\pdfoutput=1
\documentclass[10pt,twocolumn,letterpaper]{article}
\usepackage[pagenumbers]{cvpr}

\usepackage{cuted} %

\usepackage{currfile} %

\usepackage{caption} %

\usepackage{etoolbox}
\iftoggle{cvprfinal}{}{
\AtBeginEnvironment{figure*}{\nolinenumbers}
\AfterEndEnvironment{figure*}{\linenumbers}
\AtBeginEnvironment{table*}{\nolinenumbers}
\AfterEndEnvironment{table*}{\linenumbers}
\AtBeginEnvironment{table}{\nolinenumbers}
\AfterEndEnvironment{table}{\linenumbers}
}

\usepackage{booktabs}
\usepackage{multirow}
\usepackage{siunitx}
\usepackage{amsmath}
\usepackage{amssymb}
\usepackage{bbm}
\usepackage{pifont}
\usepackage{colortbl}
\definecolor{npgblue}{HTML}{3C5488}
\definecolor{npgteal}{HTML}{00A087}
\definecolor{npgorange}{HTML}{E64B35}
\definecolor{npggray}{HTML}{8491B4}
\usepackage{subcaption}
\usepackage{graphicx}
\usepackage{tikz}
\usetikzlibrary{arrows.meta,backgrounds,calc,fadings,fit}
\usepackage{titletoc}
\newcommand{\cmark}{\ding{51}}
\newcommand{\xmark}{\ding{55}}

\newcommand{\myPara}[1]{\noindent\textbf{#1.}}

\usepackage{algorithm}
\usepackage{algpseudocode}

\usepackage{fontawesome5}
\usepackage[
  breaklinks,
  colorlinks=true,
  linkcolor=blue,
  citecolor=blue,
  urlcolor=blue,
  filecolor=blue
]{hyperref}

\hypersetup{
  pdftitle={EgoMyo: Physical Hand State Estimation from Egocentric Video and Wrist Electromyography},
  pdfauthor={Guanglong Sun, Kanglei Zhou, Ziteng Wang, Nana Wang, Hengkai Tan, Xiaolong Liu, Hang Su, Yi Zhong, Jun Zhu},
  pdfsubject={Physical hand state estimation from egocentric video and wrist electromyography},
  pdfkeywords={egocentric vision, electromyography, hand pose, contact, force, multimodal learning}
}

\newcommand{\egomyolink}[3]{\href{#1}{\textcolor{npgblue}{#2\enspace#3}}}
\newcommand{\linksep}{\quad\textcolor{npggray!70}{\textbar}\quad}
\newcommand{\projectpagelink}{\egomyolink{https://egomyo.github.io/}{\faGlobeAsia}{Website}}
\newcommand{\codelink}{\egomyolink{https://github.com/SunGL001/EgoMyo}{\faGithub}{Code}}
\newcommand{\faHuggingFace}{\tikz[baseline=-0.40em,x=0.46em,y=0.46em]{%
  \fill (0,0.06) circle (1);
  \fill[white] (-0.36,0.30) ellipse (0.13 and 0.18) (0.36,0.30) ellipse (0.13 and 0.18);
  \fill[white] (-0.52,-0.12) .. controls (-0.42,-0.62) and (0.42,-0.62) .. (0.52,-0.12) -- cycle;
  \foreach \s in {-1,1}{\fill[draw=white,line width=0.35pt,rotate around={\s*35:(\s*0.78,-0.72)}] (\s*0.78,-0.72) ellipse (0.40 and 0.27);}}}
\newcommand{\datasetlink}{\egomyolink{https://huggingface.co/datasets/long2333/EgoMyo}{\faHuggingFace}{Dataset}}

\AtEndPreamble{%
  \captionsetup[subfigure]{labelformat=simple}%
  \crefname{appendixfigure}{App. Fig.}{App. Figs.}%
  \Crefname{appendixfigure}{App. Fig.}{App. Figs.}%
  \crefname{appendixtable}{App. Tab.}{App. Tabs.}%
  \Crefname{appendixtable}{App. Tab.}{App. Tabs.}%
  \crefname{appendixsubsection}{App. Sec.}{App. Secs.}%
  \Crefname{appendixsubsection}{App. Sec.}{App. Secs.}%
  \crefname{appendixalgorithm}{App. Alg.}{App. Algs.}%
  \Crefname{appendixalgorithm}{App. Alg.}{App. Algs.}%
}

\makeatletter
\patchcmd{\@maketitle}{\vskip .375in}{\vskip 0pt}{}{
  \PackageError{egomyo}{Could not adjust title top spacing}{}
}
\patchcmd{\@maketitle}{\vspace*{24pt}}{\vspace*{8pt}}{}{
  \PackageError{egomyo}{Could not adjust title-author spacing}{}
}
\patchcmd{\@maketitle}{\vspace*{12pt}}{\vspace*{0pt}}{}{
  \PackageError{egomyo}{Could not adjust title bottom spacing}{}
}
\makeatother

\title{EgoMyo: Physical Hand State Estimation \\ from Egocentric Video and Wrist Electromyography}

\author{%
  \normalsize Guanglong Sun\textsuperscript{1,2,*}\quad
  Kanglei Zhou\textsuperscript{1,*}\quad
  Ziteng Wang\textsuperscript{3}\quad
  Nana Wang\textsuperscript{1,4}\quad
  Hengkai Tan\textsuperscript{1,2}\\[2pt]
  \normalsize Xiaolong Liu\textsuperscript{1}\quad
  Hang Su\textsuperscript{1}\quad
  Yi Zhong\textsuperscript{1}\quad
  Jun Zhu\textsuperscript{1,2,\textdagger}\\[3pt]
  \small\textsuperscript{1}Tsinghua University\quad
    \textsuperscript{2}GensPI Technology\quad
  \textsuperscript{3}AstrAlive Intelligence\quad
  \textsuperscript{4}Cortix AI\\[2pt]
  \footnotesize\textsuperscript{*}Equal contribution.\quad
  \textsuperscript{\textdagger}Corresponding author.\quad
  \href{mailto:dcszj@tsinghua.edu.cn}{\textcolor{black}{\texttt{dcszj@tsinghua.edu.cn}}}\\[3pt]
  \small\projectpagelink\linksep\codelink\linksep\datasetlink
}

\begin{document}
\maketitle
\begin{strip}
  \definecolor{tVis}{HTML}{3C5488}\definecolor{tEmg}{HTML}{E64B35}%
\definecolor{tSurf}{HTML}{00A087}\definecolor{tInk}{HTML}{2B2B2B}\definecolor{tTouch}{HTML}{7E6148}%
\definecolor{tFuse}{HTML}{8491B4}\definecolor{tMute}{HTML}{858B93}\definecolor{tReg}{HTML}{7E6148}%
\definecolor{rT}{HTML}{E64B35}\definecolor{rI}{HTML}{4DBBD5}\definecolor{rM}{HTML}{00A087}%
\definecolor{rR}{HTML}{3C5488}\definecolor{rP}{HTML}{F39B7F}\definecolor{rU}{HTML}{8491B4}%
\definecolor{rL}{HTML}{91D1C2}%
\def\teaserTint{4}%
\begin{minipage}{\textwidth}
\centering
\newcommand{\teaserFont}{\fontfamily{phv}\fontsize{6}{7}\selectfont}
\def\teaserFrameTime{2.70}%

\resizebox{\textwidth}{!}{%
{\teaserFont%
\begin{tikzpicture}[
  y=0.95cm,
  font=\teaserFont, line width=0.5pt, >={Stealth[length=1.4mm]},
  panel/.style={draw=none, fill=#1!\teaserTint, rounded corners=4pt, anchor=south west, inner sep=0pt},
  ptitle/.style={font=\teaserFont\bfseries, text=#1, anchor=base west, inner sep=0pt},
  cap/.style={text=#1, align=center, inner sep=1pt},
  score/.style={font=\teaserFont, text=tInk, align=left, anchor=west, inner sep=0pt},
  imgframe/.style={inner sep=0pt, draw=#1, line width=0.6pt},
  stat/.style={draw=#1!45, fill=white, rounded corners=2pt, minimum width=17mm, minimum height=10mm, inner sep=1pt, align=center},
  chip/.style={draw=#1!55, fill=#1!10, text=#1!80!black, rounded corners=1.5pt, inner xsep=1.5pt, inner ysep=1.4pt, font=\teaserFont, anchor=west},
  sub/.style={font=\teaserFont\bfseries, text=black, anchor=west, inner sep=0pt},
]
\def\HandPath{(-0.95,-3.75) .. controls (-1.00,-3.10) and (-1.40,-2.45) .. (-1.85,-1.75) .. controls (-2.22,-1.18) and (-2.85,-0.68) .. (-3.08,-0.14) .. controls (-3.29,0.36) and (-2.91,0.77) .. (-2.54,0.44) .. controls (-2.15,0.05) and (-1.76,-0.45) .. (-1.44,-0.56) .. controls (-1.18,-0.58) and (-1.17,0.00) .. (-1.20,0.35) .. controls (-1.23,0.85) and (-1.46,2.25) .. (-1.43,2.82) .. controls (-1.42,3.35) and (-0.85,3.40) .. (-0.71,2.90) .. controls (-0.58,2.25) and (-0.50,0.95) .. (-0.39,0.48) .. controls (-0.35,0.30) and (-0.22,0.28) .. (-0.18,0.52) .. controls (-0.10,1.25) and (-0.23,3.05) .. (-0.17,3.55) .. controls (-0.12,4.13) and (0.48,4.13) .. (0.55,3.55) .. controls (0.58,2.85) and (0.51,1.13) .. (0.59,0.59) .. controls (0.63,0.35) and (0.78,0.36) .. (0.83,0.60) .. controls (0.92,1.25) and (1.00,2.55) .. (1.07,3.05) .. controls (1.14,3.57) and (1.72,3.53) .. (1.78,3.00) .. controls (1.79,2.30) and (1.61,0.88) .. (1.67,0.31) .. controls (1.70,0.12) and (1.83,0.06) .. (1.90,0.27) .. controls (2.07,0.80) and (2.29,1.56) .. (2.40,1.95) .. controls (2.54,2.47) and (3.04,2.34) .. (3.05,1.89) .. controls (3.01,1.36) and (2.57,-0.06) .. (2.41,-0.58) .. controls (2.25,-1.61) and (2.02,-2.79) .. (1.52,-3.75) -- cycle}%
\def\tipT{(-2.75,0.30)}\def\tipI{(-1.07,2.90)}\def\tipM{(0.19,3.55)}\def\tipR{(1.42,3.00)}
\newcommand{\TeaserHand}[3]{%
  \begin{scope}[shift={(#1)}, rotate=#2, scale=0.118]
    \fill[white] \HandPath; \draw[tMute, line width=0.45pt] \HandPath; #3
  \end{scope}}
\newcommand{\MeshDots}{\foreach \p in {(-2.6,0.2),(-2.1,-0.4),(-1.5,-1.1),(-1.0,2.6),(-1.0,1.4),(-0.9,0.3),(0.2,3.3),(0.2,2.0),(0.15,0.6),(1.35,2.8),(1.3,1.6),(1.2,0.4),(2.55,1.7),(2.3,0.7),(-0.5,-1.5),(0.6,-1.4),(1.5,-1.2),(-0.2,-2.8),(1.0,-2.8)} \fill[tMute] \p circle (0.17);}
\newcommand{\GTOutline}{\draw[tMute!55, densely dashed, line width=0.4pt] \HandPath;}
\newcommand{\Contact}[1]{\draw[black, line width=0.55pt] #1 circle (0.75);}
\newcommand{\Missed}[1]{\draw[tVis, line width=0.5pt, densely dotted] #1 circle (0.75);}
\newcommand{\Force}[2]{\shade[inner color=red!90!orange, outer color=orange!0] #1 circle (#2);}
\def\teaserPanelHeight{42.5mm}
\begin{scope}[on background layer]
  \node[draw=none, fill=tSurf!\teaserTint, rounded corners=4pt, anchor=south west, inner sep=0pt, minimum width=58mm, minimum height=\teaserPanelHeight, outer sep=0pt, text height=0pt, text depth=0pt] (teaserLeftPanel) at (0,0) {};
  \node[panel=tSurf, minimum width=114.75mm, minimum height=\teaserPanelHeight, outer sep=0pt, text height=0pt, text depth=0pt] (teaserRightPanel) at (6.00,0) {};
\end{scope}
\begin{scope}
\begin{scope}[yshift=0.9mm]
\def\paX{0.14}\def\paW{5.52}
\colorlet{paSub}{tInk!72}
\tikzset{paCard/.style={fill=white, fill opacity=0.75, draw=white, draw opacity=1,
                        line width=0.5pt, rounded corners=2.4pt}}
\pgfmathsetmacro{\paC}{\paX+\paW/2}
\def\paHead{3.992}
\node[anchor=base west, inner sep=0pt,
      font=\fontfamily{phv}\fontsize{9}{10}\selectfont\bfseries, text=tSurf]
  at (\paX,\paHead) {EgoMyo};
\node[anchor=base east, inner sep=0pt,
      font=\fontfamily{phv}\fontsize{4.8}{5.5}\selectfont, text=paSub]
  at (\paX+\paW,\paHead)
  {ego video $+$ sEMG $\rightarrow$ pose $\cdot$ contact $\cdot$ force};
\draw[tSurf, line width=0.6pt] (\paX,3.846) -- (\paX+\paW,3.846);
\newcommand{\regT}{(-4,1.2) -- (-1.30,0.30) -- (-1.62,-2.40) -- (-4,-2.40) -- cycle}
\newcommand{\regI}{(-1.62,0.35) rectangle (-0.28,4.5)}
\newcommand{\regM}{(-0.28,0.35) rectangle (0.71,4.5)}
\newcommand{\regR}{(0.71,0.35) rectangle (1.73,4.5)}
\newcommand{\regP}{(1.73,0.05) rectangle (3.6,4.5)}
\newcommand{\regU}{(-1.40,-1.55) rectangle (4,0.35)}
\newcommand{\regL}{(-4,-4.2) rectangle (4,-1.55)}
\newcommand{\miniHand}[4]{%
  \begin{scope}[shift={(#1,#2)}, x=1cm, y=1cm, scale=0.112]
    \begin{scope}
      \clip \HandPath;
      \foreach \r in {#3}{\fill[r\r!65] \csname reg\r\endcsname;}
      \draw[tMute!35, line width=0.2pt]
        (-1.30,0.30) -- (-1.62,-2.40)  (-1.62,0.35) -- (1.73,0.35)
        (-0.28,0.35) -- (-0.28,4.5) (0.71,0.35) -- (0.71,4.5) (1.73,0.05) -- (1.73,4.5)
        (-1.55,-1.55) -- (4,-1.55);
    \end{scope}
    \draw[tInk!65, line width=0.35pt] \HandPath;
  \end{scope}
  \node[anchor=base, inner sep=0pt, text=paSub,
        font=\fontfamily{phv}\fontsize{4.4}{5}\selectfont] at (#1-0.013,#2-0.58) {#4};}
\tikzset{paIco/.style={x=1cm, y=1cm, line join=round, line cap=round}}
\newcommand{\icoPeople}{\foreach \i in {0,...,8}\foreach \j in {0,...,3}
  \fill[\pc] (0.02+\i*0.057,-0.085+\j*0.057) circle (0.017);}
\newcommand{\icoHours}{%
  \fill[\pc, rounded corners=0.3pt] (0,-0.13) rectangle (0.48,-0.10);
  \foreach \x in {0,0.12,0.24,0.36,0.48} \draw[\pc, line width=0.5pt] (\x,-0.10) -- (\x,0.12);
  \foreach \x in {0,0.12,0.24,0.36} \foreach \d in {0.03,0.06,0.09}
    \draw[\pc, line width=0.25pt] (\x+\d,-0.10) -- (\x+\d,0.0);}
\newcommand{\icoTrials}{%
  \foreach \k/\h in {0/0.07,1/0.11,2/0.15,3/0.19,4/0.23}
    \draw[\pc, fill=\pc!12, line width=0.3pt, rounded corners=0.4pt]
      (\k*0.083,-0.13) rectangle ++(0.058,\h);
  \fill[\pc, rounded corners=0.4pt] (5*0.083,-0.13) rectangle ++(0.058,0.27);}
\newcommand{\icoHand}[3]{%
  \begin{scope}[shift={(#1,0)}, rotate=#2, scale=#3]
    \fill[\pc!12] \HandPath; \draw[\pc, line width=0.35pt] \HandPath;
  \end{scope}}
\tikzset{paLine/.style={\pc, line width=0.35pt}}
\newcommand{\icoConfig}{%
  \foreach \k in {0,...,35}{
    \pgfmathtruncatemacro{\cc}{mod(\k,9)}\pgfmathtruncatemacro{\rr}{div(\k,9)}
    \ifnum\k<34 \fill[\pc] (0.03+\cc*0.052,0.075-\rr*0.05) circle (0.016);
    \else \draw[paLine] (0.03+\cc*0.052,0.075-\rr*0.05) circle (0.014); \fi}}
\newcommand{\icoDon}{%
  \draw[paLine] (0.05,-0.03) -- (0.05,-0.13) (0.45,-0.03) -- (0.45,-0.13);
  \draw[paLine] (0.25,-0.03) ellipse (0.20 and 0.045);
  \draw[paLine] (0.05,-0.13) arc[start angle=180, end angle=360, x radius=0.20, y radius=0.045];
  \foreach \x in {0.15,0.25,0.35} \fill[\pc] (\x,-0.115) circle (0.010);
  \draw[paLine, -{Stealth[length=0.7mm,width=0.7mm]}]
    (0.06,0.06) arc[start angle=165, end angle=15, x radius=0.195, y radius=0.07];}
\newcommand{\icoCams}{%
  \draw[paLine] (0.10,0.07)--(0.21,0.02) (0.40,0.07)--(0.29,0.02)
                (0.10,-0.07)--(0.21,-0.02) (0.40,-0.07)--(0.29,-0.02);
  \foreach \x/\y in {0.0/0.04,0.40/0.04,0.0/-0.12,0.40/-0.12}
    \draw[paLine, rounded corners=0.3pt] (\x,\y) rectangle ++(0.10,0.08);
  \draw[paLine] (0.25,0) circle (0.04);}
\newcommand{\icoLegend}{%
  \foreach \c [count=\k from 0] in {rT,rI,rM,rR,rP,rU,rL}
    \fill[\c!70, rounded corners=0.3pt] (\k*0.072,-0.045) rectangle ++(0.060,0.09);}
\colorlet{paIcoC}{tInk!60}\def\pc{paIcoC}
\def\paGap{0.13}                       %
\pgfmathsetmacro{\paSW}{\paW/2-0.95-\paGap}   %
\def\paSG{0.11}\def\paSTop{3.736}\def\paSBot{1.123}  %
\pgfmathsetmacro{\paSH}{(\paSTop-\paSBot-\paSG)/2}
\newcommand{\paSide}[5]{%
  \draw[paCard]
    (#1,#2) rectangle ++(\paSW,-\paSH);
  \pgfmathsetmacro{\ym}{#2-\paSH/2}
  \node[anchor=west, inner sep=0pt] at (#1+0.15,\ym)
    {\begin{tabular}{@{}l@{}}
       \fontfamily{phv}\fontsize{11}{12}\selectfont\bfseries\color{tInk}#4\\[0.3pt]
       \fontfamily{phv}\fontsize{4.6}{5.4}\selectfont\color{paSub}%
       \begin{tabular}[t]{@{}l@{}}#5\end{tabular}
     \end{tabular}};
  \begin{scope}[shift={(#1+\paSW-0.58,\ym+0.22)}, paIco, scale=0.85] #3 \end{scope}}
\pgfmathsetmacro{\paRX}{\paX+\paW-\paSW}
\draw[paCard] (\paX+\paSW+0.95*\paSG,\paSTop) rectangle (\paRX-0.95*\paSG,\paSBot);
\begin{scope}[x=1cm, y=1cm]
  \def\mhL{\paC-0.457}\def\mhR{\paC+0.483}
  \miniHand{\mhL}{2.966}{T,I}{pinch}
  \miniHand{\mhR}{2.966}{T,I,M}{tripod}
  \miniHand{\mhL}{1.826}{I,M,R,P}{finger press}
  \miniHand{\mhR}{1.826}{U,L}{palm press}
\end{scope}
\paSide{\paX}{\paSTop}{\icoConfig}{34}{contact\\configurations}
\paSide{\paX}{\paSTop-\paSH-\paSG}{\icoDon}{2}{donning\\sessions}
\paSide{\paRX}{\paSTop}{\icoLegend}{7}{hand\\regions}
\paSide{\paRX}{\paSTop-\paSH-\paSG}{\icoCams}{4}{external\\cameras}
\draw[paCard]
  (\paX,0.053) rectangle (\paX+\paW,1.013);
\foreach \k in {1,2}
  \draw[tMute!20, line width=0.3pt] (\paX+\k*\paW/3,0.181) -- (\paX+\k*\paW/3,0.885);
\newcommand{\paStat}[3]{%
  \node[anchor=base, inner sep=0pt] at (#1,0.553)
    {\fontfamily{phv}\fontsize{11}{11}\selectfont\bfseries\textcolor{tInk}{#2}};
  \node[anchor=base, inner sep=0pt] at (#1,0.303)
    {\fontfamily{phv}\fontsize{4.6}{5}\selectfont\textcolor{paSub}{#3}};}
\paStat{\paX+\paW/6}{50}{participants}
\paStat{\paX+\paW/2}{5.75M}{video frames}
\paStat{\paX+5*\paW/6}{33{,}989}{trials}

\end{scope}
\node[sub, anchor=base west] at (6.15,4.20) {INPUTS};
\node[imgframe=tVis, anchor=north west] (ego) at (6.15,4.05)
  {\includegraphics[width=26mm]{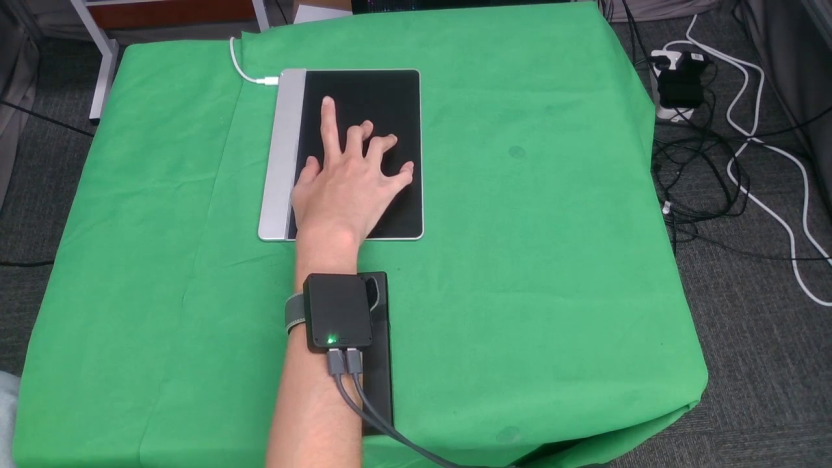}};
\node[cap=tVis, anchor=north] (egocap) at ([yshift=-0.6mm]ego.south) {egocentric video $V$};
\node[font=\teaserFont, text=white, fill=black!55, fill opacity=0.45, text opacity=1,
      rounded corners=0.6pt, inner xsep=1.2pt, inner ysep=0.5pt,
      anchor=south east, inner sep=0pt] at ([xshift=-0.9pt,yshift=0.9pt]ego.south east) {\teaserFrameTime\,s};
\node[imgframe=tEmg, fill=white, minimum width=26mm, minimum height=15mm, anchor=north west] (emg) at ([yshift=-0.9mm]egocap.south -| ego.west) {};
\node[inner sep=0pt] at (emg.center)
  {\includegraphics[width=24.5mm, height=13.5mm]{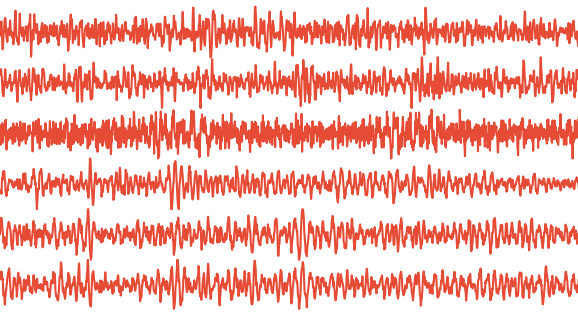}};
\node[cap=tEmg, anchor=north] at ([yshift=-0.6mm]emg.south) {wrist sEMG $E$};
\draw[tMute!30, line width=0.4pt] (8.95,0.08) -- (8.95,4.35);
\def\lab{9.17}
\def\cGT{10.43}\def\cV{12.00}\def\cT{13.45}\def\cF{14.90}\def\cO{16.35}
\foreach \x/\t/\c in {\cGT/GT/tInk,\cV/WiLoR/tVis,\cT/EgoPressure/tTouch,\cF/EMGFormer/tFuse,\cO/MyoBridge/tSurf}{
  \node[cap=\c, font=\teaserFont\bfseries, anchor=base] at (\x,4.20) {\t};}
\node[font=\teaserFont, text=tInk, rotate=90, anchor=center] at (\lab,1.08) {Total force (N)};
\foreach \m/\x/\s/\y/\h in {gt/\cGT/3.4mm/3.20/16mm,wilor/\cV/2.2mm/3.20/16mm,egopressure/\cT/2.2mm/3.20/16mm,emgformer/\cF/2.2mm/3.20/16mm,ours/\cO/2.2mm/3.20/16mm}{
  \IfFileExists{figs/teaser/\m_field.png}{%
    \node[inner sep=0pt] at ([xshift=\s]\x,\y) {\includegraphics[width=18mm,height=\h,keepaspectratio]{figs/teaser/\m_field.png}};}{}}
\node[inner sep=0pt, anchor=south west] at (9.64,3.24) {\includegraphics[width=0.12cm,height=0.855cm]{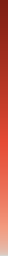}};
\draw[tInk, line width=.4pt] (9.64,3.24) rectangle (9.76,4.14);
\draw[tInk, line width=.4pt] (9.64,3.240) -- (9.540,3.240) node[left=1pt, inner sep=0pt, font=\teaserFont, text=tInk] {0};
\draw[tInk, line width=.4pt] (9.64,3.487) -- (9.540,3.487) node[left=1pt, inner sep=0pt, font=\teaserFont, text=tInk] {1};
\draw[tInk, line width=.4pt] (9.64,3.733) -- (9.540,3.733) node[left=1pt, inner sep=0pt, font=\teaserFont, text=tInk] {4};
\draw[tInk, line width=.4pt] (9.64,4.140) -- (9.540,4.140) node[left=1pt, inner sep=0pt, font=\teaserFont, text=tInk] {13};
\node[font=\teaserFont, text=tInk, rotate=90, anchor=center] at (9.13,3.69) {Force (N)};

\def\fmWilorPose{\phantom{0}6.77}
\def\fmWilorContact{\phantom{0}0.22}
\def\fmWilorForce{\phantom{0}4.18}
\def\fmEgoPressurePose{10.39}
\def\fmEgoPressureContact{\phantom{0}0.25}
\def\fmEgoPressureForce{10.27}
\def\fmEMGFormerPose{\phantom{0}9.06}
\def\fmEMGFormerContact{\phantom{0}0.25}
\def\fmEMGFormerForce{\phantom{0}5.85}
\def\fmOursPose{\phantom{0}4.09}
\def\fmOursContact{\phantom{0}0.80}
\def\fmOursForce{\phantom{0}1.39}
\def\fmBestPose{Ours}
\def\fmBestContact{Ours}
\def\fmBestForce{Ours}
  \begin{scope}[every node/.style={font=\teaserFont, inner sep=0pt, anchor=base west}]
    \foreach \y/\t in {2.66/Force MAE $\downarrow$, 2.42/Contact F1 $\uparrow$, 2.18/Pose MPJPE $\downarrow$}{%
      \node[text=tInk] at (9.04,\y) {\t};}
  \end{scope}%
  \begin{scope}[every node/.style={font=\teaserFont, inner sep=0pt, anchor=base west}]
    \foreach \x/\c/\k in {11.33/tVis/Wilor, 12.78/tTouch/EgoPressure, 14.23/tFuse/EMGFormer}{%
      \node[text=\c] at (\x,2.18) {\csname fm\k Pose\endcsname};
      \node[text=\c] at (\x,2.42) {\csname fm\k Contact\endcsname};
      \node[text=\c] at (\x,2.66) {\csname fm\k Force\endcsname};}
    \node[text=tSurf, font=\teaserFont\bfseries] at (15.68,2.18) {\fmOursPose};
    \node[text=tSurf, font=\teaserFont\bfseries] at (15.68,2.42) {\fmOursContact};
    \node[text=tSurf, font=\teaserFont\bfseries] at (15.68,2.66) {\fmOursForce};
  \end{scope}
\draw[tMute!22, line width=.3pt] (9.850,1.125) -- (17.150,1.125);
\draw[tMute!22, line width=.3pt] (9.850,1.829) -- (17.150,1.829);
\draw[tMute!22, line width=.3pt] (11.245,0.420) -- (11.245,1.970);
\draw[tMute!22, line width=.3pt] (12.640,0.420) -- (12.640,1.970);
\draw[tMute!22, line width=.3pt] (14.035,0.420) -- (14.035,1.970);
\draw[tMute!22, line width=.3pt] (15.430,0.420) -- (15.430,1.970);
\draw[tMute!22, line width=.3pt] (16.825,0.420) -- (16.825,1.970);
\draw[tVis, line width=0.55pt, line join=round] (9.850,0.420) -- (9.896,0.420) -- (9.943,0.420) -- (9.989,0.420) -- (10.036,0.420) -- (10.082,0.420) -- (10.129,0.420) -- (10.175,0.420) -- (10.222,0.420) -- (10.268,0.420) -- (10.315,0.420) -- (10.361,0.420) -- (10.408,0.420) -- (10.454,0.420) -- (10.501,0.420) -- (10.547,0.420) -- (10.594,0.420) -- (10.640,0.420) -- (10.687,0.420) -- (10.733,0.420) -- (10.780,0.420) -- (10.826,0.420) -- (10.873,0.420) -- (10.919,0.420) -- (10.966,0.420) -- (11.012,0.420) -- (11.059,0.420) -- (11.105,0.420) -- (11.152,0.421) -- (11.198,0.421) -- (11.245,0.429) -- (11.291,0.534) -- (11.338,0.584) -- (11.384,0.813) -- (11.431,0.927) -- (11.477,0.978) -- (11.524,1.061) -- (11.570,1.094) -- (11.617,1.156) -- (11.663,1.161) -- (11.710,1.198) -- (11.756,1.221) -- (11.803,1.239) -- (11.849,1.219) -- (11.896,1.220) -- (11.942,1.212) -- (11.989,1.210) -- (12.035,1.235) -- (12.082,1.244) -- (12.128,1.243) -- (12.175,1.245) -- (12.221,1.251) -- (12.268,1.245) -- (12.314,1.252) -- (12.361,1.254) -- (12.407,1.242) -- (12.454,1.239) -- (12.500,1.271) -- (12.547,1.277) -- (12.593,1.287) -- (12.640,1.248) -- (12.686,1.242) -- (12.733,1.232) -- (12.779,1.246) -- (12.826,1.243) -- (12.872,1.241) -- (12.919,1.241) -- (12.965,1.244) -- (13.012,1.236) -- (13.058,1.226) -- (13.105,1.245) -- (13.151,1.248) -- (13.198,1.224) -- (13.244,1.216) -- (13.291,1.219) -- (13.337,1.247) -- (13.384,1.217) -- (13.430,1.237) -- (13.477,1.228) -- (13.523,1.218) -- (13.570,1.203) -- (13.616,1.211) -- (13.663,1.204) -- (13.709,1.205) -- (13.756,1.206) -- (13.802,1.206) -- (13.849,1.186) -- (13.895,1.159) -- (13.942,1.156) -- (13.988,1.140) -- (14.035,1.065) -- (14.081,1.085) -- (14.128,1.080) -- (14.174,1.050) -- (14.221,1.087) -- (14.267,1.093) -- (14.314,1.043) -- (14.360,1.049) -- (14.407,1.062) -- (14.453,1.034) -- (14.500,0.986) -- (14.546,0.898) -- (14.593,0.912) -- (14.639,0.855) -- (14.686,0.859) -- (14.732,0.854) -- (14.779,0.824) -- (14.825,0.855) -- (14.872,0.820) -- (14.918,0.753) -- (14.965,0.831) -- (15.011,0.801) -- (15.058,0.816) -- (15.104,0.817) -- (15.151,0.747) -- (15.197,0.801) -- (15.244,0.827) -- (15.290,0.750) -- (15.337,0.698) -- (15.383,0.621) -- (15.430,0.629) -- (15.476,0.659) -- (15.523,0.630) -- (15.569,0.616) -- (15.616,0.620) -- (15.662,0.623) -- (15.709,0.572) -- (15.755,0.464) -- (15.802,0.438) -- (15.848,0.420) -- (15.895,0.420) -- (15.941,0.420) -- (15.988,0.420) -- (16.034,0.420) -- (16.081,0.420) -- (16.127,0.420) -- (16.174,0.420) -- (16.220,0.420) -- (16.267,0.420) -- (16.313,0.420) -- (16.360,0.420) -- (16.406,0.420) -- (16.453,0.420) -- (16.499,0.420) -- (16.546,0.420) -- (16.592,0.420) -- (16.639,0.420) -- (16.685,0.420) -- (16.732,0.420) -- (16.778,0.420) -- (16.825,0.420) -- (16.871,0.420) -- (16.918,0.420) -- (16.964,0.420) -- (17.011,0.420) -- (17.057,0.420) -- (17.104,0.420) -- (17.150,0.420);
\draw[tTouch, line width=0.55pt, line join=round] (9.850,0.420) -- (9.896,0.420) -- (9.943,0.420) -- (9.989,0.420) -- (10.036,0.420) -- (10.082,0.420) -- (10.129,0.420) -- (10.175,0.420) -- (10.222,0.420) -- (10.268,0.420) -- (10.315,0.420) -- (10.361,0.420) -- (10.408,0.420) -- (10.454,0.420) -- (10.501,0.420) -- (10.547,0.420) -- (10.594,0.420) -- (10.640,0.420) -- (10.687,0.420) -- (10.733,0.420) -- (10.780,0.420) -- (10.826,0.420) -- (10.873,0.420) -- (10.919,0.420) -- (10.966,0.420) -- (11.012,0.420) -- (11.059,0.420) -- (11.105,0.420) -- (11.152,0.420) -- (11.198,0.420) -- (11.245,0.442) -- (11.291,0.448) -- (11.338,0.458) -- (11.384,0.465) -- (11.431,0.503) -- (11.477,0.543) -- (11.524,0.705) -- (11.570,0.756) -- (11.617,0.640) -- (11.663,0.789) -- (11.710,0.744) -- (11.756,0.772) -- (11.803,0.791) -- (11.849,1.240) -- (11.896,1.455) -- (11.942,1.326) -- (11.989,1.236) -- (12.035,1.246) -- (12.082,1.271) -- (12.128,1.247) -- (12.175,1.196) -- (12.221,1.118) -- (12.268,1.133) -- (12.314,1.100) -- (12.361,1.059) -- (12.407,1.155) -- (12.454,1.199) -- (12.500,1.144) -- (12.547,1.120) -- (12.593,1.062) -- (12.640,1.068) -- (12.686,1.043) -- (12.733,0.987) -- (12.779,1.039) -- (12.826,1.089) -- (12.872,0.975) -- (12.919,1.030) -- (12.965,1.038) -- (13.012,0.970) -- (13.058,0.935) -- (13.105,1.034) -- (13.151,1.037) -- (13.198,1.119) -- (13.244,1.055) -- (13.291,1.078) -- (13.337,0.948) -- (13.384,0.944) -- (13.430,0.950) -- (13.477,0.943) -- (13.523,0.857) -- (13.570,0.891) -- (13.616,0.892) -- (13.663,0.820) -- (13.709,0.694) -- (13.756,0.648) -- (13.802,0.604) -- (13.849,0.650) -- (13.895,0.648) -- (13.942,0.676) -- (13.988,0.568) -- (14.035,0.495) -- (14.081,0.550) -- (14.128,0.530) -- (14.174,0.544) -- (14.221,0.501) -- (14.267,0.541) -- (14.314,0.562) -- (14.360,0.628) -- (14.407,0.609) -- (14.453,0.583) -- (14.500,0.692) -- (14.546,0.487) -- (14.593,0.468) -- (14.639,0.615) -- (14.686,0.605) -- (14.732,0.475) -- (14.779,0.511) -- (14.825,0.477) -- (14.872,0.484) -- (14.918,0.474) -- (14.965,0.496) -- (15.011,0.474) -- (15.058,0.475) -- (15.104,0.461) -- (15.151,0.463) -- (15.197,0.478) -- (15.244,0.478) -- (15.290,0.481) -- (15.337,0.471) -- (15.383,0.565) -- (15.430,0.647) -- (15.476,0.482) -- (15.523,0.488) -- (15.569,0.467) -- (15.616,0.474) -- (15.662,0.517) -- (15.709,0.520) -- (15.755,0.436) -- (15.802,0.431) -- (15.848,0.421) -- (15.895,0.420) -- (15.941,0.420) -- (15.988,0.420) -- (16.034,0.420) -- (16.081,0.420) -- (16.127,0.420) -- (16.174,0.420) -- (16.220,0.420) -- (16.267,0.420) -- (16.313,0.420) -- (16.360,0.420) -- (16.406,0.420) -- (16.453,0.420) -- (16.499,0.420) -- (16.546,0.420) -- (16.592,0.420) -- (16.639,0.420) -- (16.685,0.420) -- (16.732,0.420) -- (16.778,0.420) -- (16.825,0.420) -- (16.871,0.420) -- (16.918,0.420) -- (16.964,0.420) -- (17.011,0.420) -- (17.057,0.420) -- (17.104,0.420) -- (17.150,0.420);
\draw[tFuse, line width=0.55pt, line join=round] (9.850,0.420) -- (9.896,0.420) -- (9.943,0.420) -- (9.989,0.420) -- (10.036,0.420) -- (10.082,0.420) -- (10.129,0.420) -- (10.175,0.420) -- (10.222,0.420) -- (10.268,0.420) -- (10.315,0.420) -- (10.361,0.420) -- (10.408,0.420) -- (10.454,0.420) -- (10.501,0.420) -- (10.547,0.420) -- (10.594,0.420) -- (10.640,0.420) -- (10.687,0.420) -- (10.733,0.420) -- (10.780,0.420) -- (10.826,0.420) -- (10.873,0.420) -- (10.919,0.420) -- (10.966,0.420) -- (11.012,0.420) -- (11.059,0.420) -- (11.105,0.420) -- (11.152,0.420) -- (11.198,0.438) -- (11.245,0.504) -- (11.291,0.527) -- (11.338,0.720) -- (11.384,0.859) -- (11.431,0.947) -- (11.477,1.074) -- (11.524,1.084) -- (11.570,1.131) -- (11.617,1.149) -- (11.663,1.199) -- (11.710,1.264) -- (11.756,1.330) -- (11.803,1.470) -- (11.849,1.636) -- (11.896,1.681) -- (11.942,1.650) -- (11.989,1.653) -- (12.035,1.545) -- (12.082,1.663) -- (12.128,1.568) -- (12.175,1.551) -- (12.221,1.583) -- (12.268,1.590) -- (12.314,1.655) -- (12.361,1.542) -- (12.407,1.526) -- (12.454,1.578) -- (12.500,1.600) -- (12.547,1.651) -- (12.593,1.704) -- (12.640,1.692) -- (12.686,1.637) -- (12.733,1.640) -- (12.779,1.782) -- (12.826,1.753) -- (12.872,1.777) -- (12.919,1.806) -- (12.965,1.733) -- (13.012,1.701) -- (13.058,1.656) -- (13.105,1.738) -- (13.151,1.748) -- (13.198,1.762) -- (13.244,1.660) -- (13.291,1.644) -- (13.337,1.555) -- (13.384,1.629) -- (13.430,1.511) -- (13.477,1.507) -- (13.523,1.560) -- (13.570,1.551) -- (13.616,1.599) -- (13.663,1.608) -- (13.709,1.496) -- (13.756,1.487) -- (13.802,1.295) -- (13.849,1.198) -- (13.895,1.249) -- (13.942,1.331) -- (13.988,1.297) -- (14.035,1.189) -- (14.081,1.269) -- (14.128,1.130) -- (14.174,1.132) -- (14.221,0.951) -- (14.267,0.991) -- (14.314,1.042) -- (14.360,0.999) -- (14.407,1.060) -- (14.453,1.098) -- (14.500,1.075) -- (14.546,0.969) -- (14.593,0.900) -- (14.639,0.907) -- (14.686,0.975) -- (14.732,0.881) -- (14.779,0.895) -- (14.825,0.975) -- (14.872,0.847) -- (14.918,0.804) -- (14.965,0.505) -- (15.011,0.504) -- (15.058,0.473) -- (15.104,0.452) -- (15.151,0.433) -- (15.197,0.432) -- (15.244,0.472) -- (15.290,0.496) -- (15.337,0.479) -- (15.383,0.482) -- (15.430,0.468) -- (15.476,0.525) -- (15.523,0.464) -- (15.569,0.440) -- (15.616,0.440) -- (15.662,0.431) -- (15.709,0.429) -- (15.755,0.422) -- (15.802,0.420) -- (15.848,0.420) -- (15.895,0.420) -- (15.941,0.420) -- (15.988,0.420) -- (16.034,0.420) -- (16.081,0.420) -- (16.127,0.420) -- (16.174,0.420) -- (16.220,0.420) -- (16.267,0.420) -- (16.313,0.420) -- (16.360,0.420) -- (16.406,0.420) -- (16.453,0.420) -- (16.499,0.420) -- (16.546,0.420) -- (16.592,0.420) -- (16.639,0.420) -- (16.685,0.420) -- (16.732,0.420) -- (16.778,0.420) -- (16.825,0.420) -- (16.871,0.420) -- (16.918,0.420) -- (16.964,0.420) -- (17.011,0.420) -- (17.057,0.420) -- (17.104,0.420) -- (17.150,0.420);
\draw[tSurf, line width=0.8pt, line join=round] (9.850,0.420) -- (9.896,0.420) -- (9.943,0.420) -- (9.989,0.420) -- (10.036,0.420) -- (10.082,0.420) -- (10.129,0.420) -- (10.175,0.420) -- (10.222,0.420) -- (10.268,0.420) -- (10.315,0.420) -- (10.361,0.420) -- (10.408,0.420) -- (10.454,0.420) -- (10.501,0.420) -- (10.547,0.420) -- (10.594,0.420) -- (10.640,0.420) -- (10.687,0.420) -- (10.733,0.420) -- (10.780,0.420) -- (10.826,0.420) -- (10.873,0.420) -- (10.919,0.420) -- (10.966,0.420) -- (11.012,0.420) -- (11.059,0.420) -- (11.105,0.420) -- (11.152,0.420) -- (11.198,0.420) -- (11.245,0.424) -- (11.291,0.448) -- (11.338,0.493) -- (11.384,0.628) -- (11.431,0.918) -- (11.477,1.019) -- (11.524,1.237) -- (11.570,1.540) -- (11.617,1.460) -- (11.663,1.544) -- (11.710,1.745) -- (11.756,1.641) -- (11.803,1.732) -- (11.849,1.658) -- (11.896,1.732) -- (11.942,1.664) -- (11.989,1.790) -- (12.035,1.707) -- (12.082,1.858) -- (12.128,1.861) -- (12.175,1.946) -- (12.221,1.891) -- (12.268,1.831) -- (12.314,1.797) -- (12.361,1.935) -- (12.407,1.816) -- (12.454,1.894) -- (12.500,1.962) -- (12.547,1.965) -- (12.593,1.906) -- (12.640,1.881) -- (12.686,1.812) -- (12.733,1.833) -- (12.779,1.884) -- (12.826,1.927) -- (12.872,1.889) -- (12.919,1.882) -- (12.965,1.738) -- (13.012,1.726) -- (13.058,1.807) -- (13.105,1.703) -- (13.151,1.757) -- (13.198,1.752) -- (13.244,1.743) -- (13.291,1.843) -- (13.337,1.922) -- (13.384,1.946) -- (13.430,1.933) -- (13.477,1.966) -- (13.523,1.838) -- (13.570,1.794) -- (13.616,1.613) -- (13.663,1.587) -- (13.709,1.560) -- (13.756,1.580) -- (13.802,1.517) -- (13.849,1.515) -- (13.895,1.627) -- (13.942,1.554) -- (13.988,1.466) -- (14.035,1.410) -- (14.081,1.339) -- (14.128,1.235) -- (14.174,1.223) -- (14.221,1.262) -- (14.267,1.239) -- (14.314,1.128) -- (14.360,1.138) -- (14.407,1.074) -- (14.453,1.110) -- (14.500,1.082) -- (14.546,0.983) -- (14.593,0.906) -- (14.639,0.959) -- (14.686,1.053) -- (14.732,0.970) -- (14.779,0.879) -- (14.825,0.849) -- (14.872,0.869) -- (14.918,0.824) -- (14.965,0.778) -- (15.011,0.695) -- (15.058,0.579) -- (15.104,0.554) -- (15.151,0.552) -- (15.197,0.544) -- (15.244,0.564) -- (15.290,0.554) -- (15.337,0.527) -- (15.383,0.494) -- (15.430,0.526) -- (15.476,0.518) -- (15.523,0.526) -- (15.569,0.523) -- (15.616,0.535) -- (15.662,0.528) -- (15.709,0.522) -- (15.755,0.469) -- (15.802,0.453) -- (15.848,0.424) -- (15.895,0.420) -- (15.941,0.420) -- (15.988,0.420) -- (16.034,0.420) -- (16.081,0.420) -- (16.127,0.420) -- (16.174,0.420) -- (16.220,0.420) -- (16.267,0.420) -- (16.313,0.420) -- (16.360,0.420) -- (16.406,0.420) -- (16.453,0.420) -- (16.499,0.420) -- (16.546,0.420) -- (16.592,0.420) -- (16.639,0.420) -- (16.685,0.420) -- (16.732,0.420) -- (16.778,0.420) -- (16.825,0.420) -- (16.871,0.420) -- (16.918,0.420) -- (16.964,0.420) -- (17.011,0.420) -- (17.057,0.420) -- (17.104,0.420) -- (17.150,0.420);
\draw[tInk, line width=.8pt, densely dashed, line join=round] (9.850,0.421) -- (9.896,0.421) -- (9.943,0.421) -- (9.989,0.421) -- (10.036,0.421) -- (10.082,0.421) -- (10.129,0.421) -- (10.175,0.421) -- (10.222,0.421) -- (10.268,0.421) -- (10.315,0.421) -- (10.361,0.421) -- (10.408,0.421) -- (10.454,0.421) -- (10.501,0.421) -- (10.547,0.421) -- (10.594,0.421) -- (10.640,0.421) -- (10.687,0.421) -- (10.733,0.421) -- (10.780,0.421) -- (10.826,0.421) -- (10.873,0.421) -- (10.919,0.421) -- (10.966,0.421) -- (11.012,0.421) -- (11.059,0.421) -- (11.105,0.421) -- (11.152,0.421) -- (11.198,0.425) -- (11.245,0.425) -- (11.291,0.455) -- (11.338,0.588) -- (11.384,0.712) -- (11.431,0.884) -- (11.477,1.062) -- (11.524,1.264) -- (11.570,1.397) -- (11.617,1.493) -- (11.663,1.572) -- (11.710,1.612) -- (11.756,1.654) -- (11.803,1.686) -- (11.849,1.705) -- (11.896,1.728) -- (11.942,1.750) -- (11.989,1.772) -- (12.035,1.784) -- (12.082,1.803) -- (12.128,1.817) -- (12.175,1.829) -- (12.221,1.847) -- (12.268,1.860) -- (12.314,1.884) -- (12.361,1.908) -- (12.407,1.939) -- (12.454,1.953) -- (12.500,1.958) -- (12.547,1.958) -- (12.593,1.914) -- (12.640,1.872) -- (12.686,1.867) -- (12.733,1.866) -- (12.779,1.892) -- (12.826,1.902) -- (12.872,1.911) -- (12.919,1.949) -- (12.965,1.954) -- (13.012,1.970) -- (13.058,1.960) -- (13.105,1.960) -- (13.151,1.932) -- (13.198,1.905) -- (13.244,1.877) -- (13.291,1.850) -- (13.337,1.822) -- (13.384,1.795) -- (13.430,1.768) -- (13.477,1.730) -- (13.523,1.703) -- (13.570,1.643) -- (13.616,1.623) -- (13.663,1.581) -- (13.709,1.547) -- (13.756,1.535) -- (13.802,1.491) -- (13.849,1.454) -- (13.895,1.415) -- (13.942,1.346) -- (13.988,1.346) -- (14.035,1.307) -- (14.081,1.255) -- (14.128,1.216) -- (14.174,1.169) -- (14.221,1.110) -- (14.267,1.072) -- (14.314,1.022) -- (14.360,1.011) -- (14.407,0.992) -- (14.453,0.955) -- (14.500,0.956) -- (14.546,0.944) -- (14.593,0.927) -- (14.639,0.913) -- (14.686,0.900) -- (14.732,0.879) -- (14.779,0.870) -- (14.825,0.832) -- (14.872,0.802) -- (14.918,0.768) -- (14.965,0.735) -- (15.011,0.712) -- (15.058,0.678) -- (15.104,0.641) -- (15.151,0.623) -- (15.197,0.593) -- (15.244,0.560) -- (15.290,0.549) -- (15.337,0.555) -- (15.383,0.567) -- (15.430,0.572) -- (15.476,0.557) -- (15.523,0.553) -- (15.569,0.536) -- (15.616,0.508) -- (15.662,0.478) -- (15.709,0.459) -- (15.755,0.428) -- (15.802,0.421) -- (15.848,0.420) -- (15.895,0.420) -- (15.941,0.420) -- (15.988,0.420) -- (16.034,0.420) -- (16.081,0.420) -- (16.127,0.420) -- (16.174,0.420) -- (16.220,0.420) -- (16.267,0.420) -- (16.313,0.420) -- (16.360,0.420) -- (16.406,0.420) -- (16.453,0.420) -- (16.499,0.420) -- (16.546,0.420) -- (16.592,0.420) -- (16.639,0.420) -- (16.685,0.420) -- (16.732,0.420) -- (16.778,0.420) -- (16.825,0.420) -- (16.871,0.420) -- (16.918,0.420) -- (16.964,0.420) -- (17.011,0.420) -- (17.057,0.420) -- (17.104,0.420) -- (17.150,0.420);
\draw[tMute, line width=.4pt] (13.616,0.420) -- (13.616,1.970);
\draw[tInk, line width=.4pt] (9.850,1.125) -- (9.750,1.125) node[left=1pt, inner sep=0pt, font=\teaserFont, text=tInk] {50};
\draw[tInk, line width=.4pt] (9.850,1.829) -- (9.750,1.829) node[left=1pt, inner sep=0pt, font=\teaserFont, text=tInk] {100};
\draw[tInk, line width=.4pt] (9.850,0.420) -- (9.850,0.320) node[below=1pt, inner sep=0pt, font=\teaserFont, text=tInk] {0};
\draw[tInk, line width=.4pt] (11.245,0.420) -- (11.245,0.320) node[below=1pt, inner sep=0pt, font=\teaserFont, text=tInk] {1};
\draw[tInk, line width=.4pt] (12.640,0.420) -- (12.640,0.320) node[below=1pt, inner sep=0pt, font=\teaserFont, text=tInk] {2};
\draw[tInk, line width=.4pt] (14.035,0.420) -- (14.035,0.320) node[below=1pt, inner sep=0pt, font=\teaserFont, text=tInk] {3};
\draw[tInk, line width=.4pt] (15.430,0.420) -- (15.430,0.320) node[below=1pt, inner sep=0pt, font=\teaserFont, text=tInk] {4};
\draw[tInk, line width=.4pt] (16.825,0.420) -- (16.825,0.320) node[below=1pt, inner sep=0pt, font=\teaserFont, text=tInk] {5};
\draw[tInk, line width=.5pt] (9.850,0.420) -- (17.150,0.420) (9.850,0.420) -- (9.850,1.970);
\coordinate (selT) at (13.616,1.970);
  \node[font=\teaserFont, text=tMute, fill=tSurf!\teaserTint, anchor=north west, align=center,
        inner sep=0.6pt] at (14.22,1.86) {frame shown\\above};
  \node[font=\teaserFont, text=tInk, anchor=south east, inner sep=0pt] at (17.15,0.44) {Time (s)};
  \def\tmWilorForce{14.0}
\def\tmEgoPressureForce{23.8}
\def\tmEMGFormerForce{6.6}
\def\tmOursForce{3.2}
    \node[font=\teaserFont, text=tInk, anchor=base, fill=tSurf!\teaserTint, inner sep=0.6pt] at (10.55,1.80) {Trial MAE (N)};
    \foreach \y/\c/\w/\k/\f in {1.50/tVis/.55/Wilor/, 1.25/tTouch/.55/EgoPressure/, 1.00/tFuse/.55/EMGFormer/, 0.75/tSurf/.8/Ours/\bfseries}{%
      \node[font=\teaserFont\f, text=\c, anchor=base east, inner sep=0pt] at (10.88,\y) {\csname tm\k Force\endcsname};
      \node[font=\teaserFont, anchor=base east, inner sep=0pt] (tmw) at (10.88,\y) {\phantom{00.0}};
      \draw[\c, line width=\w pt] ([xshift=-3pt,yshift=0.07cm]tmw.base west) -- ++(-0.30,0);}
  \fill[tSurf!\teaserTint] (15.76,0.76) rectangle (17.14,1.97);
  \foreach \y/\c/\w/\d/\t in {1.86/tInk/.8/densely dashed/GT, 1.61/tVis/.55/solid/WiLoR,
      1.36/tTouch/.55/solid/EgoPressure, 1.11/tFuse/.55/solid/EMGFormer, 0.86/tSurf/.8/solid/MyoBridge}{
    \def\x{15.85}
    \draw[\c, line width=\w pt, \d] (\x,\y) -- ++(0.32,0) node[right=3pt, inner sep=0pt, font=\teaserFont, text=tInk] {\t};}

\end{scope}
\end{tikzpicture}}}

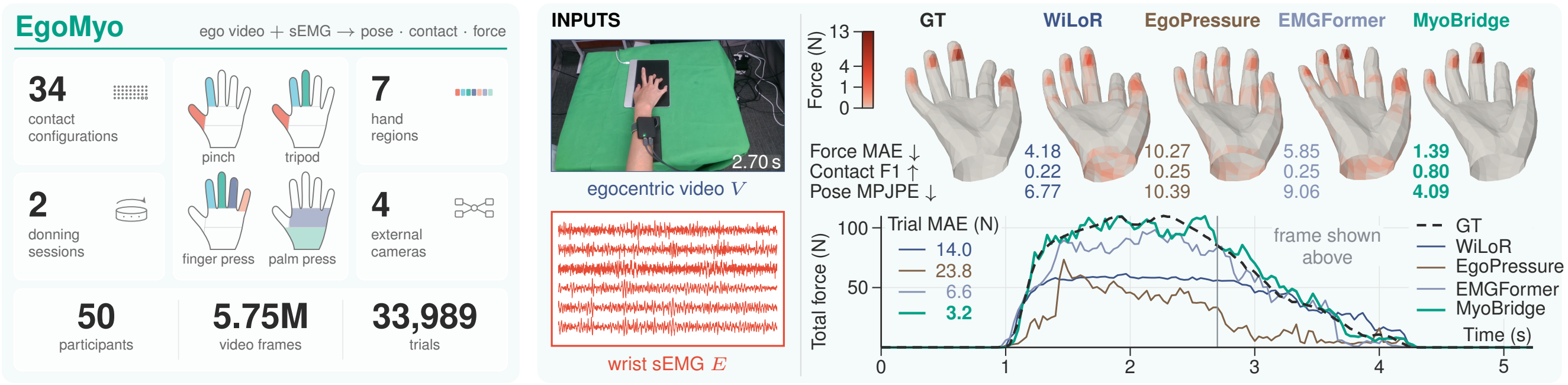
\captionof{figure}{\textbf{From complementary cues to a complete physical hand state.} Vision provides spatial geometry but misses exerted magnitude, while wrist sEMG senses exertion but not contact location. EgoMyo fills the supervision gap by synchronizing both streams with pose and bare-hand pressure on the hand mesh, making regional contact and force measurable. In the held-out trial shown, \textcolor{tVis}{WiLoR}~\cite{potamias2024wilor}, \textcolor{tTouch}{EgoPressure}~\cite{zhao2025egopressure}, and \textcolor{tFuse}{EMGFormer}~\cite{xi2026egoemg} misplace contact even with added output heads, whereas \textcolor{tSurf}{MyoBridge} routes sEMG by region and confines force to predicted contact support, achieving the lowest pose and force errors and highest contact F1.}
\label{fig:teaser}
\end{minipage}
\vspace{3pt}

\end{strip}
\begin{abstract}
Physical hand interaction depends on configuration, contact, and force, yet nearly identical appearances can conceal different contact states and exerted forces. We therefore formulate physical hand state estimation from egocentric video and wrist surface electromyography (sEMG), combining visual geometry with neuromuscular evidence to jointly recover these coupled components. To support this task, we introduce EgoMyo, a benchmark spanning 5.75M video frames, 50 participants, 34 contact configurations, and repeated donnings, synchronizing these inputs with 3D hand pose and with contact and normal force registered onto the hand surface from bare-hand pressure. Controlled analysis reveals that vision resolves spatial structure while sEMG better estimates force magnitude, that sEMG's gains concentrate in regions where vision is least reliable, and that contact localization limits force accuracy. Guided by these findings, we propose MyoBridge, which makes the hand region the unit of vision--sEMG fusion, lets each region gate its own share of sEMG, and distributes regional force magnitudes only over predicted contact vertices at inference. On held-out trials, MyoBridge reduces pose and force error by 39.8\% and 18.5\% over the strongest baseline and achieves the lowest pose error on unseen sessions and subjects.
\end{abstract}

\section{Introduction}
\label{sec:intro}

Hand sensing supports applications from augmented and virtual reality~\cite{kim2012digits,han2020megatrack,zhou2021stgae,zhou2023mixed} to human--computer interaction~\cite{mistry2009sixthsense,harrison2011omnitouch}, teleoperation~\cite{qin2023anyteleop,huang2025dihtele}, and dexterous manipulation~\cite{handa2020dexpilot,an2026dexterous,jiang2025dexmimicgen}. Many of these applications depend on hand motion and interaction with the environment~\cite{huang2025dihtele,an2026dexterous}, yet nearly identical configurations can correspond to no contact, light touch, or firm pressing~\cite{grady2022pressurevision}. We therefore define the \textbf{physical hand state} by three coupled components. \emph{Kinematics} describe hand configuration, \emph{contact} specifies where the hand touches its environment, and \emph{interaction force} captures how strongly it acts. Egocentric vision naturally observes the hand and its interaction scene from the user's viewpoint~\cite{han2020megatrack,banerjee2025hot3d}. However, kinematics can often be inferred from appearance, whereas contact and force are weakly observable and easily obscured by occlusion.

Existing methods predominantly rely on vision, progressing from hand geometry recovery~\cite{rong2021frankmocap,park2022handoccnet,pavlakos2024hamer,zhang2025hawor} to contact and force estimation~\cite{grady2022pressurevision,zhao2025egopressure,ilic2026egophi}, but remain limited to external manifestations of interaction. Wrist surface electromyography (sEMG) instead senses the neuromuscular drive underlying articulation and exertion~\cite{liu2021neuropose,salter2024emg2pose}, with muscle activation scaling with exerted force, but lacks scene context and varies across sessions~\cite{lippold1952relation,yang2024emgbench}. The modalities are complementary (\cref{fig:teaser}), yet existing combinations remain partial. PiMForce~\cite{seo2024posture} estimates pressure from sEMG conditioned on externally captured posture, ForceBand~\cite{he2026forceband} assigns force to sEMG and motion to video in largely separate roles, and EMGFormer~\cite{xi2026egoemg} fuses egocentric video and sEMG for pose only. Joint physical hand state estimation from both modalities remains unexplored, raising a central question. \emph{Can egocentric vision and wrist sEMG jointly reveal hand configuration, contact, and force?}

Answering this question is nontrivial at two levels. At the \emph{benchmark level}, reliable contact and force supervision requires instrumented measurements synchronized with sEMG, egocentric video, and hand pose. Assessing sEMG variability also requires repeated donnings, so existing datasets cover only subsets of this space (\cref{tab:dataset_comparison}). At the \emph{modeling level}, the streams resolve the state at different granularities. Vision supplies spatial structure but sees load only indirectly, while sEMG senses hand-level exertion that maps to no single digit, so a model must decide where sEMG evidence applies. Contact and force are coupled because force acts only where the hand touches, a constraint that independent output heads do not enforce.

\begin{table}[t]
  \centering
  \setlength{\tabcolsep}{3pt}
  \small
  \caption{\textbf{Comparison with representative datasets.} EgoMyo pairs egocentric video and wrist sEMG with bare-hand supervision of all three physical hand state components. Scale is as reported in the original papers, with ``---'' if not reported. ``Bare'' means no gloves or hand-worn sensors, and ``Force'' includes pressure.}
  \label{tab:dataset_comparison}
  \resizebox{\linewidth}{!}{
  \begin{tabular}{lcccccccc}
    \toprule
    \multirow{2.5}{*}{\textbf{Dataset}} & \multicolumn{2}{c}{\textbf{Scale}} & \multicolumn{3}{c}{\textbf{Sensing}} & \multicolumn{3}{c}{\textbf{Supervision}} \\
    \cmidrule(lr){2-3} \cmidrule(lr){4-6} \cmidrule(l){7-9}
    & Subjects & Hours & Ego & sEMG & Bare & Pose & Contact & Force \\
    \midrule
    emg2pose~\cite{salter2024emg2pose}       & 193 & 370  & \xmark & \cmark & \cmark & \cmark & \xmark & \xmark \\
    PiMForce~\cite{seo2024posture}           & 21  & ---  & \xmark & \cmark & \xmark & \cmark & \xmark & \cmark \\
    \midrule
    EgoPressure~\cite{zhao2025egopressure}   & 21  & 5    & \cmark & \xmark & \cmark & \cmark & \cmark & \cmark \\
    EgoPHI~\cite{ilic2026egophi}             & 8   & ---  & \cmark & \xmark & \cmark & \xmark & \cmark & \cmark \\
    OpenTouch~\cite{song2025opentouch}       & --- & 5.1  & \cmark & \xmark & \xmark & \cmark & \cmark & \cmark \\
    EgoTouch~\cite{zhou2026touchanything}    & --- & 20   & \cmark & \xmark & \xmark & \cmark & \cmark & \cmark \\
    EgoTactile~\cite{zeng2026egotactile}     & 12  & ---  & \cmark & \xmark & \xmark & \xmark & \cmark & \cmark \\
    \midrule
    EgoEMG~\cite{xi2026egoemg}               & 41  & $>$10 & \cmark & \cmark & \cmark & \cmark & \xmark & \xmark \\
    ForceBand~\cite{he2026forceband}         & 4   & 10   & \cmark & \cmark & \xmark & \xmark & \xmark & \cmark \\
    \midrule
    \textbf{EgoMyo (ours)}                   & 50 & 53 & \cmark & \cmark & \cmark & \cmark & \cmark & \cmark \\
    \bottomrule
  \end{tabular}
  }
\end{table}

At the benchmark level, we introduce \textbf{EgoMyo}, an \textbf{Ego}centric \textbf{Myo}electric benchmark for physical hand state estimation. It pairs egocentric video and 18-channel wrist sEMG with 3D hand pose and with contact and normal force obtained by registering bare-hand pressure maps onto a reconstructed hand mesh, covering 5.75M video frames from 50 participants who press all 34 digit and palm contact configurations across repeated wristband donnings. Controlled analysis (\cref{fig:empirical}) yields three findings that shape our design. Vision resolves spatial structure while sEMG better estimates force magnitude. sEMG's absolute force gains track vision's error ($\rho=0.93$), and force accuracy depends on correctly localized contact support.

At the modeling level, we develop \textbf{MyoBridge}, which bridges hand-level sEMG and the vertex-level surface state through region tokens. Unlike prior fusion, which acts on the whole hand~\cite{xi2026egoemg,liang2025mixture,wang2025hbridge}, MyoBridge makes the hand region the unit of fusion. Granularity-matched encoding keeps each stream in its own structure, region-gated fusion lets each region receive its share of sEMG, and region-to-surface decoding distributes sEMG-informed regional magnitudes over predicted contact. At inference, force is zero off predicted contact and sums to each regional magnitude.

Experiments show that MyoBridge lowers MPJPE by 39.8\% and regional force MAE by 18.5\% over the runner-up on held-out trials, and lowers MPJPE on unseen sessions and subjects by 49.9\% and 45.6\% over Mixture-of-Transformers (MoT)~\cite{liang2025mixture}. Our contributions are threefold:
\begin{itemize}
    \item We formulate \textbf{physical hand state estimation} from egocentric vision and wrist sEMG, defining kinematics, contact, and force jointly on hand-mesh vertices.
    \item We introduce \textbf{EgoMyo}, to our knowledge the first benchmark to pair egocentric video and wrist sEMG with bare-hand pose, contact, and force, with 50 participants, 5.75M video frames, 34 contact configurations, and two donnings, and analyze where sEMG complements vision.
    \item We develop \textbf{MyoBridge}, which follows these findings to fuse vision and sEMG per region and confine force to predicted contact by construction at inference.
\end{itemize}

\section{Related Work}
\label{sec:related}

\myPara{Vision-based Hand Estimation}
Multi-camera capture enabled large-scale 3D pose benchmarks~\cite{moon2020interhand,ohkawa2023assemblyhands}, and MANO-based~\cite{romero2017mano} monocular methods recover hand pose and mesh from single images~\cite{zimmermann2017handpose,ge2019handpose,hasson2019handobject,rong2021frankmocap,lin2021posemesh,park2022handoccnet,pavlakos2024hamer}, improved by regression~\cite{huang2020awr,zheng2021jointaware}, part-structure reasoning~\cite{zheng2021sar,ren2023poseguided}, multi-view self-supervision~\cite{zheng2023hamuco}, and temporal modeling~\cite{zhou2021stgae,zhou2024multitask,zhang2025hawor}. Egocentric benchmarks carried these tools to first-person hand--object interaction~\cite{kwon2021h2o,liu2022hoi4d,banerjee2025hot3d}. Geometry alone, however, does not uniquely determine contact or force magnitude, so later work extends vision to contact and force, with contact annotated for grasping~\cite{brahmbhatt2020contactpose,taheri2020grab}, force regressed from egocentric images by EgoPressure~\cite{zhao2025egopressure} and EgoPHI~\cite{ilic2026egophi}, and full-hand tactile supervision paired with video in OpenTouch~\cite{song2025opentouch}, EgoTouch~\cite{zhou2026touchanything}, and EgoTactile~\cite{zeng2026egotactile}. Visual cues remain ambiguous under occlusion, while gloves alter natural touch by covering the skin~\cite{sundaram2019stag}.

\myPara{sEMG-based Hand Estimation}
sEMG records the neuromuscular activity driving hand articulation and exertion. Early studies focused on discrete gesture and grasp recognition~\cite{amma2015hdemg,atzori2015benchmark,du2017emgdomain,kaczmarek2019putemg}, followed by continuous finger kinematics~\cite{ngeo2014emgkinematics} and 3D hand pose estimation with NeuroPose~\cite{liu2021neuropose} and vemg2pose~\cite{salter2024emg2pose}. Wristband decoders trained on thousands of users generalize across people~\cite{ctrllabs2025generic} but require large-scale collection. Beyond kinematics, sEMG has also been used for pressure and force estimation~\cite{seo2024posture,he2026forceband}. However, electrode shift and physiological changes cause substantial cross-session and -user variation~\cite{yang2024emgbench}, and without scene context, sEMG alone cannot reliably determine whether finger activation results in contact or free motion.

\myPara{Vision--sEMG Hand Estimation}
Recent work combines sEMG with visual or geometric observations. PiMForce~\cite{seo2024posture} fuses sEMG with 3D hand posture for pressure estimation, with posture from external capture rather than a wearer-mounted view. ForceBand~\cite{he2026forceband} pairs wrist sEMG with human video for force-aware robot learning, while EMGFormer~\cite{xi2026egoemg} targets pose. ActionSense~\cite{delpreto2022actionsense} records the same streams for activity recognition rather than hand-state estimation. None jointly estimates kinematics, contact, and force from egocentric vision and wrist sEMG, or exploits their coupling (\cref{tab:dataset_comparison}). Our fusion design builds on modality-specific parameterization in multimodal transformers~\cite{liang2025mixture}, but allocates evidence per hand region and decodes the physical hand state along the dependency among its components, from contact support to force.

\section{Task and Benchmark}
\label{sec:benchmark}

Learning to infer hand configuration, contact location, and exerted force from egocentric video and wrist sEMG requires these inputs synchronized with ground-truth 3D hand pose, contact, and force. Existing datasets offer this combination only partially (\cref{tab:dataset_comparison}). Acquiring such data is challenging, since hand-worn tactile instrumentation compromises natural bare-hand interaction, whereas external pressure sensors measure force in their own coordinate frame rather than on the hand surface. EgoMyo reconciles these constraints by registering pressure maps from bare-hand interactions onto a reconstructed hand mesh, yielding contact and force directly on the hand surface, and further enumerates contact configurations across repeated wristband donnings and body poses. Below, we formalize physical hand state estimation and describe the construction of EgoMyo.

\begin{figure}[t]
  \centering
  \begin{subcaptiongroup}%
    \phantomsubcaption\label{fig:dataset:acquisition:setup}%
    \phantomsubcaption\label{fig:dataset:acquisition:regions}%
    \phantomsubcaption\label{fig:dataset:acquisition:configs}%
    \phantomsubcaption\label{fig:dataset:acquisition:video}%
    \phantomsubcaption\label{fig:dataset:acquisition:joints}%
    \phantomsubcaption\label{fig:dataset:acquisition:surface_force}%
    \phantomsubcaption\label{fig:dataset:acquisition:regional_force}%
    \phantomsubcaption\label{fig:dataset:acquisition:emg}%
    \phantomsubcaption\label{fig:dataset:acquisition:pressure}%
    \phantomsubcaption\label{fig:dataset:acquisition:contact}%
    \phantomsubcaption\label{fig:dataset:acquisition:force_regions}%
    \phantomsubcaption\label{fig:dataset:acquisition:force_participants}%
    \phantomsubcaption\label{fig:dataset:acquisition:emg_sessions}%
  \end{subcaptiongroup}%
  \includegraphics[width=\linewidth]{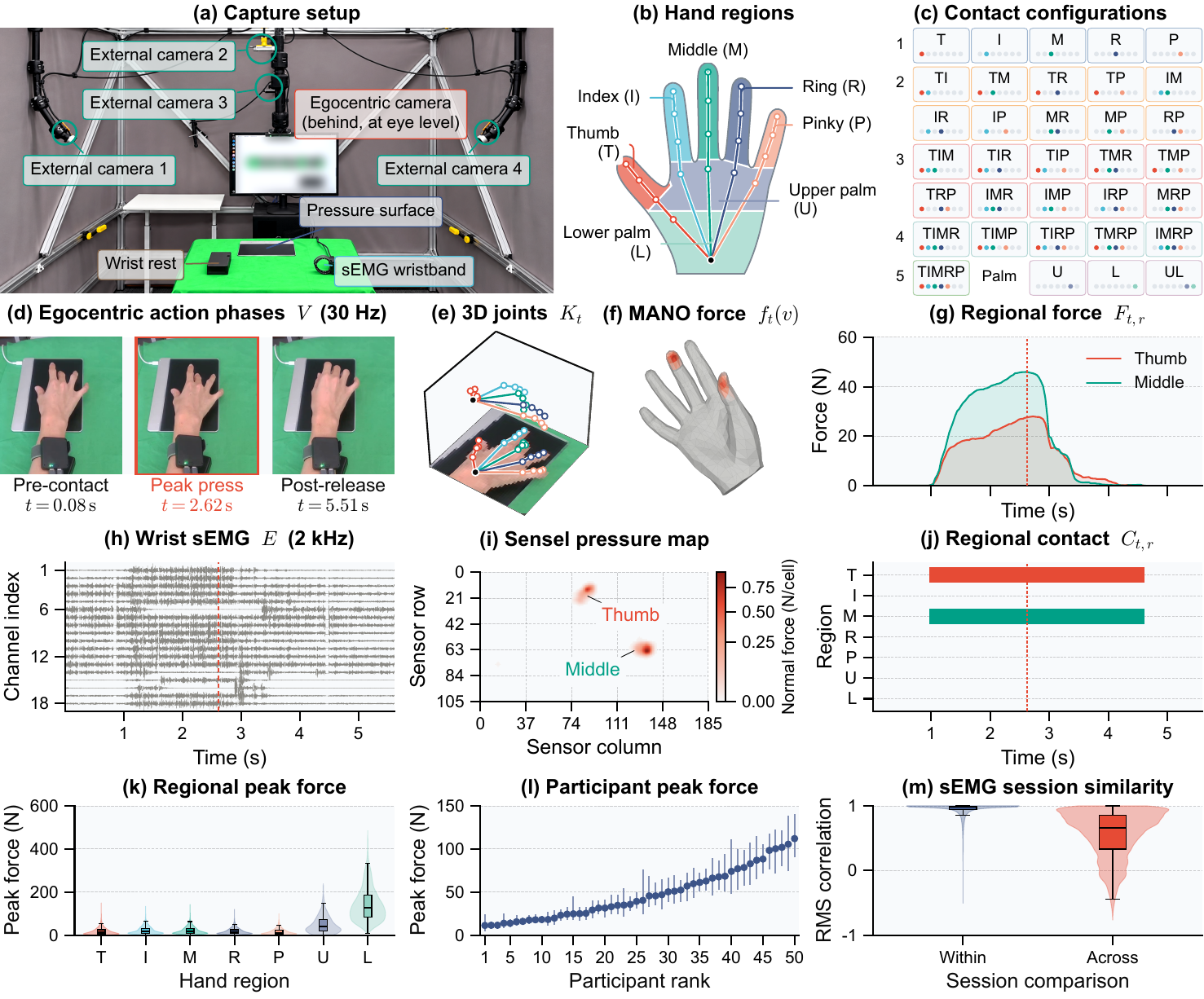}
  \caption{\textbf{EgoMyo capture, annotations, and signal statistics.} (a) Capture setup, (b) hand regions, (c) contact configurations, (d) video $V$, (e) joints $K_t$, (f) MANO force $f_t(v)$, (g) regional force $F_{t,r}$, (h) sEMG $E$, (i) Sensel force map, and (j) contact $C_{t,r}$. Panels (d)--(j) show one trial, and dashed lines mark the highlighted frame. For available valid trials lasting at least 3 s, (k) shows active-region peak force and (l) participant medians and interquartile ranges (IQRs) of peak total force. (m) Within- and across-session sEMG root mean square (RMS) correlations for matched participants, hands, and configurations, where across-session pairs come from the two recordings of the same hand.}
  \label{fig:dataset:acquisition}
\end{figure}

\myPara{Physical Hand State Estimation}\label{sec:benchmark:task}
Contact and force both arise on the hand surface, whose shape is itself determined by the kinematics, so we represent the entire physical hand state on this surface. We use the MANO~\cite{romero2017mano} mesh $M_t=\mathcal{M}(\theta_t,\beta_t)$, where $\theta_t$ denotes the hand pose, including global rotation and translation, and $\beta_t$ the hand shape. Each mesh vertex $v$ carries the normal force $f_t(v)\geq0$ attributed to it and a contact label $c_t(v)\in\{0,1\}$ (\cref{sec:benchmark:annotation}), so contact and force are two views of a single surface field, and the physical hand state is $S_t=\{M_t,c_t,f_t\}$. Given egocentric video $V$ and wrist sEMG $E$ in trailing windows ending at time $t$, a model $\Phi$ estimates
\begin{equation}
  \hat{S}_t=\Phi\!\left(V_{t-W_V:t},\,E_{t-W_E:t}\right).
  \label{eq:perception}
\end{equation}
Joints $\hat{K}_t$ are then obtained from the estimated mesh by MANO's joint regressor. Surface fields are also aggregated over the seven regions (\cref{fig:dataset:acquisition:regions}), giving region-level force $F_{t,r}=\sum_{v\in r}f_t(v)$ and contact $C_{t,r}=\max_{v\in r}c_t(v)$.

\myPara{Capture Design}\label{sec:benchmark:capture} Participants press bare-handed on a tabletop pressure surface (\cref{fig:dataset:acquisition:setup}). Model inputs pair egocentric video from a fixed participant-side camera with wrist sEMG (\cref{fig:dataset:acquisition:video,fig:dataset:acquisition:emg}). The egocentric and four surrounding cameras support mesh reconstruction, while a Sensel Morph surface~\cite{senselmorph} provides contact and normal force. Cameras are hardware-synchronized, and cross-modal streams are timestamp-aligned in software. The protocol covers 34 digit and palm contact configurations (\cref{fig:dataset:acquisition:configs}). Participants hold each cued configuration for at least 3\,s at a self-selected intensity, allowing force to vary with nearly fixed appearance. Each hand is recorded in two sessions with independent wristband donning to capture sEMG variability. Settings appear in \cref{tab:suppl:hardware}.

\myPara{Annotation}\label{sec:benchmark:annotation} We triangulate per-view 2D joints estimated with HBR Engine~\cite{hbr2025engine} and refine their trajectories with SeqHand~\cite{ren2025prioraware}. MANO fitting combines 3D-joint and 2D-reprojection constraints, initialized by available egocentric WiLoR~\cite{potamias2024wilor} predictions, to obtain $M_t$ (\cref{fig:dataset:acquisition:joints}). Calibrated registration maps baseline-corrected pressure onto the mesh to obtain $f_t$ (\cref{fig:dataset:acquisition:pressure,fig:dataset:acquisition:surface_force}). Force is converted to newtons by the sensor SDK without independent calibration against reference loads, so values are comparable across methods but carry the SDK conversion error. Regional aggregation and release-aware labeling yield force and contact annotations (\cref{fig:dataset:acquisition:regional_force,fig:dataset:acquisition:contact}). Reprojection checks and contact--cue consistency guide quality control, which excludes unassigned cells, flags uncertain labels, and re-records invalid trials. \Cref{sec:suppl:annotation} details the settings, labeling rules, and acceptance criteria.

\myPara{Statistics and Comparison}\label{sec:benchmark:stats}
EgoMyo comprises 50 participants (100 hands), with 340 planned trials per hand and 33{,}989 trials with 5.75M video frames (53.3 hours). One additional participant was recorded but is excluded, because its uncalibrated peak forces lie beyond the 1.5 interquartile range fence of participant medians, leaving sensor bias and exerted effort inseparable. Participant demographics, recruitment, and ethics appear in \cref{sec:suppl:configuration}. Force varies within each configuration (median peak force coefficient of variation 0.23) and across participants (median peak force interquartile range 23--67\,N, \cref{fig:dataset:acquisition:force_participants}), so contact alone cannot determine force. EgoMyo (\cref{tab:dataset_comparison}) keeps the hand bare unlike glove-based tactile datasets~\cite{song2025opentouch,zhou2026touchanything,zeng2026egotactile}, adds wrist sEMG and systematic configuration coverage to visual pressure datasets~\cite{grady2022pressurevision,zhao2025egopressure}, and adds contact and force supervision that sEMG datasets~\cite{salter2024emg2pose,xi2026egoemg}, which record muscle activity alone, cannot provide on the hand surface.

\myPara{Evaluation Protocol}\label{sec:benchmark:eval}
We evaluate \cref{eq:perception} under three protocols on the same 100 hands, assigning complete trials so TRAIN and TEST share neither trials nor input windows. \emph{Cross-trial} holds out repetitions within each participant, hand, session, and configuration. \emph{Cross-session} trains on the first donning and tests on the second, and \emph{cross-subject} holds out 15 of the 50 participants. Main comparisons and ablations report TEST results, while development validation within TRAIN supports hyperparameter selection and the auxiliary window study. Kinematics use root-relative MPJPE and PVE in mm, regional contact uses macro-F1, and regional force uses MAE in N. Vertex-field methods also report surface contact F1 and contact-vertex force MAE. See \cref{sec:suppl:splits,sec:suppl:metrics} for definitions. Normalization statistics are fitted on TRAIN only.

\section{When and Where Does sEMG Help?}
\label{sec:empirical}
\def\fdViolTau{0.1}
\def\fdParticipants{50}
\def\fdVisMpjpe{7.39}
\def\fdVisFone{0.843}
\def\fdVisMag{6.35}
\def\fdVisTotal{8.76}
\def\fdVisOracle{7.31}
\def\fdVisRegional{1.65}
\def\fdVisViol{18.7}
\def\fdVisUngatedViol{91.5}
\def\fdVisUngatedTotal{46.2}
\def\fdEmgMpjpe{12.40}
\def\fdEmgFone{0.706}
\def\fdEmgMag{5.33}
\def\fdEmgTotal{8.47}
\def\fdEmgOracle{5.62}
\def\fdEmgRegional{1.91}
\def\fdEmgViol{26.8}
\def\fdEmgUngatedViol{99.3}
\def\fdEmgUngatedTotal{78.7}
\def\fdFullMpjpe{6.89}
\def\fdFullFone{0.892}
\def\fdFullMag{4.28}
\def\fdFullTotal{5.25}
\def\fdFullOracle{4.48}
\def\fdFullRegional{1.08}
\def\fdFullViol{12.7}
\def\fdFullUngatedViol{88.7}
\def\fdFullUngatedTotal{45.7}
\def\fdMisMpjpe{7.54}
\def\fdMisFone{0.853}
\def\fdMisMag{6.42}
\def\fdMisTotal{8.71}
\def\fdMisOracle{7.41}
\def\fdMisRegional{1.65}
\def\fdMisViol{16.0}
\def\fdMisUngatedViol{91.8}
\def\fdMisUngatedTotal{43.8}
\def\fdDtwoMpjpe{7.27}
\def\fdDtwoFone{0.890}
\def\fdDtwoMag{4.29}
\def\fdDtwoTotal{5.29}
\def\fdDtwoOracle{4.49}
\def\fdDtwoRegional{1.10}
\def\fdDtwoViol{13.0}
\def\fdDtwoUngatedViol{89.1}
\def\fdDtwoUngatedTotal{46.7}
\def\fdZeroMag{16.84}
\def\fdZeroTotal{21.37}
\def\fdMedianMag{11.91}
\def\fdMedianTotal{14.70}
\def\fdEvVPoseDiff{5.01}
\def\fdEvVPoseLo{4.61}
\def\fdEvVPoseHi{5.40}
\def\fdEvVFoneDiff{13.7}
\def\fdEvVFoneLo{12.2}
\def\fdEvVFoneHi{15.4}
\def\fdEvVMagDiff{1.02}
\def\fdEvVMagLo{0.73}
\def\fdEvVMagHi{1.33}
\def\fdFvVMagDiff{2.07}
\def\fdFvVMagLo{1.72}
\def\fdFvVMagHi{2.43}
\def\fdFvVFoneDiff{4.9}
\def\fdFvVFoneLo{4.5}
\def\fdFvVFoneHi{5.3}
\def\fdFvVTotalDiff{3.50}
\def\fdFvVTotalLo{2.99}
\def\fdFvVTotalHi{4.03}
\def\fdFvMMagDiff{2.14}
\def\fdFvMMagLo{1.79}
\def\fdFvMMagHi{2.49}
\def\fdFvMTotalDiff{3.45}
\def\fdFvMTotalLo{2.94}
\def\fdFvMTotalHi{3.99}
\def\fdDvFPoseDiff{0.38}
\def\fdDvFPoseLo{0.34}
\def\fdDvFPoseHi{0.42}
\def\fdGvUViolDiff{76.0}
\def\fdGvUViolLo{74.3}
\def\fdGvUViolHi{77.7}
\def\fdOvGTotalDiff{0.78}
\def\fdOvGTotalLo{0.67}
\def\fdOvGTotalHi{0.89}
\def\fdPHolmMax{0.0011}
\def\fdPHolmRegionMax{0.0044}
\def\fdEvVPalmLowerDiff{2.53}
\def\fdEvVPalmLowerLo{0.91}
\def\fdEvVPalmLowerHi{4.13}
\def\fdEvVPalmLowerP{0.0044}
\def\fdGainFoneT{3.2}
\def\fdVisFoneT{0.88}
\def\fdGainFoneM{6.0}
\def\fdVisFoneM{0.81}
\def\fdGainFoneR{7.6}
\def\fdVisFoneR{0.80}
\def\fdGainFoneU{1.6}
\def\fdVisFoneU{0.89}
\def\fdGainFoneL{1.2}
\def\fdVisFoneL{0.90}
\def\fdGainFoneMin{1.2}
\def\fdGainFoneMax{7.6}
\def\fdRelMagMin{20}
\def\fdRelMagMax{38}
\def\fdRelMagDigitMin{27}
\def\fdGainMagDigitMin{1.01}
\def\fdGainMagDigitMax{1.78}
\def\fdGainMagU{7.34}
\def\fdGainMagL{4.98}
\def\fdVisMagDigitMax{4.75}
\def\fdVisMagU{19.84}
\def\fdVisMagL{25.15}
\def\fdGainForceMin{1.01}
\def\fdGainForceMax{7.34}
\def\fdD2PoseCost{0.38}
\def\fdRhoFone{1.00}
\def\fdPermFone{0.0004}
\def\fdRhoForce{0.93}
\def\fdPermForce{0.007}
\def\fdRhoMagAbs{0.93}
\def\fdPermMagAbs{0.007}
\def\fdRhoMagRel{0.18}
\def\fdPermMagRel{0.71}

\begin{figure}[t]
  \centering
  \begin{subcaptiongroup}%
    \phantomsubcaption\label{fig:empirical:pose}%
    \phantomsubcaption\label{fig:empirical:force}%
    \phantomsubcaption\label{fig:empirical:contact_gain}\label{fig:region_gain}%
    \phantomsubcaption\label{fig:empirical:force_gain}%
    \phantomsubcaption\label{fig:empirical:support}%
    \phantomsubcaption\label{fig:empirical:consistency}%
  \end{subcaptiongroup}%
  \includegraphics[width=\linewidth]{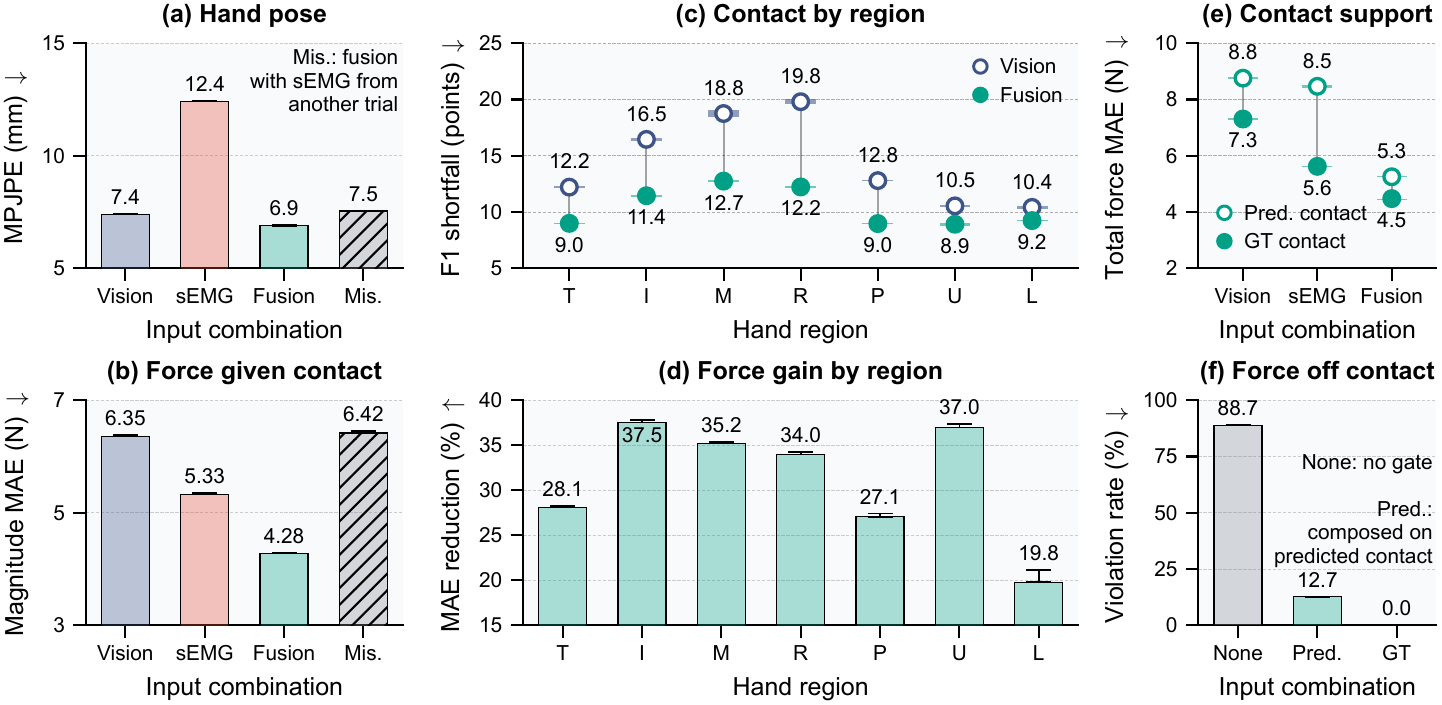}
  \caption{\textbf{Granularity, regional complementarity, and contact--force coupling.}
  (a, b) Pose error and force magnitude error on contacting regions per input, where Mismatch pairs fusion with sEMG from another trial.
  (c, d) Contact error and relative magnitude reduction per region, vision against fusion.
  (e, f) Force composed on predicted or ground-truth contact, and frames with force on non-contacting regions. Cross-trial TEST, bars are means over three seeds and upper whiskers their standard error.}
  \label{fig:empirical}
\end{figure}

Wrist sEMG records the superimposed activity of several forearm muscles and reflects overall hand exertion rather than the digit producing it, which has confined prior methods to aggregate force estimation~\cite{seo2024posture,he2026forceband}. Contact and force are resolved per hand region, so a hand-level signal has to be tested on whether it can help recover a spatially resolved state. We examine that question at three levels (\cref{fig:empirical}), \emph{what} each modality resolves in isolation, \emph{where} sEMG adds to vision, and \emph{how} contact and force constrain one another. All variants share one neutral baseline, with split, schedule, and supervision held fixed (\cref{sec:suppl:analysis:settings}).

\phantomsection\label{sec:empirical:finding1}\myPara{Finding 1: The modalities resolve the state at different granularities} Alone, sEMG localizes the hand poorly, reaching \fdEmgMpjpe\,mm MPJPE against \fdVisMpjpe\,mm for vision (\cref{fig:empirical:pose}), and separates contacting from non-contacting regions less well, at \fdEmgFone\ against \fdVisFone\ contact F1 (\cref{tab:suppl:findings}). On contacting regions, however, it estimates the force magnitude more accurately (\fdEmgMag\ vs.\ \fdVisMag\,N, \cref{fig:empirical:force}), on every region except the lower palm. Given ground-truth contact, its total force error falls to \fdEmgOracle\,N against \fdVisOracle\,N for vision (\cref{fig:empirical:support}), so sEMG is limited by where the hand touches rather than by how strongly it presses. Each stream thus supplies what the other lacks, which favors encoding vision and sEMG each in a structure of its own over forcing both through shared parameters.

\phantomsection\label{sec:empirical:finding2}\myPara{Finding 2: Complementarity is regional, not global} Adding sEMG to vision improves contact F1 by \fdGainFoneMin\ to \fdGainFoneMax\ points and reduces magnitude MAE by \fdGainForceMin\ to \fdGainForceMax\,N, but unevenly across the seven regions (\cref{fig:region_gain,fig:empirical:force_gain,tab:suppl:findings}). Absolute gains and vision's error move together, at a rank correlation of $\rho=\fdRhoFone$ for contact and $\fdRhoForce$ for force (exact permutation $p=\fdPermFone$ and $\fdPermForce$). sEMG therefore helps contact most where vision is least reliable, on the middle and ring fingers, and least where vision is accurate, on the thumb and the lower palm, so fusion narrows the spread of error across regions rather than shifting it. Zeroing sEMG in the pose branch alone, with the contact and force path unchanged, raises pose error by \fdD2PoseCost\,mm, so the stream also carries information vision does not already resolve from image appearance alone.

\phantomsection\label{sec:empirical:finding3}\myPara{Finding 3: The components are coupled through the contact support} Force depends on knowing where the hand touches. Composing the predicted magnitude of the fused model on ground-truth instead of predicted contact lowers its total force error from \fdFullTotal\ to \fdFullOracle\,N (\cref{fig:empirical:support}). Supervised only where contact occurs, the raw magnitude places more than \fdViolTau\,N on a non-contacting region in \fdFullUngatedViol\% of frames (\cref{fig:empirical:consistency}). Restricting it to the predicted contact support lowers this to \fdFullViol\%, a drop of \fdGvUViolDiff\ points, and every remaining violation is a contact false positive. The same restriction acts on the single-stream models, from \fdVisUngatedViol\% to \fdVisViol\% for vision and from \fdEmgUngatedViol\% to \fdEmgViol\% for sEMG, so the force estimate is bounded by the support it is decoded on, not by the magnitude of the prediction.

\section{MyoBridge: From Muscles to Surface}
\label{sec:method}

\begin{figure*}[t]
  \centering
  \includegraphics[width=\textwidth]{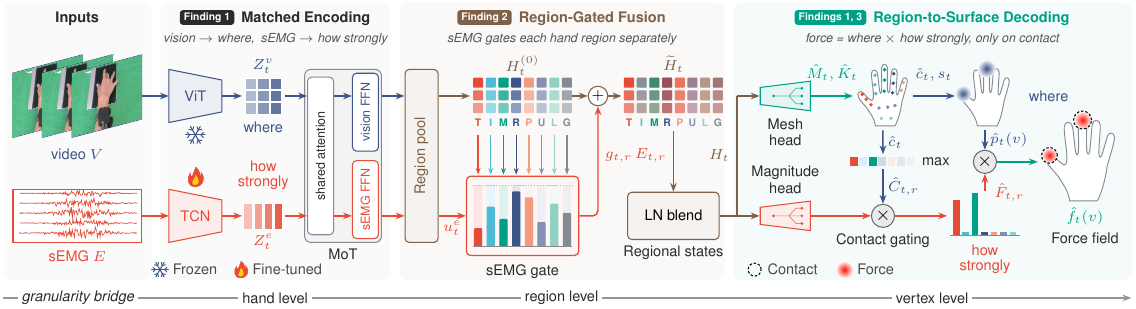}
  \caption{\textbf{Framework of MyoBridge.} Each stage answers a finding of \cref{sec:empirical}, moving from hand to region to vertex level. A frozen WiLoR encoder supplies visual tokens, a fine-tuned TCN encodes sEMG, and MoT shares attention with stream-specific FFNs (Finding~\hyperref[sec:empirical:finding1]{1}). Region states gate the sEMG residual per region (Finding~\hyperref[sec:empirical:finding2]{2}). The decoder predicts the mesh, and the force field distributes contact-gated regional forces over predicted contact vertices (Findings~\hyperref[sec:empirical:finding1]{1} and~\hyperref[sec:empirical:finding3]{3}). Blue marks where, red how strongly.}
  \label{fig:framework}
\end{figure*}

\cref{sec:empirical} constrains how the two streams should meet. sEMG resolves how strongly the hand presses but not where (Finding~\hyperref[sec:empirical:finding1]{1}), its benefit differs across hand regions (Finding~\hyperref[sec:empirical:finding2]{2}), and force is only as reliable as its contact support (Finding~\hyperref[sec:empirical:finding3]{3}). MyoBridge therefore bridges hand-level sEMG and the vertex-level surface state through region states (\cref{fig:framework}). Unlike prior vision--sEMG models, it lets sEMG act on each region through its own gate and confines the resulting force to the predicted contact support. From \cref{eq:perception}, it predicts the MANO pose and shape $(\hat{\theta}_t,\hat{\beta}_t)$, per-vertex contact $\hat{c}_t(v)$, and force $\hat{f}_t(v)$, with joints and regional values derived as in \cref{sec:benchmark:task}, causally at time $t$ (\cref{alg:egomyo}).

\subsection{Granularity-Matched Encoding (GME)}
\label{sec:modality_encoding}

The streams carry evidence of different kinds (Finding~\hyperref[sec:empirical:finding1]{1}), so each is encoded in a structure that preserves its own. The granularity gap measured there sets this separation rather than any architectural default. Following WiLoR~\cite{potamias2024wilor}, a frozen visual encoder and a linear projection map the video window to spatial patch tokens $Z^{v}_t$ that retain both hand geometry and appearance. A temporal encoder maps the 18-channel sEMG window to multichannel tokens $Z^{e}_t$ that preserve the activation dynamics of each channel. The sEMG encoder is pretrained on training windows only, by aligning them with frozen visual features of the synchronized frames through a symmetric InfoNCE objective at a temperature of 0.07, with in-batch negatives and same-trial windows excluded from each other's negatives. Within this encoder, a causal convolutional stem $f_{\mathrm{stem}}$ maps the window $E_{t-W_E:t}$ to temporal features $X^e_t$, residual temporal convolutions refine them before averaging, and channel embeddings $B_{\mathrm{ch}}$ keep the 18 electrodes distinct,
\begin{equation}
Z^e_t=\mathrm{LN}\big(\mathrm{Mean}_{\tau}\big[X^e_t+f_{\mathrm{TCN}}(X^e_t)\big]+B_{\mathrm{ch}}\big).
\label{eq:emg_adapter}
\end{equation}
The two token sets are processed jointly by a Mixture-of-Transformers (MoT) backbone~\cite{liang2025mixture}, which shares attention across all tokens so that each stream can draw on the other, while each modality keeps its own feed-forward expert instead of parameters fitted to the other kind of evidence.

\subsection{Region-Gated Fusion (RGF)}
\label{sec:region_fusion}

sEMG helps most where vision alone is least reliable (Finding~\hyperref[sec:empirical:finding2]{2}). Prior vision--sEMG models fuse the streams for the whole hand, through one multimodal residual~\cite{xi2026egoemg} or shared attention over all tokens~\cite{liang2025mixture,wang2025hbridge}, and part-aware hand estimators~\cite{zheng2021sar,ren2023poseguided} reason over parts within one modality. MyoBridge instead makes the hand region the unit of fusion. After MoT, let $\bar{Z}^{v}_t$ and $\bar{Z}^{e}_t$ denote the two output streams. We average each stream over its tokens, concatenate the summaries, and map them through an MLP $f_{\mathrm{pool}}$ to seven regional states for $\{T,I,M,R,P,U,L\}$ and one global state for the whole hand,
\begin{equation}
\begin{aligned}
u^v_t&=\operatorname{Mean}(\bar{Z}^{v}_t),\quad
u^e_t=\operatorname{Mean}(\bar{Z}^{e}_t),\\
H^{(0)}_t&=\operatorname{reshape}_{8\times d}\!\big(f_{\mathrm{pool}}([u^v_t;u^e_t])\big).
\end{aligned}
\label{eq:vamr}
\end{equation}
A linear projection then produces one sEMG residual per state, $E_t=\operatorname{reshape}_{8\times d}(W_eu^e_t+b_e)$, and each state predicts its own scalar gate from its content,
\begin{equation}
g_{t,r}=\sigma(f_{\mathrm{gate}}(h^{(0)}_{t,r})),\qquad
\widetilde{h}_{t,r}=h^{(0)}_{t,r}+g_{t,r}E_{t,r}.
\label{eq:vamr_route}
\end{equation}
Conditioning each gate on the state it updates lets every region admit its own share of the residual, as Finding~\hyperref[sec:empirical:finding2]{2} calls for, rather than one share for the hand. MoT already mixes the streams, so the gates control only this additional residual. The global state is updated likewise, and each state is blended with its layer-normalized value,
\begin{equation}
h_{t,r}=0.9\,\widetilde{h}_{t,r}+0.1\operatorname{LN}(\widetilde{h}_{t,r}),
\label{eq:state_normalization}
\end{equation}
which mildly pulls the gated states toward a common scale. The resulting $H_t$ drives both the mesh head and the force magnitude, so sEMG reaches kinematics as well as force, in line with its pose cost in Finding~\hyperref[sec:empirical:finding2]{2}.

\subsection{Region-to-Surface Decoding (RSD)}
\label{sec:structured_decoding}

Following Findings~\hyperref[sec:empirical:finding1]{1} and~\hyperref[sec:empirical:finding3]{3}, the decoder lifts the regional states to the mesh surface and composes force from an sEMG-informed regional magnitude and a within-region distribution on the predicted contact support.

\myPara{Mesh}\label{sec:method:kinematics}
A kinematic head regresses $(\hat{\theta}_t,\hat{\beta}_t)=D_M(H_t)$ from the region and global states, which yields the mesh $\hat{M}_t=\mathcal{M}(\hat{\theta}_t,\hat{\beta}_t)$ and, through MANO's fixed joint regressor, the joints $\hat{K}_t$, while the predicted mesh anchors the vertex queries of the lifting stage described next.

\myPara{Vertex Lifting}\label{sec:surface_lifting}
Each vertex $v$ forms a query from a learnable embedding $e_v$, the state of its region $r(v)$, and its position on the mesh predicted in the previous step,
\begin{equation}
q_{t,v}=e_v+h_{t,r(v)}+\operatorname{PE}\big(\hat{M}_t(v)\big),
\label{eq:vertex_query}
\end{equation}
and attends over $[Z^{v}_t;Z^{e}_t;H_t]$ to obtain a vertex feature $z_{t,v}$. Transformer mesh regressors build vertex queries from template positions and image features~\cite{lin2021posemesh}. The region state lets each vertex read both streams through its own region rather than through a hand-level sEMG vector.

\myPara{Contact-Supported Force Field}\label{sec:method:force}
Visual estimators regress pressure as a dense map from appearance~\cite{grady2022pressurevision,zhao2025egopressure}, drawing where and how strongly from the same evidence, while sEMG-based estimators~\cite{seo2024posture,he2026forceband} predict force for predefined hand areas rather than over the surface. Yet sEMG resolves how strongly a region presses but not where (Finding~\hyperref[sec:empirical:finding1]{1}), and unconstrained magnitudes leak onto non-contacting regions (Finding~\hyperref[sec:empirical:finding3]{3}). The Contact-Supported Force Field (CSFF) reads the magnitude from the sEMG-informed state of \cref{eq:state_normalization} and scales it by regional contact,
\begin{equation}
\hat{F}_{t,r}=\hat{C}_{t,r}\,\operatorname{softplus}\big(D_m(h_{t,r})\big),
\label{eq:csff_magnitude}
\end{equation}
where $\hat{c}_t(v)=\sigma(D_c(z_{t,v}))$ is the soft contact probability. Training uses $\tilde{c}_t(v)=\hat{c}_t(v)$, while inference uses $\tilde{c}_t(v)=\hat{c}_t(v)\mathbf{1}[\hat{c}_t(v)\geq0.5]$. Regional contact is $\hat{C}_{t,r}=\max_{v\in r}\tilde{c}_t(v)$. Scores $s_t(v)=D_s(z_{t,v})$ allocate force over the vertices of each hand region,
\begin{equation}
\hat{f}_t(v)=\hat{F}_{t,r(v)}\,\hat{p}_t(v),~
\hat{p}_t(v)=\frac{\tilde{c}_t(v)\,e^{s_t(v)}}{\sum_{u\in r(v)}\tilde{c}_t(u)\,e^{s_t(u)}}.
\label{eq:csff}
\end{equation}
An empty contact support gives zero force and distribution. At inference, vertices below the contact threshold carry exactly zero force, so leakage outside reference contact can arise only from contact false positives. During training, soft probabilities instead attenuate force continuously. Both sum vertex forces to regional force, consistent with \cref{sec:benchmark:task}.

\subsection{Training Objective}
\label{sec:training_objectives}
 
MyoBridge is trained end to end by minimizing
\begin{equation}
\mathcal{L}=\sum_{k\in\{M,c,F,p\}}\Big(\frac{1}{2\sigma_k^{2}}\mathcal{L}_k+\log\sigma_k\Big),
\label{eq:total_loss}
\end{equation}
where each $\sigma_k$ is a learned homoscedastic uncertainty~\cite{kendall2018multi} that balances the units and scales of the four terms without loss-weight hyperparameters. The mesh term $\mathcal{L}_M$ and the regional-force term $\mathcal{L}_F$ are $\ell_1$ losses on the vertices $\hat{M}_t(v)$, which also determine the joints, and on the forces $\hat{F}_{t,r}$ of \cref{eq:csff_magnitude}. The contact term is a Dice loss, which remains insensitive to the scarcity of contacting vertices on the mesh,
\begin{equation}
\mathcal{L}_c=1-\frac{2\sum_{v}\hat{c}_t(v)\,c_t(v)}{\sum_{v}\hat{c}_t(v)+\sum_{v}c_t(v)},
\label{eq:contact_loss}
\end{equation}
and the distribution term compares $\hat{p}_t$ of \cref{eq:csff} with the ground-truth distribution $p_t(v)=f_t(v)/F_{t,r(v)}$ on every region that carries load in the frame,
\begin{equation}
\mathcal{L}_p=-\sum_{v:\,F_{t,r(v)}>0} p_t(v)\log\hat{p}_t(v),
\label{eq:distribution_loss}
\end{equation}
so that $\mathcal{L}_F$ and $\mathcal{L}_p$ supervise how strongly and where force acts separately, mirroring the factorization in \cref{eq:csff}.

\begin{table}[t]

  \centering
  \scriptsize
  \setlength{\tabcolsep}{1.5pt}
  \renewcommand{\arraystretch}{1.08}
  \caption{\textbf{Comparison with recent strong baselines.} Mean $\pm$ SD, bold best, $V$ vision, $E$ sEMG, force in N.}
  \label{tab:main}
  \newcommand{\mainDigits}{1}
  \def\mainNumber#1.#2\relax{\makebox[\mainDigits\fontcharwd\font`0][r]{#1}.#2}
  \newcommand{\mainValue}[2]{\mbox{\mainNumber#1\relax{\fontsize{4.5}{5}\selectfont\kern0.12em\textpm\kern0.12em#2}}}
  \newcommand{\mainBestValue}[2]{\mbox{\textbf{\mainNumber#1\relax{\fontsize{4.5}{5}\selectfont\kern0.12em\textpm\kern0.12em#2}}}}
  \newcommand{\mainSingle}[1]{\mbox{\mainNumber#1\relax}}
  \newcommand{\mainBestSingle}[1]{\mbox{\textbf{\mainNumber#1\relax}}}
  {\fontsize{6.5}{7.5}\selectfont
  \resizebox{\columnwidth}{!}{%
  \begin{tabular}{@{}ll*{2}{>{\renewcommand{\mainDigits}{2}}c}*{4}{c}@{}}
    \toprule
    \multirow{2.5}{*}{\textbf{Method}} & \multirow{2.5}{*}{\textbf{Input}} & \multicolumn{2}{c}{\textbf{Kinematics (mm)}} & \multicolumn{2}{c}{\textbf{Region}} & \multicolumn{2}{c}{\textbf{Vertex}} \\
    \cmidrule(lr){3-4} \cmidrule(lr){5-6} \cmidrule(l){7-8}
    & & MPJPE$\downarrow$ & PVE$\downarrow$ & F1$\uparrow$ & MAE$\downarrow$ & F1$\uparrow$ & MAE$\downarrow$ \\
    \midrule
    \multicolumn{8}{c}{\textbf{Cross-Trial}} \\
    \midrule
    HaMeR~\cite{pavlakos2024hamer} & $V$ & \mainValue{7.34}{0.25} & \mainValue{7.30}{0.23} & \mainValue{0.89}{0.01} & \mainValue{2.02}{0.18} & \mainValue{0.81}{0.01} & \mainValue{0.38}{0.01} \\ %
    WiLoR~\cite{potamias2024wilor} & $V$ & \mainValue{6.16}{0.25} & \mainValue{6.15}{0.20} & \mainValue{0.87}{0.01} & \mainValue{1.71}{0.04} & \mainValue{0.79}{0.01} & \mainValue{0.37}{0.01} \\ %
    OpenTouch~\cite{song2025opentouch} & $V$ & \mainValue{9.21}{0.17} & \mainValue{9.11}{0.16} & \mainValue{0.84}{0.02} & \mainValue{1.97}{0.06} & \mainValue{0.74}{0.03} & \mainValue{0.40}{0.01} \\ %
    EgoTouch~\cite{zhou2026touchanything} & $V$ & \mainValue{9.42}{0.07} & \mainValue{9.32}{0.06} & \mainValue{0.84}{0.01} & \mainValue{2.05}{0.04} & \mainValue{0.72}{0.02} & \mainValue{0.41}{0.01} \\ %
    EgoTactile~\cite{zeng2026egotactile} & $V$ & \mainValue{9.23}{0.21} & \mainValue{9.13}{0.20} & \mainValue{0.85}{0.00} & \mainValue{1.99}{0.11} & \mainValue{0.75}{0.01} & \mainValue{0.40}{0.01} \\ %
    EgoPressure~\cite{zhao2025egopressure} & $V$ & \mainValue{17.34}{0.38} & \mainValue{16.91}{0.35} & \mainValue{0.59}{0.03} & \mainValue{2.97}{0.04} & \mainValue{0.44}{0.02} & \mainValue{0.49}{0.00} \\ %
    NeuroPose~\cite{liu2021neuropose} & $E$ & \mainValue{20.93}{0.38} & \mainValue{20.34}{0.37} & \mainValue{0.16}{0.15} & \mainValue{3.11}{0.01} & \mainValue{0.14}{0.13} & \mainValue{0.47}{0.02} \\ %
    vemg2pose~\cite{salter2024emg2pose} & $E$ & \mainValue{15.09}{0.10} & \mainValue{17.83}{0.59} & \mainValue{0.70}{0.01} & \mainValue{2.31}{0.02} & \mainValue{0.64}{0.01} & \mainValue{0.35}{0.00} \\ %
    EMGFormer~\cite{xi2026egoemg} & $V, E$ & \mainValue{9.00}{0.03} & \mainValue{9.21}{0.03} & \mainValue{0.85}{0.00} & \mainValue{1.74}{0.00} & \mainValue{0.77}{0.00} & \mainValue{0.34}{0.00} \\ %
    MoT~\cite{liang2025mixture} & $V, E$ & \mainValue{6.51}{0.07} & \mainValue{6.43}{0.07} & \mainValue{0.92}{0.00} & \mainValue{1.41}{0.16} & \mainValue{0.84}{0.00} & \mainValue{0.31}{0.00} \\ %
    HBridge~\cite{wang2025hbridge} & $V, E$ & \mainValue{5.42}{0.04} & \mainValue{5.36}{0.04} & \mainValue{0.94}{0.00} & \mainValue{1.06}{0.01} & \mainValue{0.87}{0.00} & \mainBestValue{0.28}{0.00} \\ %
    \textbf{MyoBridge} & $V, E$ & \mainBestValue{3.26}{0.03} & \mainBestValue{3.30}{0.03} & \mainBestValue{0.95}{0.00} & \mainBestValue{0.86}{0.01} & \mainBestValue{0.88}{0.00} & \mainValue{0.29}{0.00} \\ %
    \midrule
    \multicolumn{8}{c}{\textbf{Cross-Session}} \\
    \midrule
    HaMeR~\cite{pavlakos2024hamer} & $V$ & \mainValue{8.92}{1.98} & \mainValue{8.82}{1.94} & \mainValue{0.80}{0.06} & \mainValue{2.35}{0.42} & \mainValue{0.71}{0.06} & \mainValue{0.43}{0.02} \\ %
    WiLoR~\cite{potamias2024wilor} & $V$ & \mainValue{7.09}{1.17} & \mainValue{7.03}{1.13} & \mainValue{0.83}{0.04} & \mainValue{2.05}{0.20} & \mainValue{0.75}{0.02} & \mainValue{0.42}{0.02} \\ %
    OpenTouch~\cite{song2025opentouch} & $V$ & \mainValue{12.51}{0.20} & \mainValue{12.31}{0.19} & \mainValue{0.68}{0.01} & \mainValue{2.85}{0.07} & \mainValue{0.59}{0.00} & \mainValue{0.46}{0.00} \\ %
    EgoTouch~\cite{zhou2026touchanything} & $V$ & \mainValue{12.75}{0.02} & \mainValue{12.54}{0.03} & \mainValue{0.68}{0.01} & \mainValue{2.90}{0.16} & \mainValue{0.59}{0.02} & \mainValue{0.47}{0.01} \\ %
    EgoTactile~\cite{zeng2026egotactile} & $V$ & \mainValue{12.72}{0.06} & \mainValue{12.50}{0.07} & \mainValue{0.68}{0.01} & \mainValue{2.86}{0.13} & \mainValue{0.59}{0.02} & \mainValue{0.46}{0.01} \\ %
    EgoPressure~\cite{zhao2025egopressure} & $V$ & \mainValue{19.07}{0.05} & \mainValue{18.65}{0.05} & \mainValue{0.58}{0.02} & \mainValue{3.12}{0.05} & \mainValue{0.50}{0.01} & \mainValue{0.48}{0.00} \\ %
    NeuroPose~\cite{liu2021neuropose} & $E$ & \mainValue{22.31}{1.40} & \mainValue{21.67}{1.33} & \mainValue{0.14}{0.09} & \mainValue{3.11}{0.01} & \mainValue{0.13}{0.09} & \mainValue{0.49}{0.02} \\ %
    vemg2pose~\cite{salter2024emg2pose} & $E$ & \mainValue{22.90}{1.25} & \mainValue{22.13}{0.85} & \mainValue{0.38}{0.00} & \mainValue{3.44}{0.05} & \mainValue{0.35}{0.00} & \mainValue{0.44}{0.00} \\ %
    EMGFormer~\cite{xi2026egoemg} & $V, E$ & \mainValue{14.56}{0.11} & \mainValue{14.38}{0.08} & \mainValue{0.61}{0.00} & \mainValue{2.80}{0.01} & \mainValue{0.52}{0.00} & \mainValue{0.44}{0.00} \\ %
    MoT~\cite{liang2025mixture} & $V, E$ & \mainValue{10.69}{0.20} & \mainValue{10.49}{0.20} & \mainValue{0.71}{0.00} & \mainValue{2.46}{0.11} & \mainValue{0.62}{0.00} & \mainValue{0.40}{0.00} \\ %
    HBridge~\cite{wang2025hbridge} & $V, E$ & \mainValue{10.79}{0.09} & \mainValue{10.57}{0.07} & \mainValue{0.71}{0.00} & \mainValue{2.29}{0.01} & \mainValue{0.60}{0.00} & \mainValue{0.40}{0.00} \\ %
    \textbf{MyoBridge} & $V, E$ & \mainBestValue{5.35}{0.02} & \mainBestValue{5.33}{0.02} & \mainBestValue{0.87}{0.00} & \mainBestValue{1.59}{0.02} & \mainBestValue{0.78}{0.00} & \mainBestValue{0.38}{0.00} \\ %
    \midrule
    \multicolumn{8}{c}{\textbf{Cross-Subject}} \\
    \midrule
    HaMeR~\cite{pavlakos2024hamer} & $V$ & \mainValue{13.37}{1.97} & \mainValue{13.21}{1.91} & \mainValue{0.71}{0.06} & \mainValue{3.18}{0.27} & \mainValue{0.63}{0.06} & \mainValue{0.50}{0.03} \\ %
    WiLoR~\cite{potamias2024wilor} & $V$ & \mainValue{10.15}{2.26} & \mainValue{10.07}{2.20} & \mainValue{0.80}{0.08} & \mainValue{3.14}{0.56} & \mainValue{0.72}{0.06} & \mainValue{0.53}{0.05} \\ %
    OpenTouch~\cite{song2025opentouch} & $V$ & \mainValue{16.24}{0.56} & \mainValue{16.02}{0.55} & \mainValue{0.60}{0.03} & \mainValue{4.11}{0.47} & \mainValue{0.53}{0.02} & \mainValue{0.56}{0.02} \\ %
    EgoTouch~\cite{zhou2026touchanything} & $V$ & \mainValue{16.07}{0.54} & \mainValue{15.86}{0.52} & \mainValue{0.60}{0.04} & \mainValue{3.95}{0.14} & \mainValue{0.53}{0.02} & \mainValue{0.56}{0.01} \\ %
    EgoTactile~\cite{zeng2026egotactile} & $V$ & \mainValue{15.83}{0.53} & \mainValue{15.62}{0.44} & \mainValue{0.60}{0.01} & \mainValue{3.81}{0.38} & \mainValue{0.52}{0.00} & \mainValue{0.56}{0.05} \\ %
    EgoPressure~\cite{zhao2025egopressure} & $V$ & \mainValue{20.61}{0.43} & \mainValue{20.19}{0.35} & \mainValue{0.54}{0.00} & \mainValue{3.70}{0.17} & \mainValue{0.46}{0.01} & \mainValue{0.53}{0.02} \\ %
    NeuroPose~\cite{liu2021neuropose} & $E$ & \mainValue{23.66}{0.19} & \mainValue{23.11}{0.18} & \mainValue{0.19}{0.04} & \mainValue{3.24}{0.01} & \mainValue{0.15}{0.04} & \mainValue{0.46}{0.03} \\ %
    vemg2pose~\cite{salter2024emg2pose} & $E$ & \mainValue{25.03}{0.05} & \mainValue{24.01}{0.26} & \mainValue{0.34}{0.01} & \mainValue{3.74}{0.12} & \mainValue{0.31}{0.00} & \mainValue{0.45}{0.00} \\ %
    EMGFormer~\cite{xi2026egoemg} & $V, E$ & \mainValue{19.47}{0.11} & \mainValue{18.98}{0.09} & \mainValue{0.54}{0.00} & \mainValue{3.36}{0.02} & \mainValue{0.46}{0.00} & \mainBestValue{0.43}{0.00} \\ %
    MoT~\cite{liang2025mixture} & $V, E$ & \mainValue{14.41}{0.05} & \mainValue{14.17}{0.04} & \mainValue{0.63}{0.00} & \mainValue{3.26}{0.13} & \mainValue{0.55}{0.00} & \mainValue{0.46}{0.01} \\ %
    HBridge~\cite{wang2025hbridge} & $V, E$ & \mainValue{14.47}{0.15} & \mainValue{14.23}{0.15} & \mainValue{0.62}{0.01} & \mainValue{3.08}{0.02} & \mainValue{0.53}{0.01} & \mainValue{0.46}{0.00} \\ %
    \textbf{MyoBridge} & $V, E$ & \mainBestValue{7.84}{0.10} & \mainBestValue{7.81}{0.10} & \mainBestValue{0.83}{0.01} & \mainBestValue{2.44}{0.04} & \mainBestValue{0.74}{0.00} & \mainValue{0.47}{0.00} \\ %
    \bottomrule
  \end{tabular}}
  }
\end{table}

\begin{table}[t]
\captionsetup{type=table}
  \centering
  \small
  \setlength{\tabcolsep}{2pt}
  \renewcommand{\arraystretch}{1.0}
  \caption{\textbf{Computational cost}. Parameters (M), task FLOPs (G), and inference time (ms), with native RGB input starred.}
  \label{tab:compute}
  \resizebox{\columnwidth}{!}{%
  \begin{tabular}{@{}lrrrr@{\hspace{10pt}}lrrrr@{}}
    \toprule
    \textbf{Method} & \textbf{Total} & \textbf{Train} & \textbf{FLOPs} & \textbf{Infer.} & \textbf{Method} & \textbf{Total} & \textbf{Train} & \textbf{FLOPs} & \textbf{Infer.} \\
    \midrule
    HaMeR$*$ & 672.02 & 672.02 & 251.099 & 0.6397 & NeuroPose & 6.61 & 6.61 & 0.497 & 0.0359 \\
    WiLoR$*$ & 641.04 & 641.04 & 275.940 & 0.9885 & vemg2pose & 4.47 & 4.47 & 0.316 & 0.0074 \\
    OpenTouch & 89.74 & 3.15 & 0.017 & 0.0021 & EMGFormer & 92.13 & 5.55 & 0.404 & 0.0131 \\
    EgoTouch & 89.99 & 3.40 & 0.016 & 0.0029 & MoT & 97.56 & 10.98 & 0.514 & 0.0106 \\
    EgoTactile & 90.53 & 3.94 & 0.017 & 0.0030 & HBridge & 95.90 & 9.31 & 0.558 & 0.0122 \\
    EgoPressure & 89.67 & 3.08 & 0.006 & 0.0008 & \textbf{MyoBridge} & 648.69 & 9.41 & 0.712 & 0.0284 \\
    \bottomrule
  \end{tabular}}
\vspace{\floatsep}
  \centering
  \captionsetup{type=figure}
  \begin{subcaptiongroup}
    \phantomsubcaption\label{fig:hparam:kinematics}
    \phantomsubcaption\label{fig:hparam:contact}
    \phantomsubcaption\label{fig:hparam:force}
    \phantomsubcaption\label{fig:hparam:timing:kinematics}
    \phantomsubcaption\label{fig:hparam:timing:contact}
    \phantomsubcaption\label{fig:hparam:timing:force}
  \end{subcaptiongroup}
  \includegraphics[width=\linewidth]{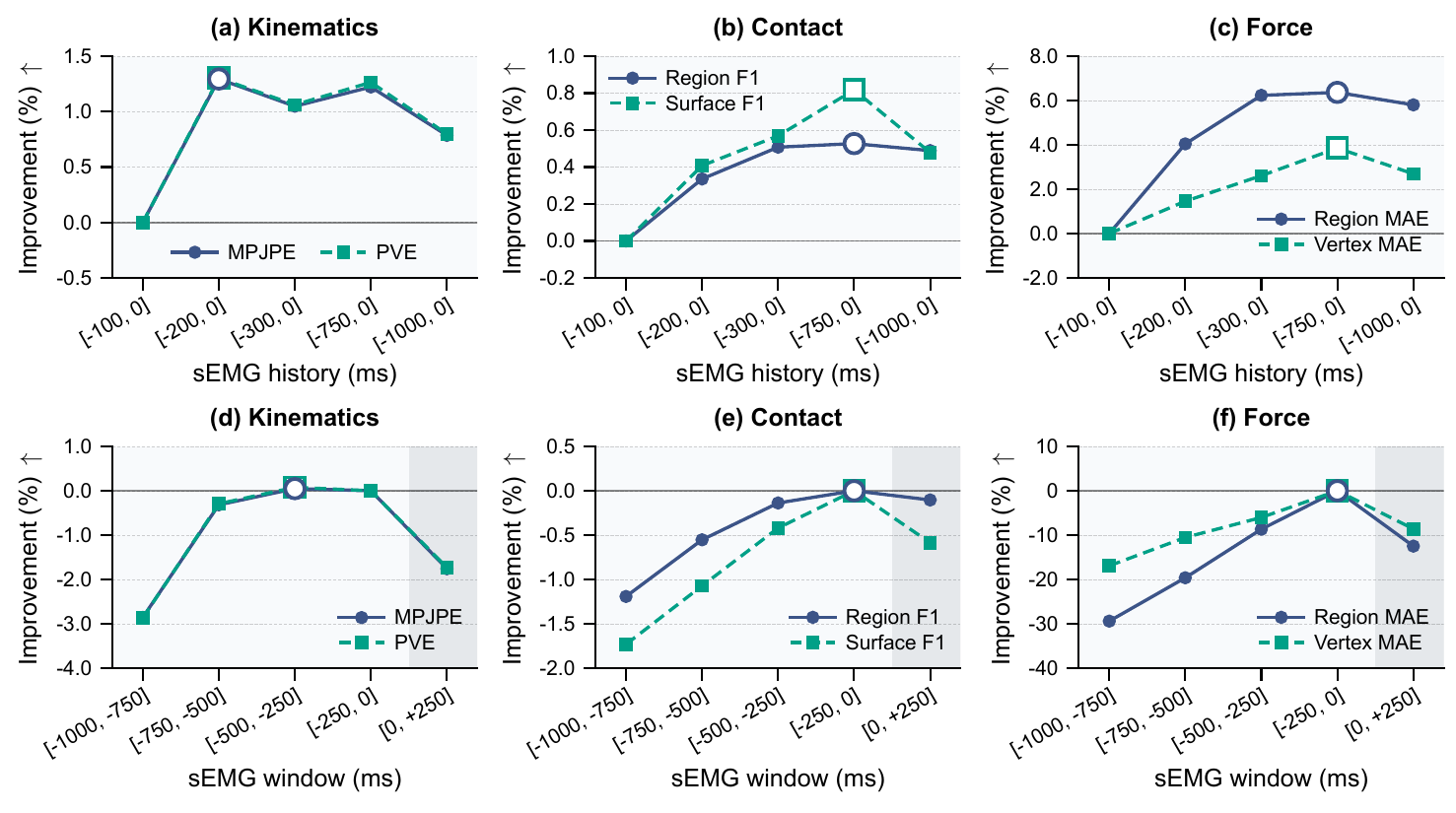}
  \caption{\textbf{How long and when to read wrist sEMG.} Improvement relative to the shortest history (a--c) and to the [\textminus250,~0]\,ms window (d--f), on cross-trial development validation. Hollow markers mark the best interval of each curve, and gray marks future sEMG, which serves only as an offline control.}
  \label{fig:hparam}
\end{table}

\begin{table}[t]

  \captionsetup{type=table}
  \centering
  \scriptsize
  \setlength{\tabcolsep}{1.5pt}
  \renewcommand{\arraystretch}{1.08}
  \newcommand{\ablationDigits}{1} 
  \def\ablationNumber#1.#2\relax{\makebox[\ablationDigits\fontcharwd\font`0][r]{#1}.#2}
  \newcommand{\ablationValue}[2]{\mbox{\ablationNumber#1\relax{\fontsize{4.5}{5}\selectfont\kern0.12em\textpm\kern0.12em#2}}}
  \newcommand{\ablationSingle}[1]{\mbox{\ablationNumber#1\relax}}
  \newcommand{\ablationBestSingle}[1]{\mbox{\textbf{\ablationNumber#1\relax}}}
  \newcommand{\ablationBestValue}[2]{\mbox{\textbf{\ablationNumber#1\relax{\fontsize{4.5}{5}\selectfont\kern0.12em\textpm\kern0.12em#2}}}}
  \caption{\textbf{Cross-trial ablations.} Best are in bold, as in \cref{tab:main}.}
  \label{tab:ablation}
  {\fontsize{6.5}{7.5}\selectfont
  \resizebox{\columnwidth}{!}{%
  \begin{tabular}{@{}l*{2}{>{\renewcommand{\ablationDigits}{2}}c}*{4}{c}@{}}
    \toprule
    \multirow{2.5}{*}{\textbf{Variant}} & \multicolumn{2}{c}{\textbf{Kinematics (mm)}} & \multicolumn{2}{c}{\textbf{Region}} & \multicolumn{2}{c}{\textbf{Vertex}} \\
    \cmidrule(lr){2-3} \cmidrule(lr){4-5} \cmidrule(l){6-7}
    & MPJPE$\downarrow$ & PVE$\downarrow$ & F1$\uparrow$ & MAE$\downarrow$ & F1$\uparrow$ & MAE$\downarrow$ \\
    \midrule
    \textbf{MyoBridge} & \ablationBestValue{3.26}{0.03} & \ablationBestValue{3.30}{0.03} & \ablationBestValue{0.95}{0.00} & \ablationBestValue{0.86}{0.01} & \ablationBestValue{0.88}{0.00} & \ablationBestValue{0.29}{0.00} \\ %
    \midrule
    w/o sEMG & \ablationValue{5.73}{0.06} & \ablationValue{5.72}{0.06} & \ablationValue{0.88}{0.00} & \ablationValue{1.93}{0.03} & \ablationValue{0.81}{0.00} & \ablationValue{0.41}{0.00} \\ %
    w/o Video & \ablationValue{12.70}{0.12} & \ablationValue{12.44}{0.12} & \ablationValue{0.70}{0.01} & \ablationValue{1.72}{0.03} & \ablationValue{0.66}{0.01} & \ablationValue{0.36}{0.00} \\ %
    \midrule
    w/o GME (w/ separate streams) & \ablationValue{3.83}{0.05} & \ablationValue{3.86}{0.05} & \ablationValue{0.93}{0.00} & \ablationValue{1.06}{0.02} & \ablationValue{0.87}{0.00} & \ablationValue{0.30}{0.00} \\ %
    w/o RGF (w/ ungated residual) & \ablationValue{4.31}{0.18} & \ablationValue{4.34}{0.19} & \ablationValue{0.92}{0.01} & \ablationValue{1.02}{0.04} & \ablationValue{0.86}{0.01} & \ablationValue{0.33}{0.01} \\ %
    w/o RSD (w/ direct force head) & \ablationValue{3.44}{0.08} & \ablationValue{3.47}{0.08} & \ablationValue{0.93}{0.00} & \ablationValue{1.09}{0.02} & \ablationValue{0.87}{0.00} & \ablationValue{0.34}{0.00} \\ %
    \bottomrule
  \end{tabular}}
  }
\end{table}

\section{Experiments}

\myPara{Baselines} We compare published methods adapted to EgoMyo, namely the visual pose estimators HaMeR~\cite{pavlakos2024hamer} and WiLoR~\cite{potamias2024wilor}, the sEMG estimators NeuroPose~\cite{liu2021neuropose} and vemg2pose~\cite{salter2024emg2pose}, the multimodal EMGFormer~\cite{xi2026egoemg}, MoT~\cite{liang2025mixture}, and HBridge~\cite{wang2025hbridge}, and the contact and force methods OpenTouch~\cite{song2025opentouch}, EgoTouch~\cite{zhou2026touchanything}, EgoTactile~\cite{zeng2026egotactile}, and EgoPressure~\cite{zhao2025egopressure}. Adaptations, including the MoT baseline, are architecture-level rather than reproductions of the original checkpoints. Readouts, budgets, and initialization per method are detailed in \cref{sec:suppl:baselines}.

\myPara{Implementation Details} Following WiLoR~\cite{potamias2024wilor}, the ViT in \cref{fig:framework} is a frozen ViT-H/16, and Full uses global pooling, gating, and state normalization with coefficient 0.1. Each seed runs 20 training-only alignment epochs followed by 20 task-training epochs with AdamW at learning rate $5\times10^{-4}$, batch size 256, and cosine decay on NVIDIA B200 GPUs. \cref{sec:suppl:baselines} reports the per-method training setup and confirms that no method uses extra test-time sensors.

\subsection{Comparison with Recent Strong Baselines}
\myPara{Cross-Trial Performance}
\cref{tab:main} compares all methods on the same TEST splits. On cross-trial, MyoBridge reaches 3.26\,mm MPJPE, about half the 6.16\,mm of vision-only WiLoR and well below the 5.42\,mm of the strongest fusion baseline HBridge, together with 0.95 regional contact F1, 0.86\,N regional force MAE and 0.88 surface contact F1. MoT, HBridge and MyoBridge all fall below 1.5\,N regional force MAE, while the best single-modality method stays at 1.71\,N, consistent with the complementary evidence in Finding~\hyperref[sec:empirical:finding1]{1}. The margin over HBridge is widest on geometry, at 39.8\%, so fusion also improves spatial estimation rather than only the force that sEMG informs most directly.

\myPara{Cross-Session and Cross-Subject}
All families incur higher geometry and regional force errors on both unseen recording sessions and unseen subjects. Despite sEMG shifts across donnings and users (\cref{fig:dataset:acquisition:emg_sessions}), MyoBridge keeps the lowest MPJPE, at 5.35 and 7.84\,mm, roughly half that of MoT. Its MPJPE gains over the best fusion baseline rise from 39.8\% on cross-trial to 49.9\% and 45.6\%, and contact F1 gains grow from 0.01 to 0.15 and 0.20. This cross-domain advantage is consistent with Finding~\hyperref[sec:empirical:finding2]{2}, which motivates admitting sEMG selectively by region rather than assigning a single contribution to the whole hand.

\myPara{Kinematics, Contact, and Force}
MyoBridge leads geometry, regional contact, and regional force on cross-trial, while its vertex force MAE of 0.29\,N trails HBridge by 0.01\,N despite better surface contact F1. Regional accuracy therefore does not guarantee a better surface force field. This distinction echoes Finding~\hyperref[sec:empirical:finding3]{3}, since force depends on both magnitude and its contact support. MyoBridge improves regional force without sacrificing surface contact, but its vertex force ranking shows that within-contact allocation remains a separate challenge.

\myPara{Computational Cost}
\Cref{tab:compute} shows the accuracy gains use a modest trainable task model. MyoBridge trains 9.41\,M parameters, close to MoT and HBridge, and decodes vertex contact and force at 0.712 GFLOPs and 0.0284\,ms per frame in batches, excluding visual encoding.

\subsection{Ablation Study}
\label{sec:res:ablation}

\myPara{Core Modules}
\cref{tab:ablation} removes one module at a time on cross-trial TEST. Without sEMG, regional force MAE rises from 0.86 to 1.93\,N and contact F1 falls from 0.95 to 0.88, while MPJPE rises by 76\%, so the wrist stream serves all three components. Without video, MPJPE rises nearly fourfold to 12.70\,mm, the largest degradation in the table, so vision is the spatial backbone. Removing GME costs 0.57\,mm of MPJPE, and replacing RSD with a direct force head raises force MAE to 1.09\,N. Removing RGF alone raises force MAE to 1.02\,N, so regional gating improves on an ungated sEMG residual. Vision beats sEMG on MPJPE and sEMG beats vision on regional force, as in Finding~\hyperref[sec:empirical:finding1]{1}, so the full model retains this division of labor and benefits from regional gating and contact-supported decoding.

\myPara{EMG Window Study}
\cref{fig:hparam} studies history and timing on development validation. Longer history mainly sharpens force, lowering its MAE by 6.4\% from 100 to 750\,ms, while contact F1 moves under 1\%. Moving a 250\,ms window to [\textminus1000,~\textminus750]\,ms raises force MAE by 29.4\%, and input after the frame raises it by 12.5\%, so exertion is read best just before the frame. Both trends support MyoBridge's causal sEMG window ending at the current frame.

\subsection{Beyond Aggregate Accuracy}\label{sec:res:analysis}

\begin{figure}[t]
  \centering
  \begin{subcaptiongroup}%
    \phantomsubcaption\label{fig:qualitative:pose}%
    \phantomsubcaption\label{fig:qualitative:contact}%
  \end{subcaptiongroup}%
  \includegraphics[width=\columnwidth]{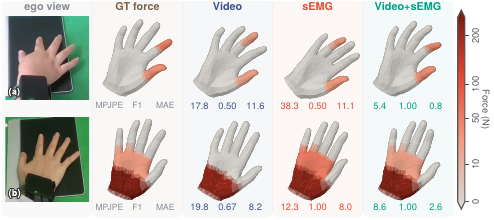}
  \caption{\textbf{Complementary examples.} (a)~Vision gives the better pose. (b)~sEMG finds the missed contact. Below each hand are the per-frame MPJPE (mm), contact F1, and force MAE (N).}
  \label{fig:qualitative}
\end{figure}

\begin{figure}[t]
  \centering
  \includegraphics[width=\columnwidth]{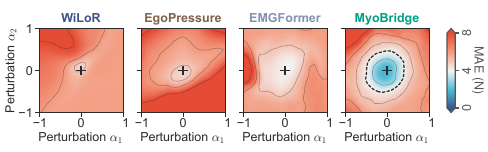}
  \caption{\textbf{Loss landscape.} Force MAE on 64 held-out frames around trained weights ($+$). White and dashed mark 4.04\,N, the lowest baseline value across the sampled perturbation slices.}
  \label{fig:landscape}
\end{figure}

\myPara{Complementarity}
\cref{fig:qualitative} illustrates Findings~\hyperref[sec:empirical:finding1]{1} and~\hyperref[sec:empirical:finding2]{2}. In \cref{fig:qualitative:pose}, vision places the hand better than sEMG (17.8 vs.\ 38.3\,mm), yet both misplace the contact. Fusion improves the visual pose and corrects the contact (F1 0.50 to 1.00), as sEMG refines regions vision localized. In \cref{fig:qualitative:contact}, vision misses an upper-palm contact that sEMG detects, and fusion recovers it with force MAE 2.6 vs.\ 8.2\,N. Here, fusion matches or beats the better stream on each component.

\myPara{Loss Landscape}
A low error may also reflect a sharp, fragile minimum. \cref{fig:landscape} perturbs each checkpoint along filter-normalized directions~\cite{li2018visualizing}. Only MyoBridge falls below the best baseline (1.88 vs.\ 4.04\,N) and stays below within a step of 0.4 across seeds, and along eight more random directions it holds the lowest force error up to a step of 0.9. A step of 0.1 moves its error by 0.06\,N, so the gain spans a broad basin in the sampled directions, consistent with stable force decoding under these parameter perturbations.

\myPara{Temporal Stability}
\cref{tab:temporal_teaser} evaluates temporal stability on 120 TEST trials, where the baselines divide by input. EMGFormer and vemg2pose time onsets more accurately (219 and 374\,ms against at least 494\,ms for vision), although NeuroPose shows no such gain, while vision-only methods hold pose more steadily. MyoBridge inherits both and leads seven metrics, with pose correlation 0.882, near-best jitter 1.224, segment F1 0.857, and onset error 105\,ms. The gain is clear on contact, where F1 rises from 0.736 to 0.857, so the support that bounds force in Finding~\hyperref[sec:empirical:finding3]{3} is more stable, alongside lower force error SD of 0.74\,N.

\begin{table}[t]
\centering
\caption{\textbf{Temporal metrics on 120 CT TEST trials.} $r$, error SD, onset ms, pose jitter mm/frame$^2$, flips/s.}
\small
\renewcommand{\arraystretch}{1.05}
\def\temporalrows{
\toprule
\multirow{2.5}{*}{\textbf{Method}} & \multicolumn{3}{c}{\textbf{Pose}} & \multicolumn{2}{c}{\textbf{Contact}} & \multicolumn{3}{c}{\textbf{Force}} \\
\cmidrule(lr){2-4}\cmidrule(lr){5-6}\cmidrule(lr){7-9}
 & \textbf{Corr.}$\uparrow$ & \textbf{SD}$\downarrow$ & \textbf{Jitter}$\downarrow$ & \textbf{F1}$\uparrow$ & \textbf{Spur.}$\downarrow$ & \textbf{Corr.}$\uparrow$ & \textbf{SD}$\downarrow$ & \textbf{Onset}$\downarrow$ \\
\midrule
HaMeR~\cite{pavlakos2024hamer} & \cellcolor{npgteal!0}0.768 & \cellcolor{npgteal!0}1.56 & \cellcolor{npgteal!0}1.320 & \cellcolor{npgteal!25}0.736 & \cellcolor{npgteal!25}0.14 & \cellcolor{npgteal!0}0.798 & \cellcolor{npgteal!0}2.04 & \cellcolor{npgteal!0}732 \\
WiLoR~\cite{potamias2024wilor} & \cellcolor{npgteal!25}0.802 & \cellcolor{npgteal!0}1.53 & \cellcolor{npgteal!45}\textbf{1.222} & \cellcolor{npgteal!0}0.659 & \cellcolor{npgteal!0}0.20 & \cellcolor{npgteal!0}0.847 & \cellcolor{npgteal!0}1.68 & \cellcolor{npgteal!0}827 \\
OpenTouch~\cite{song2025opentouch} & \cellcolor{npgteal!0}0.643 & \cellcolor{npgteal!0}1.74 & \cellcolor{npgteal!0}2.522 & \cellcolor{npgteal!0}0.519 & \cellcolor{npgteal!0}0.49 & \cellcolor{npgteal!0}0.808 & \cellcolor{npgteal!0}1.77 & \cellcolor{npgteal!0}494 \\
EgoTouch~\cite{zhou2026touchanything} & \cellcolor{npgteal!0}0.629 & \cellcolor{npgteal!0}1.71 & \cellcolor{npgteal!0}2.447 & \cellcolor{npgteal!0}0.461 & \cellcolor{npgteal!0}0.49 & \cellcolor{npgteal!0}0.769 & \cellcolor{npgteal!0}2.04 & \cellcolor{npgteal!0}872 \\
EgoTactile~\cite{zeng2026egotactile} & \cellcolor{npgteal!0}0.636 & \cellcolor{npgteal!0}1.74 & \cellcolor{npgteal!0}2.682 & \cellcolor{npgteal!0}0.511 & \cellcolor{npgteal!0}0.48 & \cellcolor{npgteal!0}0.794 & \cellcolor{npgteal!0}1.83 & \cellcolor{npgteal!0}654 \\
EgoPressure~\cite{zhao2025egopressure} & \cellcolor{npgteal!0}0.595 & \cellcolor{npgteal!0}2.49 & \cellcolor{npgteal!0}2.808 & \cellcolor{npgteal!0}0.145 & \cellcolor{npgteal!0}0.66 & \cellcolor{npgteal!0}0.527 & \cellcolor{npgteal!0}2.89 & \cellcolor{npgteal!0}2680 \\
NeuroPose~\cite{liu2021neuropose} & \cellcolor{npgteal!0}0.354 & \cellcolor{npgteal!0}3.16 & \cellcolor{npgteal!0}1.646 & \cellcolor{npgteal!0}0.028 & \cellcolor{npgteal!0}0.51 & \cellcolor{npgteal!0}0.577 & \cellcolor{npgteal!0}3.24 & \cellcolor{npgteal!0}4167 \\
vemg2pose~\cite{salter2024emg2pose} & \cellcolor{npgteal!0}0.364 & \cellcolor{npgteal!0}4.44 & \cellcolor{npgteal!0}5.951 & \cellcolor{npgteal!0}0.167 & \cellcolor{npgteal!0}1.65 & \cellcolor{npgteal!0}0.758 & \cellcolor{npgteal!0}2.19 & \cellcolor{npgteal!0}374 \\
EMGFormer~\cite{xi2026egoemg} & \cellcolor{npgteal!0}0.595 & \cellcolor{npgteal!0}2.15 & \cellcolor{npgteal!0}3.808 & \cellcolor{npgteal!0}0.348 & \cellcolor{npgteal!0}0.98 & \cellcolor{npgteal!0}0.874 & \cellcolor{npgteal!0}1.64 & \cellcolor{npgteal!0}219 \\
MoT~\cite{liang2025mixture} & \cellcolor{npgteal!0}0.690 & \cellcolor{npgteal!0}1.48 & \cellcolor{npgteal!0}2.475 & \cellcolor{npgteal!0}0.600 & \cellcolor{npgteal!0}0.39 & \cellcolor{npgteal!0}0.874 & \cellcolor{npgteal!0}1.70 & \cellcolor{npgteal!0}280 \\
HBridge~\cite{wang2025hbridge} & \cellcolor{npgteal!0}0.733 & \cellcolor{npgteal!25}1.37 & \cellcolor{npgteal!0}2.239 & \cellcolor{npgteal!0}0.728 & \cellcolor{npgteal!0}0.23 & \cellcolor{npgteal!25}0.917 & \cellcolor{npgteal!25}1.06 & \cellcolor{npgteal!25}125 \\
MyoBridge & \cellcolor{npgteal!45}\textbf{0.882} & \cellcolor{npgteal!45}\textbf{0.94} & \cellcolor{npgteal!25}1.224 & \cellcolor{npgteal!45}\textbf{0.857} & \cellcolor{npgteal!45}\textbf{0.09} & \cellcolor{npgteal!45}\textbf{0.964} & \cellcolor{npgteal!45}\textbf{0.74} & \cellcolor{npgteal!45}\textbf{105} \\
\bottomrule}
\resizebox{\columnwidth}{!}{\begin{tabular}{lcccccccc}\temporalrows\end{tabular}}
\label{tab:temporal_teaser}
\end{table}

\vspace{4.6pt}
\section{Conclusion and Discussion}
\label{sec:conclusion}

We formulated physical hand state estimation from egocentric video and wrist sEMG and introduced EgoMyo, which registers bare-hand pressure onto a reconstructed mesh across repeated donnings. Analysis on EgoMyo shows that vision localizes where the hand touches, sEMG senses how strongly it presses, and force is bounded by its contact support. MyoBridge follows these findings with region-gated fusion and contact-supported decoding.

\myPara{Limitations and Future Work}
EgoMyo covers bare-hand pressing on a planar surface, so grasping, curved objects, and shear are natural extensions. Wrist sEMG across donnings and users remains open, and repeated donnings make it measurable, motivating session-invariant encoders.

\clearpage
\section*{LLM Usage Statement}
Claude (Anthropic) and Codex (OpenAI) were used for language editing of the manuscript text and for checking numerical consistency between the text and tables. The authors take full responsibility for the methodology, experimental results, and conclusions. All scientific claims, experiments, and analyses are the authors' own.

\clearpage
\twocolumn[\centering\Large\bfseries Appendix\par\vspace{12pt}]
\raggedbottom
\setcounter{dbltopnumber}{3}
\renewcommand{\dbltopfraction}{0.85}
\renewcommand{\dblfloatpagefraction}{0.85}
\setlength{\dbltextfloatsep}{8pt plus 2pt minus 2pt}
\makeatletter
\setlength{\@dblfptop}{0pt}
\makeatother
\setcounter{table}{0}
\setcounter{figure}{0}
\setcounter{section}{0}
\setcounter{equation}{0}

\renewcommand{\theequation}{A\arabic{equation}}
\renewcommand{\thetable}{A\arabic{table}}
\renewcommand{\thefigure}{A\arabic{figure}}
\captionsetup[sub]{subrefformat=simple}
\renewcommand{\thesection}{\Alph{section}}
\renewcommand{\theHequation}{appendix.\arabic{equation}}
\renewcommand{\theHtable}{appendix.\arabic{table}}
\renewcommand{\theHfigure}{appendix.\arabic{figure}}
\renewcommand{\theHsection}{appendix.\Alph{section}}
\renewcommand{\theHsubsection}{appendix.\Alph{section}.\arabic{subsection}}

\startcontents[appendices]
\section*{Appendix Contents}
\begingroup
\renewcommand{\numberline}[1]{#1\hspace{1em}}
\printcontents[appendices]{}{1}{\setcounter{tocdepth}{2}}
\endgroup

\section{Dataset Construction Details}
\label{sec:suppl:dataset}

\textbf{EgoMyo} abbreviates \emph{Egocentric Myoelectric}. It pairs egocentric video and wrist sEMG with hand geometry, contact, and normal-force annotations on a shared mesh. This section details acquisition, contact configurations, annotation and quality control, evaluation splits, and metrics.

\subsection{Acquisition and Configuration}
\label[appendixsubsection]{sec:suppl:configuration}

\myPara{Participants} We recorded 51 right-handed adults (26 male, 25 female), aged 18--48 years (mean 25.6, SD 7.4). The benchmark excludes one participant whose uncalibrated peak forces exceed the 1.5 interquartile range fence of participant medians (\cref{sec:benchmark:stats}). Benchmark statistics and results use these 50 participants. Volunteers reported no condition affecting hand or wrist function. Both hands were recorded in two sessions with independent wristband donning. Participants gave written informed consent under institutional ethics approval and were compensated.

\myPara{Acquisition Hardware} \cref{tab:suppl:hardware} summarizes device placement, sampling rates, and output formats. Egocentric video and wrist sEMG are model inputs, while the external cameras and pressure surface provide kinematics, contact, and force supervision only during collection. The five-camera array uses the egocentric camera as the hardware synchronization master and four external cameras as slaves. RGB frames are matched separately using recorded timestamps, forming unique frame groups spanning at most 10\,ms. Reconstruction requires at least three valid views and the geometric checks in \cref{tab:suppl:annotation}.

Within each run, video, sEMG, and pressure use the host's monotonic time reference. Camera SDK timestamps are mapped to it. sEMG times follow the nominal 2\,kHz rate within continuous segments, while pressure uses host-side retrieval timestamps. Annotation alignment pairs each egocentric frame with the preceding 200\,ms of sEMG and the nearest valid pressure sample within 20\,ms in the same trial. Incomplete windows are excluded, unavailable pressure labels are masked, and alignment never crosses recording gaps. Model-specific histories appear in \cref{sec:suppl:baselines}. All methods read identical samples.

A box beside the Sensel sensing area serves as a passive wrist rest, reducing arm-support effort during sustained presses without touching the sEMG electrodes or obstructing the cued finger and palm contacts. This support keeps sustained presses repeatable across trials and donnings, and EgoMyo covers this supported tabletop interaction.

\begin{table}[t]
  \centering
  \small
  \setlength{\tabcolsep}{3pt}
  \caption{Acquisition hardware and settings. Device crops show the hardware, with the same camera model in both groups.}
  \label[appendixtable]{tab:suppl:hardware}
  \begin{tabular}{@{}>{\raggedright\arraybackslash}p{0.20\linewidth}>{\raggedright\arraybackslash}p{0.25\linewidth}>{\raggedright\arraybackslash}p{\dimexpr0.55\linewidth-12pt\relax}@{}}
    \toprule
    \textbf{Device} & \textbf{Parameter} & \textbf{Specification} \\
    \midrule
    \multirow[c]{5}{=}{\centering Egocentric camera\par\smallskip\includegraphics[width=0.95\linewidth,height=10mm,keepaspectratio]{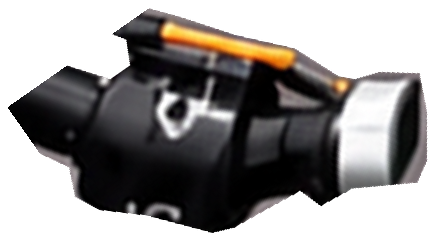}} & Number & One \\
    & Brand / model & Astralive Cube 1.0\par{\scriptsize\nolinkurl{https://www.xy-ai.tech/en/}} \\
    & Placement & Fixed at the participant's eye level on their side, facing the workspace \\
    & Frame rate & 30\,Hz \\
    & Output & Egocentric video \\
    \midrule
    \multirow[c]{5}{=}{\centering External cameras\par\smallskip\includegraphics[width=0.95\linewidth,height=10mm,keepaspectratio]{figs/devices/astralive_cube_rectified}} & Number & Four RGB cameras \\
    & Brand / model & Astralive Cube 1.0\par{\scriptsize\nolinkurl{https://www.xy-ai.tech/en/}} \\
    & Arrangement & Surrounding the interaction area \\
    & Calibration & Calibrated multi-view geometry \\
    & Role & Hand mesh and joint reconstruction \\
    \midrule
    \multirow[c]{6}{=}{\centering sEMG wristband\par\smallskip\includegraphics[width=0.95\linewidth,height=10mm,keepaspectratio]{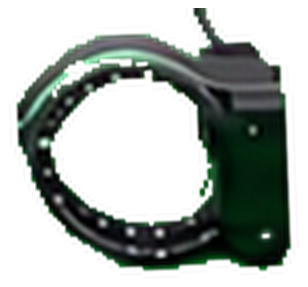}} & Number & One \\
    & Brand / model & Cortix AI\par{\scriptsize\nolinkurl{https://getcortix.ai/}} \\
    & Channels & 18 \\
    & Sampling rate & 2\,kHz \\
    & Placement & Active forearm \\
    & Output & Multichannel sEMG \\
    \midrule
    \multirow[c]{7}{=}{\centering Pressure surface\par\smallskip\includegraphics[width=0.95\linewidth,height=10mm,keepaspectratio]{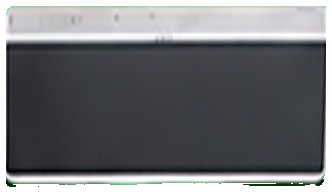}} & Number & One \\
    & Brand / model & Sensel Morph~\cite{senselmorph}\par{\scriptsize\nolinkurl{https://sensel.com/}} \\
    & Sensor array & 105 rows $\times$ 185 columns \\
    & Acquisition rate & 30\,Hz \\
    & Placement & Fixed to the tabletop \\
    & Output & Spatial normal-force map \\
    & Force unit & Normal force in newtons (N), converted by the sensor SDK \\
    \bottomrule
  \end{tabular}
\end{table}

\begin{figure}[h]
  \centering
  \setlength{\tabcolsep}{1pt}
  \begin{tabular}{@{}*{5}{>{\centering\arraybackslash}p{\dimexpr(\linewidth-8pt)/5\relax}}@{}}
    \includegraphics[width=\linewidth]{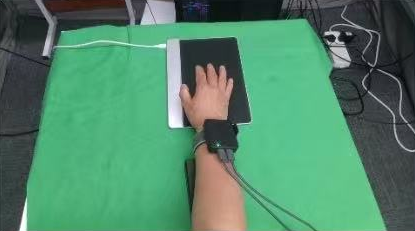} &
    \includegraphics[width=\linewidth]{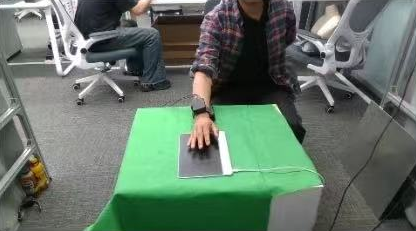} &
    \includegraphics[width=\linewidth]{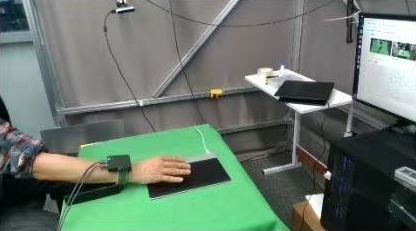} &
    \includegraphics[width=\linewidth]{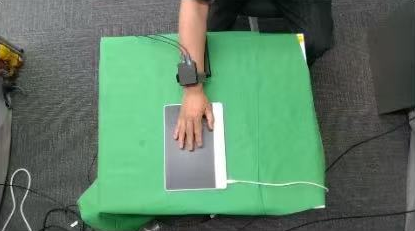} &
    \includegraphics[width=\linewidth]{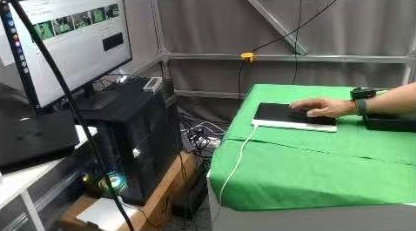} \\
    \small Ego view & \small External view 1 & \small External view 2 &
    \small External view 3 & \small External view 4
  \end{tabular}
  \caption{\textbf{Camera views of EgoMyo.} Video and wrist sEMG are model inputs. All five cameras support hand reconstruction, with pressure providing contact and force labels.}
  \label[appendixfigure]{fig:dataset:setup}
\end{figure}

\myPara{Capture Views} \cref{fig:dataset:setup} shows the egocentric view and four external views of the same interaction area. The egocentric camera is fixed on the participant's side at eye level, facing the workspace. This mounting keeps its calibration with the pressure surface constant across trials while preserving the first-person viewpoint. All five calibrated views support multiview hand reconstruction and its registration to the fixed instrumented pressure surface.

\begin{figure}[t]
  \centering
  \includegraphics[width=\linewidth]{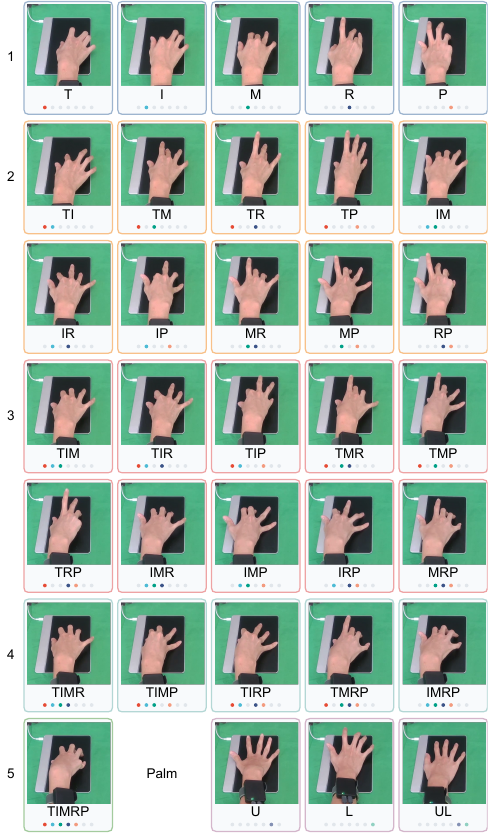}
  \caption{\textbf{EgoMyo contact configurations.} Each card shows the egocentric
  frame at peak total force for one configuration, all from the participant
  shown in \cref{fig:dataset:acquisition:video}. Names, border colors, and region
  dots follow \cref{fig:dataset:acquisition:configs}. Rows group
  configurations by active digit count, with the three palm conditions in the final row.}
  \label[appendixfigure]{fig:dataset:configs}
\end{figure}

\myPara{Configuration Coverage} \cref{fig:dataset:configs} covers 34 configurations, with 31 non-empty digit combinations and three palm conditions. Configurations follow digit count, then palm contacts. Participants repeat the set with both hands in two independently donned sessions. Five repetitions per session give 340 planned trials per hand. Participants press the cued regions, hold for at least 3\,s at a self-selected intensity, and release, varying force within each configuration. Cues specify intended regions, while frame-level labels follow measured regional forces and release status.

\subsection{Annotation Settings}
\label[appendixsubsection]{sec:suppl:annotation}

\myPara{Annotation Pipeline} The offline annotation pipeline integrates multiview reconstruction, temporal refinement, MANO fitting, and pressure-to-surface mapping. \cref{tab:suppl:annotation} summarizes the reconstruction, force mapping, and quality-control procedures used to produce synchronized hand-state labels. Calibrated multiview geometry determines the hand surface, while baseline-corrected pressure measurements provide normal force on that surface. The table lists region aggregation and trial acceptance rules for constructing the synchronized benchmark annotations.

Before recording, the board pose is checked across calibrated views and fixed in the common reference frame. Reconstructing the hand in that frame establishes its relation to the board plane for pressure correspondence. Camera extrinsics follow $X_c=R_cX_{\mathrm{ref}}+t_c$, with translation in meters. Meshes and joints use the egocentric camera frame.

\begin{table}[htbp]
  \centering
  \footnotesize
  \setlength{\tabcolsep}{3pt}
  \caption{Annotation procedures and settings for EgoMyo.}
  \label[appendixtable]{tab:suppl:annotation}
  \begin{tabular}{@{}>{\raggedright\arraybackslash}p{0.23\linewidth}>{\raggedright\arraybackslash}p{\dimexpr0.77\linewidth-6pt\relax}@{}}
    \toprule
    \textbf{Stage} & \textbf{Procedure and settings} \\
    \midrule
    Multi-view joints & Estimate 21 image-space joints per view with HBR Engine~\cite{hbr2025engine}, retaining hand confidence $\geq0.25$ and uniquely matched frame groups spanning at most 10\,ms. Triangulation requires at least three inlier views with positive depth, reprojection error $\leq4$ pixels per inlier, and a maximum ray angle $\geq1^\circ$. \\
    \midrule
    Joint smoothing & Refine triangulated trajectories with SeqHand~\cite{ren2025prioraware} before MANO fitting, using hand priors and 27-frame windows within each trial. Windows stay within recording segments and never cross gaps over 0.2\,s. \\
    \midrule
    MANO fitting & Initialize MANO~\cite{romero2017mano} from available egocentric WiLoR~\cite{potamias2024wilor} predictions. Fit refined 3D joints and multiview 2D observations with temporal regularization and shared shape. Output mesh $M_t$ and joints $K_t$ in the egocentric frame, retaining observation and completion flags. \\
    \midrule
    Pressure registration & Register the mesh to the Sensel Morph plane using the fixed board pose and camera extrinsics. Subtract per-cell means from the same donning's unloaded baseline, nominally 30\,s, retaining its recorded duration. Convert readings to newtons with the vendor SDK. Apply pressure cleanup and consistency checks. \\
    \midrule
    Surface force & For each mapped pressure cell, retain the mesh triangle, normalized nonnegative barycentric coordinates, region label, and force in newtons. Sum cell forces per triangle. Cells without a valid mesh correspondence are excluded (approximately 0.02\% of loaded pressure cells). Cell forces are lifted to vertices with the stored barycentric coordinates, restricted to the vertices of the cell's own region, renormalized to sum to one, and accumulated once per cell, so the conversion preserves the total and the regional forces. \\
    \midrule
    Force aggregation & Use fixed palmar triangle patches for the five digits and upper/lower palm. Each assigned cell has one region label and contributes its force once to that region. Vertex forces satisfy $F_{t,r}=\sum_{v\in r}f_t(v)$. Each vertex joins its incident faces' majority region, yielding seven disjoint vertex sets. \\
    \midrule
    Contact labeling & For available samples not marked as released, set $C_{t,r}=\mathbbm{1}[F_{t,r}>0]$ after pressure cleanup. Released samples are labeled as no contact. Retain availability and quality flags to distinguish uncertain or missing labels from confirmed no contact. \\
    \midrule
    Trial acceptance & Require contact lasting at least 3\,s and valid recordings. Check contact--cue agreement and sEMG channel quality. Collectors review ambiguous contacts and motion artifacts. Failed trials are re-recorded. \\
    \bottomrule
  \end{tabular}
\end{table}

MANO fitting combines confidence-weighted 3D joint and multiview reprojection terms with shape, pose, and temporal regularization. Shared shape is optimized for 12 Adam steps at a learning rate of 0.012. Pose fitting uses windows of up to 48 frames with an eight-frame overlap, 40 Adam steps at a learning rate of 0.015, and up to 12 L-BFGS iterations. Windows respect trial and segment boundaries.

\myPara{Quality Control} Geometric limits in \cref{tab:suppl:annotation} apply to triangulated observations, with model-completed joints identified separately. Failed or nonfinite MANO fits trigger review with reprojection diagnostics. A baseline-corrected pressure cell is active above $\max(5\sigma_i,4.9\times10^{-4}\,\mathrm{N})$, where $\sigma_i$ is its unloaded baseline standard deviation. Cells outside the calibrated surface are excluded, and unassigned force is reported separately from the seven regional values. Finger trials require contact-component count and order to match the cue. Palm trials are checked for the cued half or halves. Merged, extra, ambiguously ordered, or unassigned contacts trigger review. Unlabeled edge force is tolerated only when target regions agree, finger trials have the cued count, and unassigned force is at most 0.05\,N and 10\% of total force. All 18 sEMG channels must contain finite samples. A channel is flagged for trial-wide SD below 0.5 ADC counts, a constant run of at least 1000 samples (0.5\,s), or any sample at a signed 16-bit ADC rail (\textminus32768 or 32767). Collectors review finger identity and suspected motion artifacts. A topology warning alone does not establish incorrect contact. Confirmed failures are marked invalid and re-recorded. Exclusion logs retain the reason, reviewer, and timestamp.

Pressure cleanup repairs short downward anomalies only when neighboring readings are stable, pressure footprints agree, and the interval avoids transitions and timing gaps. Interpolation anchors must span at most 100\,ms, with footprint cosine similarity at least 0.95 and a 300\,ms guard from contact and release boundaries. Force stability, signal-to-noise ratio, and SDK contact evidence are checked jointly. Repairs interpolate linearly at recorded timestamps. Residual force after release is removed only when low pressure, absent SDK contacts, and a contracted footprint persist for at least 300\,ms without rebound or re-contact. Raw readings and uncertainty flags are retained. Regional and unassigned forces must sum to total board force for each finite frame, including the intermediate surface-face representation.

\subsection{Split Construction}
\label[appendixsubsection]{sec:suppl:splits}

All three protocols use 33{,}989 trials from 50 participants (100 hands), following the exclusion described in \cref{sec:benchmark:stats}. Complete trials are assigned identically to TRAIN or TEST across methods and runs (\cref{tab:suppl:splits}).

\begin{table}[t]
  \centering
  \caption{\textbf{Cross-trial, cross-session, and cross-subject splits.} Frame counts precede metric-specific validity filtering.}
  \label[appendixtable]{tab:suppl:splits}
  \small
  \setlength{\tabcolsep}{3pt}
  \begin{tabular}{lrrrr}
    \toprule
    \multirow{2.5}{*}{\textbf{Protocol}} & \multicolumn{2}{c}{\textbf{TRAIN}} & \multicolumn{2}{c}{\textbf{TEST}} \\
    \cmidrule(lr){2-3}\cmidrule(lr){4-5}
    & \textbf{Trials} & \textbf{Frames} & \textbf{Trials} & \textbf{Frames} \\
    \midrule
    Cross-trial & 20{,}399 & 3{,}453{,}937 & 13{,}590 & 2{,}300{,}961 \\
    Cross-session & 16{,}995 & 2{,}889{,}496 & 16{,}994 & 2{,}865{,}402 \\
    Cross-subject & 23{,}797 & 4{,}014{,}944 & 10{,}192 & 1{,}739{,}954 \\
    \bottomrule
  \end{tabular}
\end{table}

\myPara{Cross-Trial} Trials are stratified by participant, hand, session, and configuration. Each complete stratum assigns three repetitions to TRAIN and two to TEST. Of 6{,}800 strata, 6{,}790 are complete. Nine have four repetitions and one has three, allocated 3/1 and 2/1 to TRAIN/TEST, respectively. All trials are retained, testing unseen repetitions within the same participant, hand, and session.

\myPara{Cross-Session} For each hand, all session-1 trials form TRAIN and all session-2 trials form TEST. All 100 hands have both sessions, with independent wristband donning between sessions. Participants and contact configurations are shared across partitions, allowing evaluation of generalization across donnings for the same individuals.

\myPara{Cross-Subject} The 50 participants are randomly partitioned into 35 training and 15 test participants using a fixed split. Both hands, both sessions, and all trials of each participant remain in the same partition. Contact configurations are shared across partitions, but participants are disjoint. This protocol tests generalization to unseen individuals.

\myPara{Development and Final Training} Development validation holds out one TRAIN trial per participant, hand, session, and configuration for cross-trial, and per participant, hand, and configuration in session 1 for cross-session. Development-training/validation sizes are 13{,}599/6{,}800 and 13{,}595/3{,}400 trials. Final models train on full TRAIN, including validation trials, with fixed hyperparameters. Cross-subject uses all 35 TRAIN participants without validation and evaluates the final checkpoint.

\myPara{Leakage Controls} Temporal windows remain within trial boundaries, so TRAIN and TEST share neither trials nor input windows. Normalization and data-dependent scales use only the training subset of each stage. Dataset-specific pretraining and model fitting run separately for each protocol, as their splits differ. For cross-subject, normalization and pretraining use only the 35 training participants. The exclusion criterion uses recorded force statistics alone.

\subsection{Evaluation Metrics}
\label[appendixsubsection]{sec:suppl:metrics}

\subsubsection{Primary Metrics}

\myPara{Kinematic Errors} Let $\mathcal{T}$ contain valid TEST frames, $J=21$, and $N_v=778$. Coordinates $k^{\mathrm{rel}}$ and $m^{\mathrm{rel}}$ subtract each hand's wrist, with hats denoting predictions. MPJPE and PVE average corresponding joint and vertex distances,
\begin{align}
 \mathrm{MPJPE}&=\frac{\sum_{t\in\mathcal{T}}\sum_{j=1}^{J}\|\hat{k}^{\mathrm{rel}}_{t,j}-k^{\mathrm{rel}}_{t,j}\|_2}{|\mathcal{T}|J},\\
 \mathrm{PVE}&=\frac{\sum_{t\in\mathcal{T}}\sum_{v=1}^{N_v}\|\hat{m}^{\mathrm{rel}}_{t,v}-m^{\mathrm{rel}}_{t,v}\|_2}{|\mathcal{T}|N_v}.
\end{align}
Both use mm, with lower better. Wrist alignment retains rotation, scale, and articulation errors.

\myPara{Regional Contact F1} For each region $r$, count true positives $\mathrm{TP}_r$, false positives $\mathrm{FP}_r$, and false negatives $\mathrm{FN}_r$ over frames with valid contact labels. The main comparison averages the seven region scores equally,
\begin{equation}
 \mathrm{F1}_{\mathrm{macro}}=\frac{1}{7}\sum_{r=1}^{7}
 \frac{2\mathrm{TP}_r}{2\mathrm{TP}_r+\mathrm{FP}_r+\mathrm{FN}_r}.
\end{equation}
Higher is better. Surface fields share the region partition, with probabilities at least 0.5 labeled as contact. With no positive labels or predictions, a region scores 1. Analysis uses participant-level micro-F1 (\cref{sec:suppl:analysis:settings}).

\myPara{Regional Force MAE} Let $\mathcal{T}_r$ contain frames with valid force labels for region $r$. Mean absolute error averages these errors within regions and then equally across regions,
\begin{equation}
 \mathrm{MAE}_{\mathrm{region}}=\frac{1}{7}\sum_{r=1}^{7}
 \frac{1}{|\mathcal{T}_r|}\sum_{t\in\mathcal{T}_r}|\hat{F}_{t,r}-F_{t,r}|.
\end{equation}
The unit is N, with lower values better. All regions contribute, penalizing force outside reference contact support.

\myPara{Surface Contact F1} Let $\hat{\mathcal{C}}_t$ and $\mathcal{C}_t$ denote predicted and reference contact sets on the common MANO topology. Predictions use the 0.5 threshold. For geodesic tolerance $\delta=5$\,mm, precision and recall are
\begin{align}
 P&=\frac{\sum_{t\in\mathcal{T}}|\{v\in\hat{\mathcal{C}}_t:d_g(v,\mathcal{C}_t)\leq\delta\}|}{\sum_{t\in\mathcal{T}}|\hat{\mathcal{C}}_t|},\\
 R&=\frac{\sum_{t\in\mathcal{T}}|\{v\in\mathcal{C}_t:d_g(v,\hat{\mathcal{C}}_t)\leq\delta\}|}{\sum_{t\in\mathcal{T}}|\mathcal{C}_t|}.
\end{align}
F1 is $2PR/(P+R)$, with higher values better. The distance $d_g$ follows edges of the fixed rest mesh. Vertices outside the region partition are excluded. Counts pool valid TEST frames. Two empty pooled sets score 1, while one empty set or zero matched contact scores 0.

\myPara{Surface Force MAE} On valid contact vertices,
\begin{equation}
 \mathrm{MAE}_{\mathrm{surface}}=
 \frac{\sum_{t\in\mathcal{T}}\sum_{v\in\mathcal{C}_t}|\hat{f}_t(v)-f_t(v)|}
 {\sum_{t\in\mathcal{T}}|\mathcal{C}_t|}.
\end{equation}
The unit is N, with lower values better. Reference contact vertices inside the partition contribute equally. Regional MAE penalizes force outside this support.

\subsubsection{Temporal Metrics}

\myPara{Scope and Aggregation} \cref{tab:temporal_teaser} uses 120 fixed cross-trial TEST trials, totaling 20{,}240 frames at 30\,Hz. Final-epoch weights are evaluated without smoothing or selection. Runs are averaged within trials, then trials equally. Metrics use shared valid trials. Pose correlation, error SD, and jitter use 85, 100, and 96 trials. Contact metrics use 120. Force correlation, error SD, and onset use 118, 120, and 119 trials.

\myPara{Pose Stability} Correlation averages Pearson correlations over root-relative coordinates with reference SD at least 1\,mm and at least ten fully valid frames. Constant predictions score zero. Error SD is the population SD of per-frame MPJPE over fully valid frames. Jitter averages second differences over joints and valid frame triplets,
\begin{equation}
 \mathrm{jitter}_{\mathrm{pose}}=\frac{\sum_{t\in\mathcal{T}'}\sum_{j=1}^{J}\big\|\hat{k}^{\mathrm{rel}}_{t+1,j}-2\hat{k}^{\mathrm{rel}}_{t,j}+\hat{k}^{\mathrm{rel}}_{t-1,j}\big\|_2}{|\mathcal{T}'|J},
\end{equation}
in mm/frame$^2$, where $\mathcal{T}'$ contains valid triplet centers. Lower is better, although genuine acceleration contributes. Correlation measures agreement, jitter temporal roughness.

\myPara{Contact Segments} Probabilities at least 0.5 indicate contact. Within each region's valid run of at least three frames, segments are maximal contacting intervals. Predictions match unused references in time order at temporal IoU at least 0.5. Segment F1 is $2\mathrm{TP}/(n_{\mathrm{pred}}+n_{\mathrm{ref}})$, averaged over runs with any predicted or reference contact. Spurious flips lack reference flips within three frames, and their total count is divided by valid transition seconds.

\myPara{Force Timing} Frames require valid force labels for all regions. Pearson correlation compares predicted and reference total force, requiring both to vary. Error SD is the population SD of per-frame regional MAE. Onset compares first crossings of 20\% of the valid reference peak, when this exceeds 1\,N. Missing crossings use the first frame beyond the trial. Frame differences convert to ms at 30\,Hz.

\myPara{Units and Direction} Pose and force error SD use mm and N. Jitter uses mm/frame$^2$, spurious flips use s$^{-1}$, and onset uses ms. Higher is better for correlation and segment F1. Lower is better for error SD, jitter, spurious rate, and onset error. Temporal metrics average trials equally.

\section{Experimental and Analysis Details}
\label{sec:suppl:analysis}

\subsection{Training Protocol}
\label[appendixsubsection]{sec:suppl:protocol}

\myPara{Data and Evaluation} Main comparisons and ablations use the TEST partitions of the 50-participant population in \cref{sec:suppl:splits}. Development validation stays within TRAIN for hyperparameter selection and window studies. Normalization and loss scales are fitted separately for each protocol using training data alone. Final models train on full TRAIN without early stopping and use the last prescribed checkpoint. \cref{tab:main} reports mean and sample SD across runs.

\myPara{MyoBridge Training} Each protocol trains independently, without transferring task or alignment weights across protocols. GME uses frozen WiLoR~\cite{potamias2024wilor} visual tokens and a trainable sEMG encoder. Each run's 20-epoch TRAIN-only alignment checkpoint initializes 20 task-training epochs. Both stages use AdamW at peak learning rate $5\times10^{-4}$, weight decay $10^{-2}$, gradient clipping at 5, and two warmup epochs followed by cosine decay. Task training uses batch size 256. RGF uses pooled states, gated residuals, and normalization coefficient 0.1 (\cref{sec:region_fusion}).

\myPara{Baseline Training Budgets} HaMeR~\cite{pavlakos2024hamer} and WiLoR~\cite{potamias2024wilor} train in cross-subject for two task epochs with a peak learning rate of $10^{-4}$, weight decay $10^{-2}$, gradient clipping at 5, and cosine decay with a two-epoch warmup. Both models use batches of 32 images. NeuroPose~\cite{liu2021neuropose} and EMGFormer~\cite{xi2026egoemg} train for 20 epochs with AdamW, batch size 64, learning rate $10^{-4}$, weight decay $10^{-4}$, gradient clipping at 1, and cosine decay, and cross-subject training uses batch size 256. vemg2pose~\cite{salter2024emg2pose} trains for 20 epochs with batch size 256 and constant encoder/decoder learning rates of $10^{-4}$/$10^{-3}$. OpenTouch~\cite{song2025opentouch}, EgoTouch~\cite{zhou2026touchanything}, EgoTactile~\cite{zeng2026egotactile}, and EgoPressure~\cite{zhao2025egopressure} train their local adaptations for 20 task epochs with AdamW, batch size 256, learning rate $10^{-3}$, weight decay $10^{-4}$, and gradient clipping at 1. Their native-model learning rate is constant. For these four methods and the temporal sEMG model, added readouts use a separate AdamW optimizer at $10^{-4}$ with cosine decay and weight decay $10^{-4}$, and their losses also update the shared features. Each method keeps its own training recipe, while all methods share the same TRAIN data, labels, and TEST evaluation. Evaluation batch size leaves temporal context, masks, coverage, and aggregation unchanged.

\myPara{EMG Window Study} The top row of \cref{fig:hparam} compares 200, 400, 600, 1,500, and 2,000 samples, corresponding to histories of 100, 200, 300, 750, and 1,000\,ms at 2\,kHz. Each history ends at the target frame. Every configuration uses 20 TRAIN-only alignment epochs at its own history length and 20 task-training epochs, temperature 0.07, and state-normalization coefficient 0.1. Relative improvements use each run's 100\,ms reference and average across runs.

\myPara{EMG Window Timing} The bottom row of \cref{fig:hparam} shifts a fixed 500-sample window (250\,ms at 2\,kHz) to end at $-750$, $-500$, $-250$, 0, or $+250$\,ms relative to the target frame. All five configurations share the TRAIN-only alignment checkpoint trained with a 400-sample window. Task training lasts 20 epochs with AdamW at $5\times10^{-4}$, weight decay $10^{-2}$, gradient clipping at 5, and cosine decay with two warmup epochs. Training uses 2,301,779 development frames, and both window studies evaluate on the same 6,800 development-validation trials (1,152,158 frames). Timing curves report relative changes against $[-250,0]$\,ms. Future sEMG serves only as an offline control, and MyoBridge reads past-only sEMG ending at the target frame.

\subsection{Baseline Details}
\label[appendixsubsection]{sec:suppl:baselines}

\myPara{Joint-Task Adaptation} The methods in \cref{tab:main} are architecture-level adaptations to EgoMyo inputs and labels. Native predictions are retained, with direct auxiliary readouts from shared features for missing tasks. Added contact readouts use sigmoid outputs and binary cross-entropy. Force readouts use softplus and masked $\ell_1$ supervision. Added geometry and force losses use TRAIN-derived scales. Each row reports six metrics from one jointly trained model. Inference needs no extra sensors or reference pose.

\myPara{Visual Pose Estimators} HaMeR~\cite{pavlakos2024hamer} and WiLoR~\cite{potamias2024wilor} take RGB hand crops and retain their native joint and mesh prediction paths, using fixed MANO~\cite{romero2017mano} geometry. Added readouts predict regional and vertex-level contact and force. Only the visual encoder uses the public checkpoint. Method-specific modules and readouts start afresh, with all learned parameters trainable. Encoders retain task-specific pretraining, with crops from a pretrained detector. Both methods operate frame by frame.

\myPara{sEMG Pose Estimators} NeuroPose~\cite{liu2021neuropose} uses the encoder--residual--decoder architecture provided by the EgoEMG implementation~\cite{xi2026egoemg}. The current adaptation removes the final decoder ReLU so that signed features reach the task readouts. vemg2pose~\cite{salter2024emg2pose} uses a time-depth-separable temporal encoder with Cartesian joint and mesh readouts. Both methods receive an 18-channel, 2,000-sample sEMG history ending at the target frame, without visual input. Auxiliary readouts extend their outputs to the remaining geometry, contact, and force tasks. These Cartesian readouts are specific to the benchmark's joint and mesh labels.

\myPara{Multimodal Estimator} EMGFormer~\cite{xi2026egoemg} adapts residual fusion with its published temporal convolution, rotary-position transformer, and attention pooling. It fuses the past-only sEMG history with pooled frozen DINOv2~\cite{oquab2023dinov2} features. A learned multimodal residual refines a visual joint estimate, and auxiliary readouts predict the mesh, regional contact and force, and vertex fields. Supervision applies at the history endpoint in Cartesian coordinates.

\myPara{Compact Multimodal Transformers} MoT~\cite{liang2025mixture} and HBridge~\cite{wang2025hbridge} use frozen DINOv2~\cite{oquab2023dinov2} visual tokens and a three-layer 1D CNN mapping an 18-channel, 400-sample trailing sEMG window to 16 tokens. Six-layer MoT uses two 256-dimensional experts with separate attention projections, feed-forward layers, and normalization, plus global attention at every layer. Six-layer HBridge uses 256/192-dimensional branches, joint attention in layers 3 and 4, and 16 semantic reconstruction tokens trained by cosine reconstruction of pooled DINOv2 features. Direct readouts predict all six tasks. They have 10.98M and 9.31M trainable parameters, excluding visual encoding. HBridge replaces the original pretrained understanding/generation experts and diffusion objective with supervised hand-state prediction. Both train for 20 epochs in three runs using AdamW at $5\times10^{-4}$, weight decay $10^{-2}$, batch size 256, two warmup epochs, cosine decay, and gradient clipping at 5. TEST uses fixed epoch-20 weights.

\myPara{Visual Contact and Force Estimators} OpenTouch~\cite{song2025opentouch}, EgoTouch~\cite{zhou2026touchanything}, EgoTactile~\cite{zeng2026egotactile}, and EgoPressure~\cite{zhao2025egopressure} use local adaptations with frozen DINOv2~\cite{oquab2023dinov2} features. The first three use token-transformer and MLP predictors, EgoPressure an MLP. All predict regional contact and force, plus auxiliary geometry and vertex readouts.

\myPara{Computational Cost} In \cref{tab:compute}, total parameters include frozen encoders, while trainable parameters exclude them. Feature-based methods report task FLOPs and latency after visual encoding. HaMeR and WiLoR include their native RGB path. MyoBridge's 0.0284\,ms is batch-amortized task-forward time on a B200 with BF16 and batch size 512. It excludes detection, cropping, feature extraction, transfers, and scoring. FLOPs are counted per operator with a MANO stand-in. Timings report batch-amortized throughput.

\subsection{Modality Analysis Settings}
\label[appendixsubsection]{sec:suppl:analysis:settings}

\begin{figure*}[t]
  \centering
  \resizebox{\textwidth}{!}{\begin{tikzpicture}[
  x=1cm, y=1cm,
  font=\fontfamily{phv}\selectfont\footnotesize,
  blk/.style={draw=#1!85, line width=0.5pt, fill=#1!7, rounded corners=3pt,
              align=center, inner sep=2pt, text width=25mm, minimum height=11.5mm},
  blk/.default=npggray,
  inp/.style={blk=#1, fill=white},
  frozen/.style={blk=npgblue, dash pattern=on 2pt off 1.3pt},
  head/.style={blk=npgteal, text width=33mm, minimum height=9.5mm},
  side/.style={blk, fill=white, dash pattern=on 2pt off 1.3pt, text width=30mm, minimum height=9mm},
  lane/.style={draw=none, fill=#1!4, rounded corners=5pt, inner xsep=5pt, inner ysep=4pt},
  lanelab/.style={font=\fontfamily{phv}\selectfont\scriptsize\bfseries, text=#1!85!black,
                  anchor=south west, inner sep=0pt},
  dim/.style={font=\fontfamily{phv}\selectfont\tiny, text=black!55, fill=white,
              inner sep=1pt, midway},
  flow/.style={-{Stealth[length=1.6mm,width=1.3mm]}, draw=black!50, line width=0.55pt},
  aux/.style={flow, dash pattern=on 2pt off 1.4pt},
]
\newcommand{\op}[2]{#1\\[-1pt]{\scriptsize\color{black!55}\renewcommand{\arraystretch}{0.95}\begin{tabular}[t]{@{}c@{}}#2\end{tabular}}}
\newcommand{\x}{\,$\times$\,}
\def\cA{0}\def\cB{3.15}\def\cC{6.3}\def\cD{9.45}\def\cE{11.55}\def\cF{13.75}\def\cG{17.6}
\def\rV{2.5}\def\rF{1.25}\def\rE{0}\def\rX{-1.45}
\node[inp=npgblue] (v0) at (\cA,\rV) {\op{Egocentric frame}{RGB}};
\node[frozen]      (v1) at (\cB,\rV) {\op{DINOv2}{frozen, pooled}};
\node[blk=npgblue] (v3) at (\cD,\rV) {\op{Linear}{GELU}};
\draw[flow] (v0) -- (v1);
\draw[flow] (v1) -- node[dim]{768} (v3);
\node[inp=npgorange, minimum height=13mm] (e0) at (\cA,\rE) {\op{sEMG window}{18\x400}};
\node[blk=npgorange, minimum height=13mm] (e1) at (\cB,\rE) {\op{Conv1d}{\textit{k}\,=\,9, \textit{s}\,=\,2\\GN, GELU}};
\node[blk=npgorange, minimum height=13mm] (e2) at (\cC,\rE) {\op{Conv1d}{\textit{k}\,=\,7, \textit{s}\,=\,2\\GN, GELU}};
\node[blk=npgorange, minimum height=13mm] (e3) at (\cD,\rE) {\op{Time mean}{Linear}};
\draw[flow] (e0) -- (e1);
\draw[flow] (e1) -- node[dim]{64} (e2);
\draw[flow] (e2) -- node[dim]{128} (e3);
\begin{scope}[on background layer]
  \node[lane=npgblue, fit=(v1)(v3)] (laneV) {};
  \node[lane=npgorange, fit=(e1)(e3)] (laneE) {};
\end{scope}
\node[lanelab=npgblue] at ([yshift=2pt]laneV.north west) {Vision encoder};
\node[lanelab=npgorange] at ([yshift=2pt]laneE.north west) {sEMG encoder};
\node[circle, draw=npggray!85, fill=npggray!7, line width=0.5pt, minimum size=6.5mm,
      inner sep=0pt, font=\fontfamily{phv}\selectfont\small] (cat) at (\cE,\rF) {$\Vert$};
\node[font=\fontfamily{phv}\selectfont\scriptsize, text=black!55, anchor=east, inner sep=2pt]
  at (cat.west) {concat};
\node[blk] (mlp) at (\cF,\rF) {\op{Fusion MLP}{2 layers, 256 hidden}};
\draw[flow] (v3.east) -| node[dim, pos=0.25]{128} (cat.north);
\draw[flow] (e3.east) -| node[dim, pos=0.25]{128} (cat.south);
\draw[flow] (cat) -- node[dim, above=0.5pt, fill=none]{256} (mlp);
\node[head] (hp) at (\cG,\rV) {\op{Pose head}{6D rotations, shape}};
\node[head] (hc) at (\cG,\rF) {\op{Contact head}{per region, BCE}};
\node[head] (hm) at (\cG,\rE) {\op{Magnitude head}{softplus, L1 on GT contact}};
\coordinate (split) at ($(mlp.east)!0.45!(hc.west)$);
\draw[flow] (mlp.east) -- (hc.west);
\draw[flow] (split) |- (hp.west);
\draw[flow] (split) |- (hm.west);
\fill[black!50] (split) circle (0.9pt);
\node[side] (dtwo) at (\cF,\rX) {\op{D2 pose path}{sEMG feature zeroed}};
\draw[aux] (dtwo) -- (mlp);
\node[side, text width=33mm] (reg) at (\cG,\rX) {\op{Region embedding}{and hand side}};
\draw[aux] (reg) -- (hm);
\draw[aux] (reg.east) -- ++(0.32,0) |- (hc.east);
\end{tikzpicture}}
  \caption{Neutral baseline for \cref{sec:empirical}. Frozen DINOv2 features and an $18\times400$ sEMG window are independently projected to 128 dimensions and fused by a two-layer MLP with 256 hidden units. Parallel heads predict MANO pose and shape, regional contact, and magnitude. Region heads also receive a region embedding and hand side. Dashed arrows show region inputs and D2's zeroed sEMG.}
  \label[appendixfigure]{fig:suppl:analysis_network}
\end{figure*}
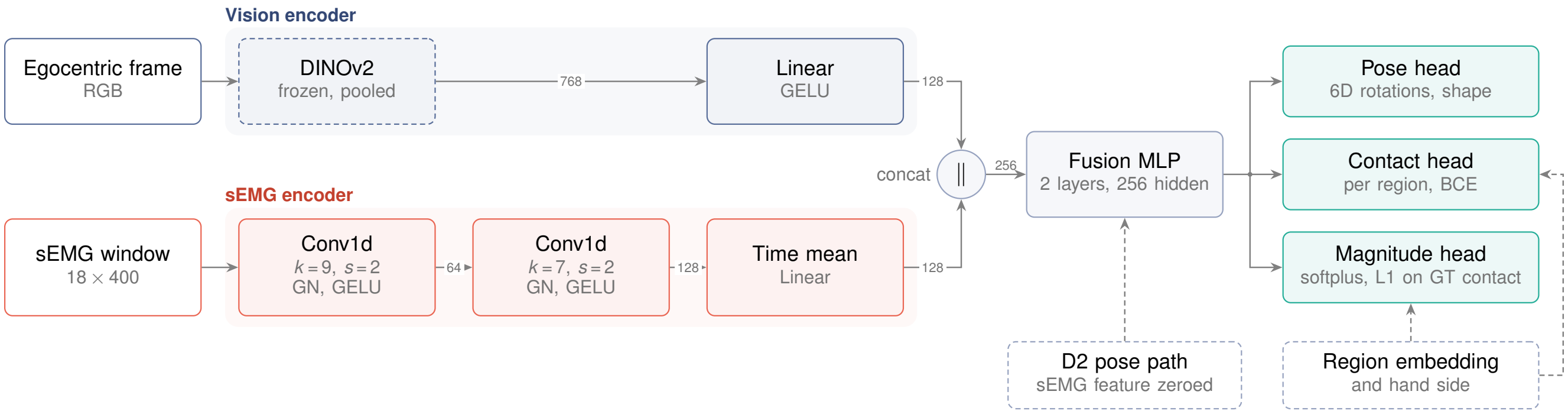

\myPara{Analysis Network} \cref{sec:empirical} uses the neutral baseline in \cref{fig:suppl:analysis_network}, independently of MyoBridge. A frozen, pooled 768-dimensional DINOv2 feature of the current frame is projected to 128 dimensions with a linear layer and GELU. The trailing 18-channel, 400-sample sEMG window passes through two strided 1D convolutions with 64/128 channels, kernels 9/7, stride 2, group normalization, and GELU, then temporal averaging and a linear projection to 128 dimensions. A two-layer MLP with 256 hidden units fuses the concatenated features. Parallel heads predict MANO rotations in 6D and shape (106 values), regional contact, and force magnitude. Region heads also read a learned region embedding and hand side. Contact uses binary cross-entropy. Softplus magnitude uses $\ell_1$ loss on reference-contacting regions only. Heads decode independently.

\myPara{Variants} \cref{tab:suppl:variants} compares five independently trained variants with the same data, supervision, and schedule.
\emph{NB-V} decodes all three tasks from the 128-dimensional visual feature alone, providing the vision reference.
\emph{NB-E} decodes the same tasks from the trailing 200\,ms sEMG window alone.
\emph{NB-VE} uses both encoders and the full fusion network in \cref{fig:suppl:analysis_network}.
\emph{NB-VE-mis} keeps the NB-VE architecture but replaces each sEMG window with one from another trial, matched by participant, hand, session, and ground-truth contact configuration within the same split. Donor selection uses contact labels without force values. Assignments are fixed before training and used in both training and evaluation. This label-informed control preserves contact configuration while disrupting frame-specific muscle activity, testing the value of synchronized sEMG.
\emph{D2} keeps the full fused feature for contact and magnitude, but computes pose through the same fusion MLP with the sEMG feature zeroed. The contact and force architecture is unchanged, with all weights retrained. Its pose gap to NB-VE measures the sEMG path's contribution.

\begin{table}[t]
  \centering
  \footnotesize
  \setlength{\tabcolsep}{4pt}
  \caption{Analysis variants and the input each head reads. $V$ and $E$ are the visual and sEMG features of the frame, $\tilde{E}$ the sEMG feature of a window from another trial with the same contact configuration, and $E$\,=\,0 a zeroed sEMG feature. NB-V and NB-E have no encoder at all for the other modality.}
  \label[appendixtable]{tab:suppl:variants}
  \begin{tabular}{@{}llll@{}}
    \toprule
    \textbf{Variant} & \textbf{Pose head} & \textbf{Region heads} & \textbf{Comparison} \\
    \midrule
    NB-V & $V$ & $V$ & vision alone (reference) \\
    NB-E & $E$ & $E$ & sEMG alone \\
    NB-VE & $V+E$ & $V+E$ & adding sEMG, vs.\ NB-V \\
    NB-VE-mis & $V+\tilde{E}$ & $V+\tilde{E}$ & matched sEMG, vs.\ NB-VE \\
    D2 & $V$, $E$\,=\,0 & $V+E$ & sEMG on pose, vs.\ NB-VE \\
    \bottomrule
  \end{tabular}
\end{table}

\myPara{Force Composition} Each trained variant supplies three evaluation-time estimates. Gated force multiplies magnitude by binary predicted contact at threshold 0.5, oracle force uses reference contact, and ungated force uses magnitude alone. Magnitude MAE covers reference-contacting regions only. Total force MAE compares sums over seven regions. A violation is a frame with more than \fdViolTau\,N on any reference non-contacting region. Zero force and per-region median force over contacting TRAIN frames provide trivial references. These binary analysis gates differ from MyoBridge's probability-weighted CSFF gate.

\myPara{Training and Selection} All variants share AdamW with weight decay $10^{-4}$, batch size 256, gradient clipping at 1, unit loss weights, 20 epochs with one epoch of linear warmup and cosine decay, and final-epoch evaluation. A rule fixed before run completion selects the peak learning rate from $10^{-3}$ and $3\times10^{-4}$ on a TRAIN-only development split. It compares the learning-rate difference in mean NB-V and NB-VE validation loss over two paired runs with between-run variation, retaining $3\times10^{-4}$ unless the former is larger. Selection uses validation losses independently of between-variant contrasts. The selected $3\times10^{-4}$ is shared by all variants, each trained for three runs on full cross-trial TRAIN and evaluated once on TEST.

\myPara{Statistics} This modality analysis weights participants equally. Errors average frames within trials, trials within hands, and hands within participants. Contact F1 pools frame--region decisions within each participant before averaging participants, distinct from the main benchmark's regional macro-F1. Contrasts average paired participant differences over three runs, with percentile 95\% CIs from 10{,}000 participant bootstrap resamples and two-sided sign-flip tests from 10{,}000 random assignments. All eleven pre-registered contrasts form one Holm family and reach adjusted $p$-value 0.0011 (\cref{tab:suppl:findings_contrasts}). Exploratory regional tests form four seven-region Holm families, comparing fusion with vision for F1 and magnitude, and fusion with mismatch and sEMG with vision for magnitude. Cross-region Spearman tests permute all $7!$ orderings without multiplicity correction. Across-run SEs in \cref{fig:empirical,tab:suppl:findings} describe run variation, while \cref{tab:suppl:findings_regions} reports run means.

\subsection{Modality Analysis Results}
\label[appendixsubsection]{sec:suppl:analysis:results}

\cref{tab:suppl:findings} lists the pooled axes of \cref{fig:empirical:pose,fig:empirical:force} for all five variants, and \cref{tab:suppl:findings_regions} the per-region values behind \cref{fig:empirical:contact_gain,fig:empirical:force_gain}. The pre-registered contrasts are given in \cref{tab:suppl:findings_contrasts}. The two trivial references belong to no variant and are therefore not tabulated. Predicting zero gives a magnitude error of \fdZeroMag\,N and a total error of \fdZeroTotal\,N, and predicting the per-region median of the TRAIN split gives \fdMedianMag\,N and \fdMedianTotal\,N.

\begin{table*}[t]
  \centering
  \small
  \caption{Variants behind \cref{sec:empirical} on the cross-trial TEST split, mean\,$\pm$\,standard error over three runs, where a standard error shown as 0.00 is rounded rather than exactly zero. Force is composed on predicted contact unless marked GT, and the violation rate counts frames with more than \fdViolTau\,N on a non-contacting region. NB-VE-mis feeds each frame an sEMG window from another trial.}
  \label[appendixtable]{tab:suppl:findings}
  \setlength{\tabcolsep}{6pt}
\begin{tabular}{@{}lr@{}lr@{}lr@{}lr@{}lr@{}l@{}}
\toprule
\textbf{Metric} & \multicolumn{2}{c}{\textbf{NB-V}} & \multicolumn{2}{c}{\textbf{NB-E}} & \multicolumn{2}{c}{\textbf{NB-VE}} & \multicolumn{2}{c}{\textbf{NB-VE-mis}} & \multicolumn{2}{c}{\textbf{D2}} \\
\midrule
MPJPE (mm) $\downarrow$ & 7&.39\,$\pm$\,0.03 & 12&.40\,$\pm$\,0.02 & 6&.89\,$\pm$\,0.02 & 7&.54\,$\pm$\,0.03 & 7&.27\,$\pm$\,0.03 \\
\addlinespace[3pt]
Contact F1 (\%) $\uparrow$ & 84&.29\,$\pm$\,0.16 & 70&.59\,$\pm$\,0.14 & 89&.21\,$\pm$\,0.07 & 85&.26\,$\pm$\,0.19 & 88&.97\,$\pm$\,0.04 \\
\addlinespace[3pt]
Force magnitude MAE (N) $\downarrow$ & 6&.35\,$\pm$\,0.03 & 5&.33\,$\pm$\,0.02 & 4&.28\,$\pm$\,0.00 & 6&.42\,$\pm$\,0.03 & 4&.29\,$\pm$\,0.01 \\
Total force MAE (N) $\downarrow$ & 8&.76\,$\pm$\,0.04 & 8&.47\,$\pm$\,0.06 & 5&.25\,$\pm$\,0.01 & 8&.71\,$\pm$\,0.04 & 5&.29\,$\pm$\,0.02 \\
Total force MAE, GT contact (N) $\downarrow$ & 7&.31\,$\pm$\,0.03 & 5&.62\,$\pm$\,0.02 & 4&.48\,$\pm$\,0.00 & 7&.41\,$\pm$\,0.04 & 4&.49\,$\pm$\,0.01 \\
\addlinespace[3pt]
Violation rate (\%) $\downarrow$ & 18&.68\,$\pm$\,0.19 & 26&.80\,$\pm$\,0.10 & 12&.68\,$\pm$\,0.10 & 15&.99\,$\pm$\,0.32 & 13&.05\,$\pm$\,0.05 \\
\bottomrule
\end{tabular}

\end{table*}

\begin{table}[t]
  \centering
  \small
  \caption{Regional F1 and magnitude MAE (N) behind \cref{fig:empirical:contact_gain,fig:empirical:force_gain}, averaged over three cross-trial TEST runs. Vision, sEMG, Fusion, and Mismatch denote NB-V, NB-E, NB-VE, and NB-VE-mis. T/I/M/R/P denote digits, U/L upper/lower palm.}
  \label[appendixtable]{tab:suppl:findings_regions}
  \footnotesize
\setlength{\tabcolsep}{3pt}
\resizebox{\linewidth}{!}{%
\begin{tabular}{@{}lccccccc@{}}
\toprule
\textbf{Metric} & \textbf{T} & \textbf{I} & \textbf{M} & \textbf{R} & \textbf{P} & \textbf{U} & \textbf{L} \\
\midrule
Vision F1 & 0.878 & 0.835 & 0.812 & 0.802 & 0.872 & 0.895 & 0.896 \\
sEMG F1 & 0.720 & 0.686 & 0.665 & 0.715 & 0.735 & 0.726 & 0.734 \\
Fusion F1 & 0.910 & 0.886 & 0.873 & 0.878 & 0.910 & 0.911 & 0.908 \\
Mismatch F1 & 0.889 & 0.837 & 0.817 & 0.814 & 0.884 & 0.927 & 0.928 \\
\midrule
Vision Mag. & 4.02 & 4.75 & 4.68 & 3.90 & 3.73 & 19.84 & 25.15 \\
sEMG Mag. & 3.41 & 3.49 & 3.58 & 3.06 & 3.19 & 15.50 & 27.69 \\
Fusion Mag. & 2.89 & 2.97 & 3.03 & 2.57 & 2.72 & 12.50 & 20.17 \\
Mismatch Mag. & 4.07 & 4.84 & 4.75 & 3.98 & 3.78 & 19.72 & 25.15 \\
\bottomrule
\end{tabular}}

\end{table}

\begin{table*}[!h]
  \centering
  \small
  \caption{The eleven pre-registered paired contrasts of \cref{sec:empirical} on the cross-trial TEST split (\fdParticipants\ participants). Differences are $a-b$ in the stated metric, averaged over runs within participant, with a participant-level bootstrap 95\% CI and a sign-flip $p$-value adjusted by Holm over all eleven contrasts. Unless stated otherwise, force is composed on the predicted contact of each model.}
  \label[appendixtable]{tab:suppl:findings_contrasts}
\small
\begin{tabular*}{0.78\textwidth}{@{\extracolsep{\fill}}llccc@{}}
\toprule
\textbf{Contrast ($a-b$)} & \textbf{Metric} & \textbf{Difference} & \textbf{95\% CI} &
$\boldsymbol{p_{\mathrm{Holm}}}$ \\
\midrule
sEMG $-$ vision & MPJPE (mm) & \phantom{0}+5.01 & [\phantom{0}+4.61, \phantom{0}+5.40] & 0.0011 \\
sEMG $-$ vision & Contact F1 (points) & \textminus13.70 & [\textminus15.40, \textminus12.20] & 0.0011 \\
sEMG $-$ vision & Magnitude MAE (N) & \phantom{0}\textminus1.02 & [\phantom{0}\textminus1.33, \phantom{0}\textminus0.73] & 0.0011 \\
Fusion $-$ vision & Magnitude MAE (N) & \phantom{0}\textminus2.07 & [\phantom{0}\textminus2.43, \phantom{0}\textminus1.72] & 0.0011 \\
Fusion $-$ vision & Contact F1 (points) & \phantom{0}+4.90 & [\phantom{0}+4.50, \phantom{0}+5.30] & 0.0011 \\
Fusion $-$ vision & Total force MAE (N) & \phantom{0}\textminus3.50 & [\phantom{0}\textminus4.03, \phantom{0}\textminus2.99] & 0.0011 \\
Fusion $-$ mismatch & Magnitude MAE (N) & \phantom{0}\textminus2.14 & [\phantom{0}\textminus2.49, \phantom{0}\textminus1.79] & 0.0011 \\
Fusion $-$ mismatch & Total force MAE (N) & \phantom{0}\textminus3.45 & [\phantom{0}\textminus3.99, \phantom{0}\textminus2.94] & 0.0011 \\
Vision-only pose $-$ fusion & MPJPE (mm) & \phantom{0}+0.38 & [\phantom{0}+0.34, \phantom{0}+0.42] & 0.0011 \\
Gated $-$ ungated (fusion) & Violation rate (points) & \textminus76.00 & [\textminus77.70, \textminus74.30] & 0.0011 \\
GT $-$ predicted contact (fusion) & Total force MAE (N) & \phantom{0}\textminus0.78 & [\phantom{0}\textminus0.89, \phantom{0}\textminus0.67] & 0.0011 \\
\bottomrule
\end{tabular*}

\end{table*}

\setcounter{algorithm}{0}
\renewcommand{\thealgorithm}{A\arabic{algorithm}}
\providecommand{\theHalgorithm}{\thealgorithm}
\renewcommand{\theHalgorithm}{appendix.\arabic{algorithm}}
\begin{algorithm}[!t]
\caption{Forward pass and training objective.}
\label[appendixalgorithm]{alg:egomyo}
\small
\begin{algorithmic}[1]
\Require video $V_{t-W_V:t}$, sEMG $E_{t-W_E:t}$, mode
\Ensure $\hat{M}_t$, $\hat{K}_t$, $\hat{c}_t$, $\hat{f}_t$
\State $Z^v_t\gets P_v(g_V(V_{t-W_V:t}))$ \Comment{frozen WiLoR~\cite{potamias2024wilor}}
\State $X^e_t\gets f_{\mathrm{stem}}(E_{t-W_E:t})$
\State $Z^e_t\gets\operatorname{LN}(\operatorname{Mean}_{\tau}[X^e_t+f_{\mathrm{TCN}}(X^e_t)]+B_{\mathrm{ch}})$
\State $[\bar{Z}^v_t;\bar{Z}^e_t]\gets\mathrm{MoT}([Z^v_t;Z^e_t])$
\State $u^v_t\gets\operatorname{Mean}(\bar{Z}^v_t)$, $u^e_t\gets\operatorname{Mean}(\bar{Z}^e_t)$
\State $H^{(0)}_t\gets\operatorname{reshape}_{8\times d}(f_{\mathrm{pool}}([u^v_t;u^e_t]))$
\State $E_t\gets\operatorname{reshape}_{8\times d}(W_eu^e_t+b_e)$
\For{each regional or global state $r$}
  \State $g_{t,r}\gets\sigma(f_{\mathrm{gate}}(h^{(0)}_{t,r}))$
  \State $\widetilde{h}_{t,r}\gets h^{(0)}_{t,r}+g_{t,r}E_{t,r}$
  \State $h_{t,r}\gets0.9\widetilde{h}_{t,r}+0.1\operatorname{LN}(\widetilde{h}_{t,r})$
\EndFor
\State $H_t\gets[h_{t,T};\ldots;h_{t,L};h_{t,\mathrm{global}}]$
\State $(\hat{\theta}_t,\hat{\beta}_t)\gets D_M(H_t)$
\State $\hat{M}_t\gets\mathcal{M}(\hat{\theta}_t,\hat{\beta}_t)$, $\hat{K}_t\gets\mathcal{J}\hat{M}_t$
\For{each vertex $v$}
  \State $q_{t,v}\gets e_v+h_{t,r(v)}+\operatorname{PE}(\hat{M}_t(v))$
  \State $z_{t,v}\gets\operatorname{Attn}(q_{t,v},[\bar{Z}^v_t;\bar{Z}^e_t;H_t])$
  \State $\hat{c}_t(v)\gets\sigma(D_c(z_{t,v}))$, $s_t(v)\gets D_s(z_{t,v})$
  \State $\tilde{c}_t(v)\gets\hat{c}_t(v)$ \Comment{training}
  \State at inference, $\tilde{c}_t(v)\gets\hat{c}_t(v)\mathbf{1}[\hat{c}_t(v)\geq0.5]$
\EndFor
\For{each region $r$}
  \State $\hat{C}_{t,r}\gets\max_{v\in r}\tilde{c}_t(v)$
  \State $\hat{F}_{t,r}\gets\hat{C}_{t,r}\operatorname{softplus}(D_m(h_{t,r}))$
  \If{$\hat{C}_{t,r}>0$}
    \State $\hat{p}_t(v)\gets\dfrac{\tilde{c}_t(v)e^{s_t(v)}}{\sum_{u\in r}\tilde{c}_t(u)e^{s_t(u)}}$, $v\in r$
  \Else
    \State $\hat{p}_t(v)\gets0$ for $v\in r$
  \EndIf
  \State $\hat{f}_t(v)\gets\hat{F}_{t,r}\hat{p}_t(v)$, $v\in r$
\EndFor
\State \textbf{Train only} with $\mathcal{L}$ in \cref{eq:total_loss}
\end{algorithmic}
\end{algorithm}

\section{Method Details}
\label{sec:suppl:method}

\textbf{MyoBridge} combines \emph{Myo} for muscle activity and \emph{Bridge} for the regional pathway to the hand surface.

\myPara{Execution Order} \cref{alg:egomyo} follows GME, RGF, and RSD in \cref{sec:method}. RGF mean-pools each stream, maps their concatenation to seven regional states and one global state, adds gated sEMG residuals, and applies state normalization. The resulting $H_t$ predicts MANO pose and shape. The mesh determines joints through the fixed regressor $\mathcal{J}$ and anchors vertex queries. Frozen WiLoR supplies visual features without an explicit pose or shape prior.

\myPara{Surface Reconstruction and Supervision} RSD combines vertex embeddings, regional states, and predicted positions into vertex queries. CSFF uses soft contact during training and threshold-masked probabilities at inference (\cref{eq:csff}). Empty support gives zero force and distribution. Nonempty support allocates regional force with normalized contact-weighted scores, preserving $\sum_{v\in r}\hat{f}_t(v)=\hat{F}_{t,r}$. Training weights mesh, contact, regional-force, and distribution losses by learned uncertainty (\cref{eq:total_loss}).

\begin{figure*}[tp]
  \centering
  \begin{subcaptiongroup}%
    \phantomsubcaption\label[appendixfigure]{fig:suppl:typical_reconstruction:a}%
    \phantomsubcaption\label[appendixfigure]{fig:suppl:typical_reconstruction:b}%
    \phantomsubcaption\label[appendixfigure]{fig:suppl:typical_reconstruction:c}%
  \end{subcaptiongroup}%
  \includegraphics[height=3.0in]{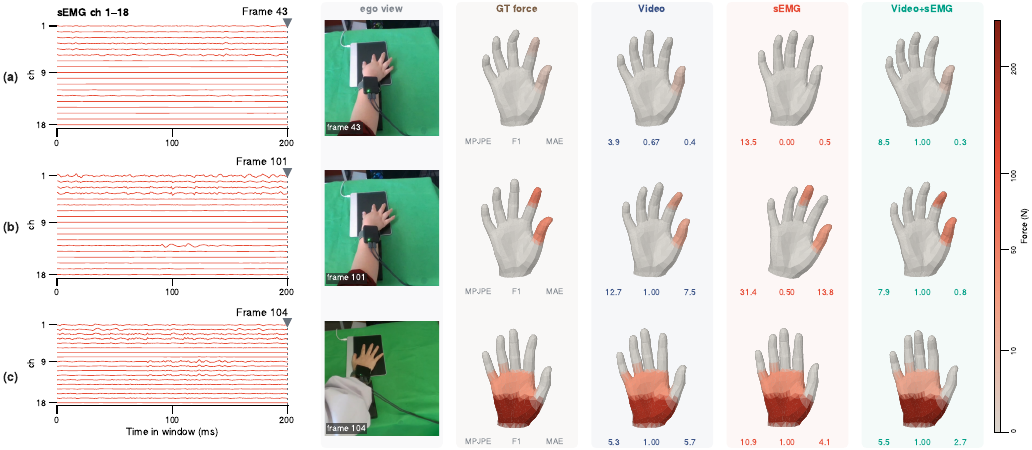}
  \caption{Representative digit and palm reconstruction by the neutral-baseline models. Rows \subref{fig:suppl:typical_reconstruction:a} and \subref{fig:suppl:typical_reconstruction:b} show thumb--index contact at lower and higher force, and \subref{fig:suppl:typical_reconstruction:c} shows palm contact. NB-VE recovers the reference contacts with lower regional force MAE than either single stream in all three examples. Left traces show all 18 sEMG channels in the preceding 200\,ms, with the displayed frame at the 200\,ms window endpoint. Signals share one TRAIN-normalized amplitude scale across channels and rows.}
  \label[appendixfigure]{fig:suppl:typical_reconstruction}
\end{figure*}

\begin{figure*}[tp]
  \centering
  \begin{subcaptiongroup}%
    \phantomsubcaption\label[appendixfigure]{fig:suppl:final_successes:a}%
    \phantomsubcaption\label[appendixfigure]{fig:suppl:final_successes:b}%
    \phantomsubcaption\label[appendixfigure]{fig:suppl:final_successes:c}%
    \phantomsubcaption\label[appendixfigure]{fig:suppl:final_successes:d}%
  \end{subcaptiongroup}%
  \includegraphics[width=\textwidth]{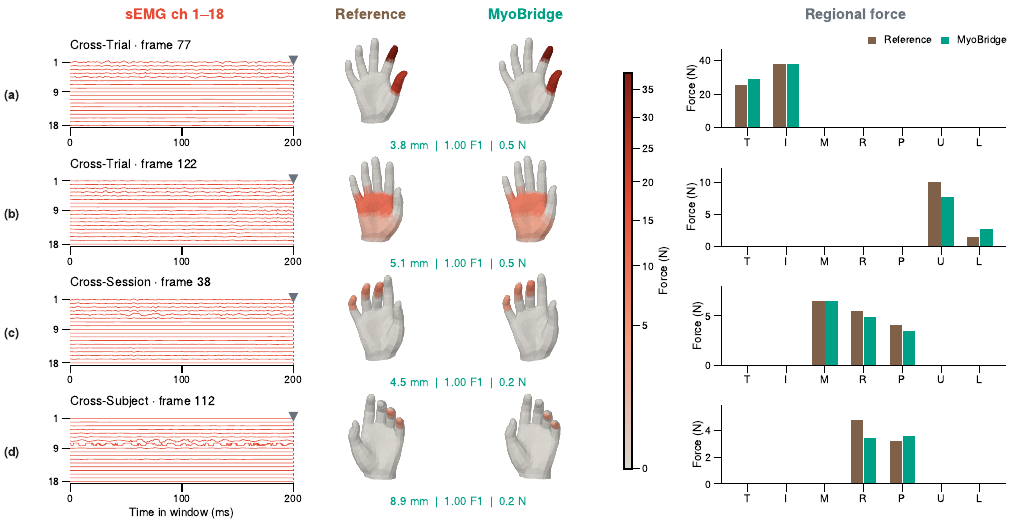}
  \caption{Successful reconstructions by the final MyoBridge checkpoints on TEST trials from all three protocols. Each row pairs the reference and predicted mesh with regional forces and the preceding 200\,ms of 18-channel sEMG. The marked 200\,ms endpoint corresponds to the displayed frame. Signals use one amplitude scale across channels within each row. Rows \subref{fig:suppl:final_successes:a} and \subref{fig:suppl:final_successes:b} come from cross-trial, \subref{fig:suppl:final_successes:c} from cross-session, and \subref{fig:suppl:final_successes:d} from cross-subject TEST trials. Metrics below the meshes give MPJPE, contact F1, and regional force MAE. T, I, M, R, P, U, and L denote thumb, index, middle, ring, little, upper palm, and lower palm.}
  \label[appendixfigure]{fig:suppl:final_successes}
\end{figure*}

\begin{figure*}[tp]
  \centering
  \begin{subcaptiongroup}%
    \phantomsubcaption\label[appendixfigure]{fig:suppl:visual_ambiguity:a}%
    \phantomsubcaption\label[appendixfigure]{fig:suppl:visual_ambiguity:b}%
    \phantomsubcaption\label[appendixfigure]{fig:suppl:visual_ambiguity:c}%
  \end{subcaptiongroup}%
  \includegraphics[height=3.0in]{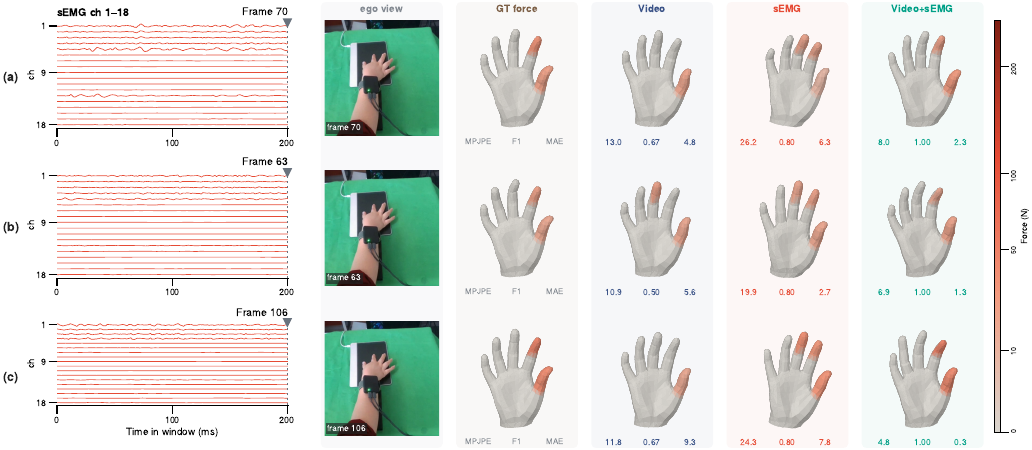}
  \caption{Contact recovery under visual ambiguity. The visual stream misses index contact in \subref{fig:suppl:visual_ambiguity:a} and \subref{fig:suppl:visual_ambiguity:c}, and predicts middle instead of index contact in \subref{fig:suppl:visual_ambiguity:b}. NB-VE recovers the reference thumb--index contacts with lower pose and regional force errors than either single stream in these examples. Left traces follow the displayed frames. Both neutral-baseline figures share the force scale and metric order.}
  \label[appendixfigure]{fig:suppl:visual_ambiguity}
\end{figure*}

\begin{figure*}[tp]
  \centering
  \includegraphics[width=\textwidth]{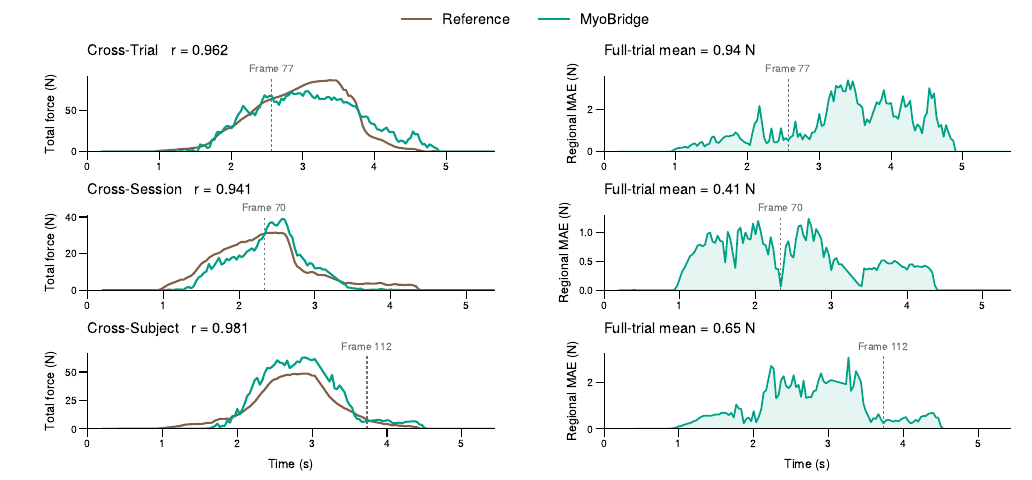}
  \caption{Complete TEST-trial force trajectories from the final MyoBridge checkpoints. Left plots compare reference and predicted total force summed over seven regions. Right plots show frame-level regional MAE. Dashed cursors mark representative frames. Internal gaps are linearly interpolated for display. Statistics use all valid frames, excluding interpolated values.}
  \label[appendixfigure]{fig:suppl:final_temporal}
\end{figure*}

\begin{figure*}[tp]
  \centering
  \includegraphics[width=\textwidth]{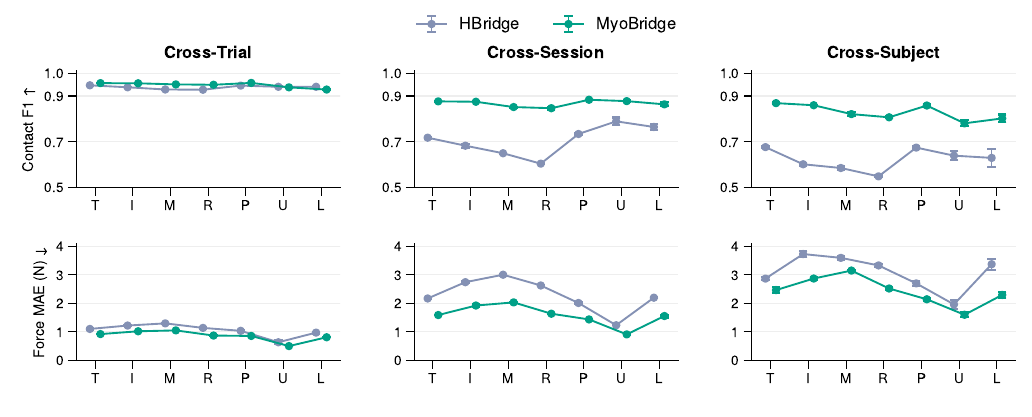}
  \caption{Regional contact and force comparison on the complete TEST sets. Points and error bars show the mean and sample standard deviation across three independent runs for HBridge and MyoBridge. Higher contact F1 and lower force MAE indicate better performance. Region abbreviations follow \cref{fig:suppl:final_successes}. Each run uses the complete TEST set and the benchmark masks and aggregation.}
  \label[appendixfigure]{fig:suppl:regional_results}
\end{figure*}

\begin{table*}[tp]
  \centering
  \footnotesize
  \setlength{\tabcolsep}{3pt}
  \renewcommand{\arraystretch}{1.08}
  \caption{\textbf{Regional gating ablations.} Cross-trial TEST mean $\pm$ sample SD over three runs. Force MAE is in N.}
  \label[appendixtable]{tab:suppl:regional_gate_ablation}
  \begin{tabular}{@{}l*{6}{c}@{}}
    \toprule
    \multirow{2.5}{*}{\textbf{Variant}} & \multicolumn{2}{c}{\textbf{Kinematics (mm)}} & \multicolumn{2}{c}{\textbf{Region}} & \multicolumn{2}{c}{\textbf{Vertex}} \\
    \cmidrule(lr){2-3} \cmidrule(lr){4-5} \cmidrule(l){6-7}
    & MPJPE$\downarrow$ & PVE$\downarrow$ & F1$\uparrow$ & MAE$\downarrow$ & F1$\uparrow$ & MAE$\downarrow$ \\
    \midrule
    Global feature + global gate & 3.32\,$\pm$\,0.16 & 3.36\,$\pm$\,0.15 & 0.913\,$\pm$\,0.004 & 0.92\,$\pm$\,0.05 & 0.865\,$\pm$\,0.004 & 0.290\,$\pm$\,0.010 \\
    No regional gate & 4.11\,$\pm$\,0.10 & 4.14\,$\pm$\,0.10 & 0.929\,$\pm$\,0.004 & 1.30\,$\pm$\,0.30 & 0.866\,$\pm$\,0.002 & 0.292\,$\pm$\,0.004 \\
    \textbf{Per-region gate (MyoBridge)} & \textbf{3.26\,$\pm$\,0.03} & \textbf{3.30\,$\pm$\,0.03} & \textbf{0.948\,$\pm$\,0.001} & \textbf{0.86\,$\pm$\,0.01} & \textbf{0.879\,$\pm$\,0.000} & \textbf{0.289\,$\pm$\,0.002} \\
    \bottomrule
  \end{tabular}
\end{table*}

\begin{table*}[tp]
\centering
\small
\setlength{\tabcolsep}{5pt}
\begin{tabular}{lrrrrrrr}
\toprule
\textbf{Protocol} & \textbf{No contact} & \textbf{Single digit} & \textbf{Multiple digits} & \textbf{Palm} & \textbf{Lower force} & \textbf{Middle force} & \textbf{Higher force} \\
\midrule
Cross-Trial & 715,810 & 215,564 & 1,083,737 & 130,654 & 475,485 & 475,338 & 479,132 \\
Cross-Session & 897,711 & 264,644 & 1,340,437 & 162,240 & 589,349 & 570,806 & 607,166 \\
Cross-Subject & 537,665 & 163,133 & 822,613 & 99,600 & 330,823 & 344,865 & 409,658 \\
\bottomrule
\end{tabular}
\caption{Frame counts for reference-defined TEST groups. Contact groups require valid inputs and all seven contact labels. Palm includes mixed digit--palm contact. Force groups additionally require all seven force labels and at least one contacting region. Force thresholds are fitted independently on TRAIN for each protocol. Each grouping partitions its eligible frames.}
\label[appendixtable]{tab:suppl:group_coverage}
\end{table*}

\begin{figure*}[tp]
\centering
\includegraphics[width=\textwidth]{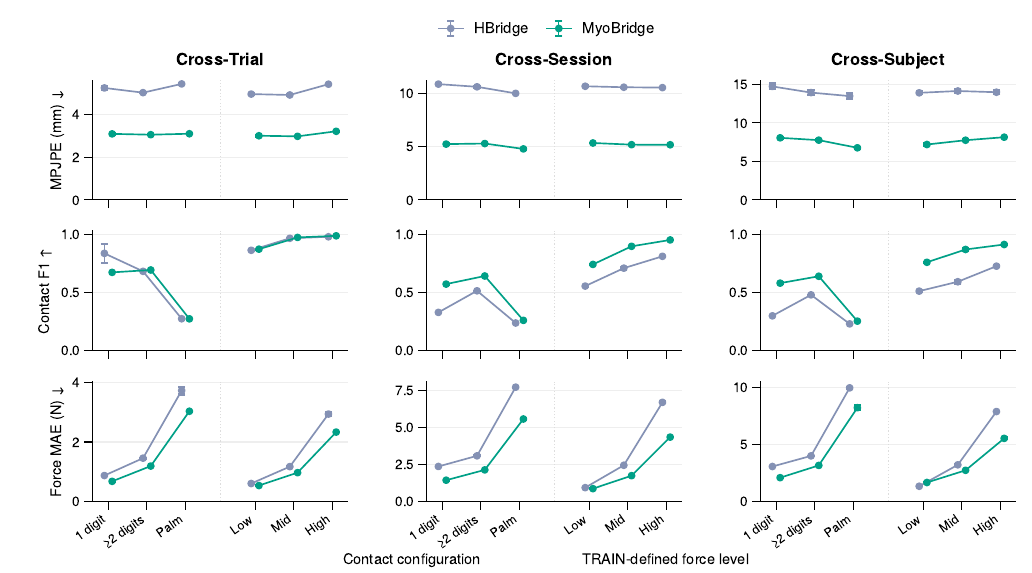}
\caption{Breakdown of \cref{tab:main} by reference contact configuration (left) and TRAIN-defined force level (right). Points and bars show means and sample SDs over three runs. Groups retain all eligible TEST frames, benchmark masks, and metric aggregation.}
\label[appendixfigure]{fig:suppl:contact_force_groups}
\end{figure*}

\begin{figure*}[t]
\centering
\includegraphics[width=\textwidth]{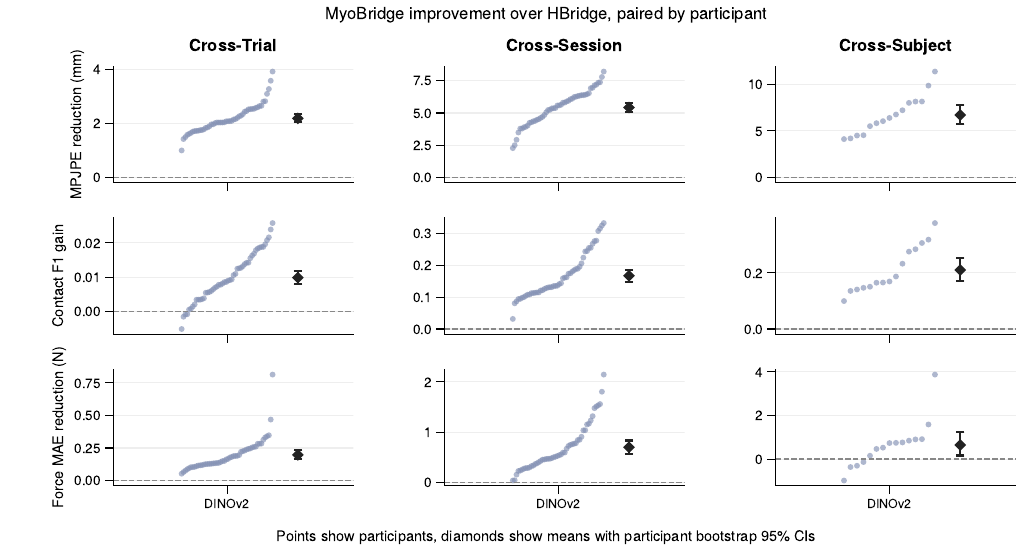}
\caption{Participant-level view of the MyoBridge--HBridge comparison in \cref{tab:main}. Positive values indicate lower MPJPE, higher contact F1, or lower regional force MAE. Every participant contributes one point after averaging three paired runs. Diamonds and bars show the equal-participant mean improvement and participant bootstrap 95\% confidence interval. All participants are retained.}
\label[appendixfigure]{fig:suppl:paired_performance}
\end{figure*}

\section{Additional Experimental Results}
\label{sec:suppl:additional_results}

\subsection{Successful Reconstruction}
\label[appendixsubsection]{sec:suppl:qualitative_successes}

\myPara{Case Selection and Display} \Cref{fig:suppl:typical_reconstruction,fig:suppl:visual_ambiguity} show six distinct frames from two cross-trial TEST sequences, supplementing the modality analysis in \cref{sec:empirical}. Video, sEMG, and Video+sEMG denote NB-V, NB-E, and NB-VE from \cref{sec:suppl:analysis:settings}. Mesh colors show regional force with predicted contact gated at 0.5. Metrics give MPJPE in mm, frame-level contact F1, and regional force MAE in N.

\myPara{Digit and Palm Reconstruction} \Cref{fig:suppl:typical_reconstruction} shows thumb--index contact at two force levels and contact over both palm regions. Fusion recovers the reference contact set in all three examples with lower regional force MAE than either single stream. At low force in \subref{fig:suppl:typical_reconstruction:a}, vision misses the index contact and sEMG detects no contact, while fusion recovers both digits. In \subref{fig:suppl:typical_reconstruction:b}, MAE is 0.8\,N, compared with 7.5\,N for vision and 13.8\,N for sEMG. The palm example in \subref{fig:suppl:typical_reconstruction:c} reaches 2.7\,N versus 5.7\,N and 4.1\,N, so fusion allocates force better over both palm regions.

\myPara{Final-Model Reconstruction across Protocols} \Cref{fig:suppl:final_successes} extends the neutral-baseline analysis with the final MyoBridge checkpoints. The four TEST frames cover thumb--index, palm, and multi-finger contact across the three protocols. All four recover the reference contact set, with regional force MAE between 0.2\,N and 0.5\,N. The cross-session example in \cref{fig:suppl:final_successes:c} reconstructs three contacting fingers with MPJPE of 4.5\,mm. The cross-subject example in \subref{fig:suppl:final_successes:d} separates ring and little-finger forces with MAE of 0.2\,N. The model recovers geometry, contact, and force jointly on unseen sessions and subjects.

\subsection{Contact Recovery under Visual Ambiguity}
\label[appendixsubsection]{sec:suppl:failure_cases}

\Cref{fig:suppl:visual_ambiguity} illustrates Finding~\hyperref[sec:empirical:finding2]{2} through missed or misassigned contacts in vision and spurious middle-finger contact in sEMG. Fusion recovers the reference contacts with F1 of 1.00 and the lowest pose and force errors in each row. Relative to vision, it lowers MPJPE by 37\% to 59\% across the three rows. In \subref{fig:suppl:visual_ambiguity:c}, force MAE is 0.3\,N, compared with 9.3\,N for vision and 7.8\,N for sEMG. Fusion resolves these errors with regional force MAE of 0.3--2.3\,N.

\subsection{Temporal Reconstruction}
\label[appendixsubsection]{sec:suppl:temporal_results}

\Cref{fig:suppl:final_temporal} extends \cref{tab:temporal_teaser} with TEST trajectories from all protocols. MyoBridge follows force onset, loading, and release, with total-force correlations of 0.962, 0.941, and 0.981 in the cross-trial, cross-session, and cross-subject trials. Regional MAEs are 0.94\,N, 0.41\,N, and 0.65\,N. These trials complement the temporal results in \cref{tab:temporal_teaser} with examples from unseen sessions and subjects.

\subsection{Regional Performance}
\label[appendixsubsection]{sec:suppl:regional_results}

\Cref{fig:suppl:regional_results} details the regional comparisons in \cref{tab:main}. MyoBridge has lower force MAE than HBridge in all seven regions and protocols. Reductions range from 16.4\% to 23.5\% in cross-trial, 26.1\% to 37.7\% in cross-session, and 12.4\% to 31.9\% in cross-subject evaluation. Contact F1 improves in all five digit regions for cross-trial and in all seven regions for cross-session and cross-subject, by 0.09 to 0.24 and 0.14 to 0.26. The largest gain falls on the ring finger in both, where Finding~\hyperref[sec:empirical:finding2]{2} finds vision least reliable.

\subsection{Regional Gating}
\label[appendixsubsection]{sec:suppl:regional_gate_ablation}

\Cref{tab:suppl:regional_gate_ablation} evaluates regional gating on cross-trial using frozen WiLoR features. The global variant uses one shared feature and gate, while No regional gate uses ungated regional states and direct heads in place of RSD. Per-region gating yields regional contact F1 of 0.948 versus 0.913 with global gating. Relative to No regional gate, MyoBridge reduces MPJPE from 4.11 to 3.26\,mm and regional force MAE from 1.30 to 0.86\,N, leading all six metrics. Its run-to-run SD is also the smallest on every metric, 0.01\,N regional force MAE against 0.30\,N without gating.

\subsection{Contact and Force Stratification}
\label[appendixsubsection]{sec:suppl:contact_force_groups}

We break down the complete TEST comparisons in \cref{tab:main} by reference contact configuration and force level. Contact groups comprise one digit without palm contact, multiple digits without palm contact, and any palm contact, including mixed digit--palm configurations. Group membership requires valid inputs and all seven contact labels. Force groups additionally require all seven force labels and at least one reference-contacting region. Their boundaries are total-force tertiles fitted independently on TRAIN for each protocol. \Cref{tab:suppl:group_coverage} gives the eligible frame counts. No-contact frames remain in the complete benchmark.

\Cref{fig:suppl:contact_force_groups} provides a stratified view of the geometry, contact, and regional-force comparisons in \cref{tab:main}. MyoBridge has lower regional force MAE than HBridge in each displayed contact-configuration group under all three protocols. For example, on cross-session multi-digit contact, MAE decreases from 3.08\,N to 2.13\,N. Across the three TRAIN-defined force levels, MyoBridge has lower MPJPE and higher contact F1 than HBridge in every group under all three protocols, and its MPJPE reduction stays between 39\% and 52\% across all contact and force groups. It also has lower regional force MAE in every middle- and higher-force group under all three protocols, and in all three force groups under cross-trial and cross-session. Its force-MAE reduction grows with reference force, from 6.70 to 4.35\,N in the higher-force cross-session group.

\subsection{Participant Consistency}
\label[appendixsubsection]{sec:suppl:paired_performance}

To examine the consistency of \cref{tab:main} across participants, we pair predictions on the same complete TEST trials. We pool metric sufficient statistics within each participant, compute paired differences, and average them over three runs. Participants then contribute equally to the summary. A recording denotes one participant, hand side, and session. Percentile 95\% confidence intervals use 10{,}000 paired participant bootstrap resamples. Recording-level intervals resample participants as clusters, retaining their recordings together. All units with valid labels contribute to each metric.

\Cref{fig:suppl:paired_performance} shows every participant's paired difference against HBridge with DINOv2. MyoBridge has lower MPJPE for every participant under all three protocols and higher contact F1 for 46 of 50, 50 of 50, and 15 of 15 participants. It has lower regional force MAE for 50 of 50 participants in both cross-trial and cross-session, and for 11 of 15 in cross-subject. The corresponding recording counts for force are 195 of 200, 98 of 100, and 44 of 60. These paired results show that the aggregate gains in \cref{tab:main} are shared across participants rather than driven by a few.

Equal-participant MPJPE reductions are 2.19\,mm, 5.43\,mm, and 6.71\,mm for cross-trial, cross-session, and cross-subject. Their participant bootstrap 95\% intervals all exclude zero, as do those of the regional force MAE reductions, at 0.17--0.23\,N, 0.57--0.84\,N, and 0.17--1.23\,N.

\section{Additional Discussion}
\label{sec:suppl:discussion}

\myPara{Complementary Evidence} The appendix results extend the three findings of \cref{sec:empirical} from aggregate statistics to individual frames, regions, and participants. In the neutral-baseline cases of \cref{fig:suppl:typical_reconstruction,fig:suppl:visual_ambiguity}, vision misses or misassigns digit contacts, sEMG adds spurious or missing contacts, and fusion recovers the reference contact set in all six frames with lower regional force MAE than either stream. These examples make Findings~\hyperref[sec:empirical:finding1]{1} and~\hyperref[sec:empirical:finding2]{2} concrete, since vision supplies where the hand is, sEMG how strongly it presses, and fusion helps most where vision errs.

\myPara{Region-Level Fusion} MyoBridge turns this complementarity into consistent regional gains. It lowers regional force MAE in all seven regions under every protocol, by 12.4\% to 37.7\%, with the largest contact gain on the ring finger under cross-session and cross-subject (\cref{fig:suppl:regional_results}). Per-region gating outperforms both a single global gate and ungated regional states on all six metrics in \cref{tab:suppl:regional_gate_ablation}, consistent with the RGF ablation in \cref{tab:ablation}. These results support admitting sEMG separately for each region rather than assigning one contribution to the whole hand.

\myPara{Robustness across Interaction States and Users} The advantage over HBridge holds across interaction states and individuals. MyoBridge has lower MPJPE in every contact-configuration and force-level group and higher contact F1 at every force level, and its force-MAE reduction grows with reference force (\cref{fig:suppl:contact_force_groups}). It improves MPJPE for every participant under all three protocols. Participant-bootstrap 95\% intervals for mean improvements in MPJPE, contact F1, and regional force MAE exclude zero in all protocols (\cref{fig:suppl:paired_performance}). Complete trajectories in \cref{fig:suppl:final_temporal} follow force onset, loading, and release on unseen sessions and subjects. Together with RSD, which confines force to the predicted contact support (\cref{eq:csff}) following Finding~\hyperref[sec:empirical:finding3]{3}, these results support joint geometry, contact, and force recovery on unseen users (\cref{tab:main}).

\end{document}